\documentclass[11pt]{article}

\usepackage[preprint]{acl}

\usepackage{subcaption} 

\usepackage[most]{tcolorbox}

\usepackage{array}

\usepackage{times}
\usepackage{latexsym}

\usepackage{multirow}
\usepackage{booktabs}

\usepackage{makecell}

\usepackage[table,dvipsnames]{xcolor}
\usepackage{colortbl}
\usepackage{enumitem}
\newtcolorbox{promptbox}[2][]{
    breakable,
    enhanced,
    colback=gray!3,
    colframe=gray!55,
    coltitle=black,
    fonttitle=\bfseries,
    title={#2},
    boxrule=0.6pt,
    arc=1.5pt,
    left=6pt,
    right=6pt,
    top=5.2pt,
    bottom=5.2pt,
    before skip=8pt,
    after skip=8pt,
    #1
}
\usepackage{listings}

\lstdefinestyle{promptcode}{
    basicstyle=\ttfamily\small,
    breaklines=true,
    breakatwhitespace=false,
    columns=fullflexible,
    keepspaces=true,
    showstringspaces=false,
    frame=none,
    aboveskip=2pt,
    belowskip=2pt
}
\usepackage[T1]{fontenc}

\usepackage[utf8]{inputenc}

\usepackage{microtype}
\usepackage{amsmath}
\usepackage{inconsolata}

\usepackage{graphicx}

\title{Prompt-Level Reward Specifications for Open-Ended Post-Training}

\author{
Zijun Weng$^{1,2}$,
Xiaohui Hu$^{2}$,
Shuangyong Song$^{2}$,
Yongxiang Li$^{2}$,
Kaidong Yu$^{2}$\thanks{\, \, Corresponding authors.},
Xuanjing Huang$^{1}$\footnotemark[1]
\\
$^1$Fudan University\\
$^2$Xingchen AGI Lab, China Telecom Artificial Intelligence Technology (Beijing) Co., Ltd.\\
\texttt{25113050287@m.fudan.edu.cn}\\
\texttt{yukd@chinatelecom.cn, xjhuang@fudan.edu.cn}
}

\begin{document}
\maketitle
\begin{abstract}
Open-ended post-training benefits from rewards that make prompt-specific success conditions explicit, rather than relying only on post-hoc scalar scores. In instruction following, writing, and decision-support tasks, response quality depends on local requirements, holistic preferences, and explicit constraints, but existing reward methods often leave these criteria implicit or cover only narrowly verifiable cases. We propose a prompt-level reward specification framework that separates reward specification from reward computation. Given only prompts, our framework constructs reusable task-adaptive rubrics and executable hard-constraint checkers offline, making reward criteria explicit before training and reusable across rollouts. At scoring time, artifact-anchored rubric and code scores are combined with an independent global score for residual holistic quality, yielding a normalized hybrid reward over requirement satisfaction, holistic quality, and deterministic constraints. The framework requires no human preference annotations, reference answers, or a separately trained reward model. Experiments show that the resulting reward improves offline RM-style response ranking and supports online reinforcement learning across multiple open-ended benchmarks. Ablations further show that rubrics, global scoring, and executable verification provide complementary supervision.
\end{abstract}

\section{Introduction}

Reward construction remains a central obstacle for open-ended language-model post-training. The difficulty is not simply to obtain a stronger scalar judge. For prompts in instruction following, writing, and decision support, a reward must determine what counts as success for that particular prompt: which local requirements should be satisfied, which qualities require holistic judgment, and which constraints can be checked exactly~\citep{flask}. We call this the \emph{\textbf{prompt-level reward specification problem}}. Transparent and consistent scoring therefore requires an explicit specification of what the reward should measure.

Existing post-training paradigms only partially address this problem. Reinforcement learning from human feedback (RLHF) and learned reward models provide broad preference supervision, but their prompt-specific criteria are usually implicit in human comparisons or model parameters~\citep{christiano2017deep,ouyang2022rlhf}. Reinforcement learning with verifiable rewards (RLVR) provides explicit and reliable supervision, but it relies on reference answers, executable tests, or other clean verification signals that are often unavailable in open-ended tasks~\citep{shao2024deepseekmath,guo2025deepseek,yu2026dapo}. Generic LLM judges can score open-ended responses~\citep{zheng2023judging,gu2025surveyllmasajudge}, but their criteria remain opaque. Rubric-based judges make criteria more explicit~\citep{arora2025healthbench,lmunit}, and recent work has used rubrics for reward modeling and online RL~\citep{xu2026rubricarm,liu2026openrubric,gunjal2025RaR,shao2025drtulureinforcementlearning,jia2026openrs}. However, these methods typically construct or use rubrics during scoring or optimization, rather than treating them as reusable prompt-level specifications built before rollout generation and shared across responses. Deterministic checkers are reliable for explicit constraints, but cover only requirements reducible to surface-level tests, such as length, format, or required strings~\citep{zhou2023instruction,pyatkin2026generalizing,followbench}.

These limitations suggest a different design principle: open-ended reward construction should separate reward specification from reward computation. Rather than relying on a judge to reconstruct all prompt-specific criteria each time a response is scored, the system should first construct reusable prompt-level reward artifacts that make decomposable and verifiable success conditions explicit. This provides an RLVR-like interface for open-ended tasks: the success conditions are specified before training and reused across rollouts, even though they are only partially verifiable.

We propose a prompt-level reward specification framework for open-ended post-training. Given only prompts, it constructs reusable task-adaptive rubrics and executable hard-constraint checkers offline. At scoring time, each prompt-response pair receives three complementary signals: rubric scoring for fine-grained requirements, global scoring for holistic quality, and code scoring for explicit constraints. These normalized signals are combined into a unified hybrid reward.

This positioning distinguishes our framework from prior rubric-based and hybrid-reward approaches. Our contribution is not simply to add rubrics or verifiers to an LLM judge, but to specify reward criteria before rollout generation and reuse them across scoring calls. This makes all candidate responses to the same prompt comparable under a shared specification, rather than asking the judge to rediscover or adapt the criteria for each response. Equally important, we treat local rubric-based supervision and global holistic judgment as complementary rather than interchangeable. Rubrics expose prompt-specific requirements and provide decomposed supervision, but independently scored criteria may miss response-level interactions among otherwise strong candidates. The global score complements this by capturing holistic coherence, usefulness, and trade-offs across criteria, while executable checkers provide deterministic feedback for explicit hard constraints.

Empirically, we evaluate the same prompt-level artifacts in two settings: offline response ranking and downstream online RL. The resulting reward performs strongly on offline reward-evaluation benchmarks and yields consistent gains across multiple open-ended RL benchmarks. Our contributions are as follows:
\begin{itemize}[leftmargin=*, itemsep=1pt, topsep=2pt, parsep=0pt, partopsep=0pt]
    \item We formulate open-ended reward construction as a prompt-level reward specification problem, arguing that rewards should expose prompt-specific success conditions before scoring responses.
    \item We propose a prompt-only reward specification framework that separates reward specification from reward computation by constructing reusable task-adaptive rubrics and executable hard-constraint checkers offline.
    \item We instantiate these artifacts as a pointwise hybrid reward combining local rubric scoring, independent global scoring, and executable verification, and show its effectiveness for both offline reward evaluation and online RL.
\end{itemize}
\begin{figure*}[t!]
    \centering
    \includegraphics[width=0.98\textwidth]{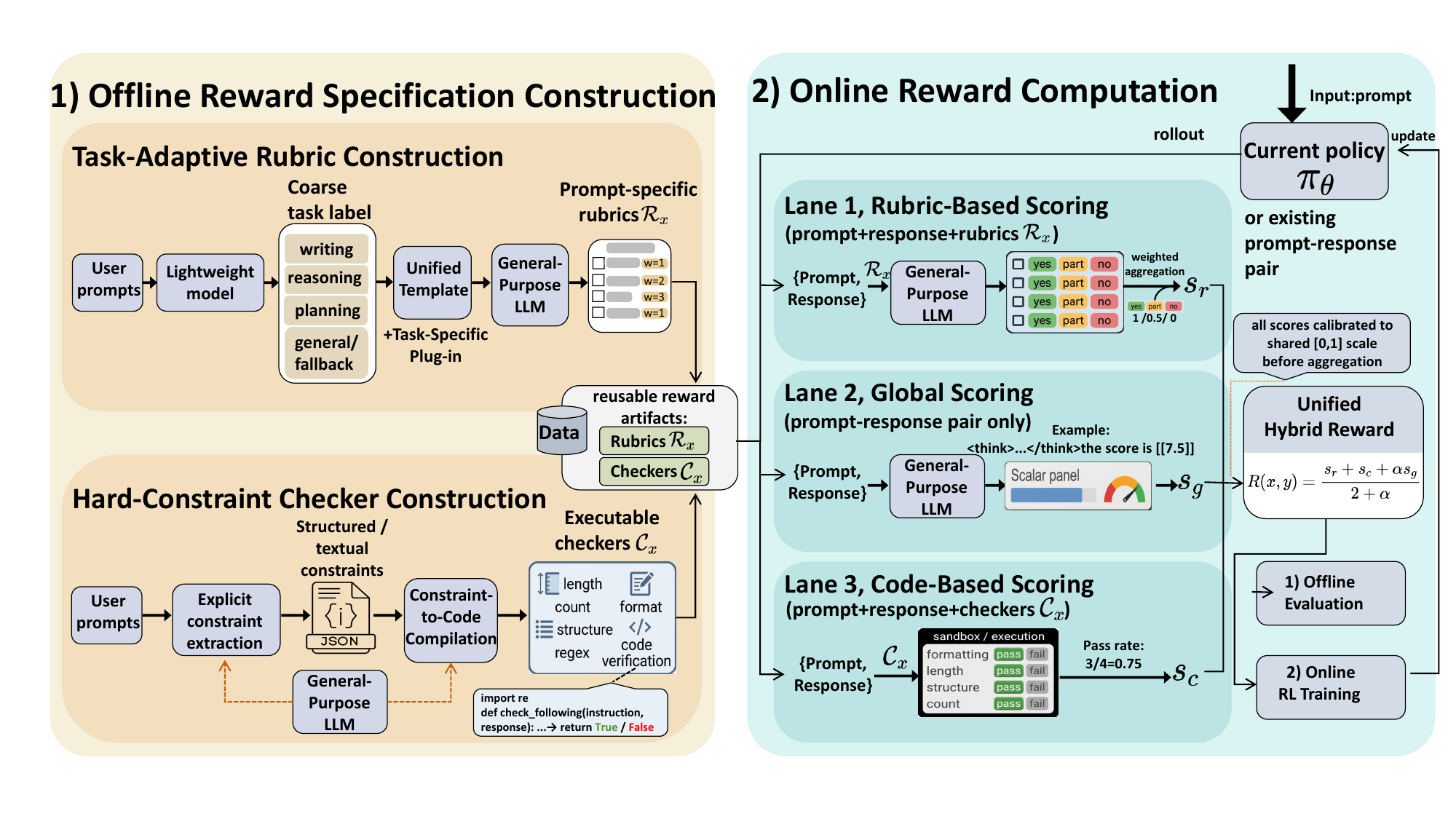}
    \caption{\textbf{Left}: offline reward specification construction builds reusable reward artifacts from prompts alone, including prompt-specific rubrics $\mathcal{R}_x$ and executable hard-constraint checkers $\mathcal{C}_x$. \textbf{Right}: online reward computation combines rubric-based, global, and code-based scoring to produce a unified reward for evaluation and training.}
    \label{fig:fig2}
\end{figure*}

\section{Related Work}

\paragraph{RL for LLM post-training.}
Reinforcement learning has become a widely used component of LLM post-training, ranging from PPO-based RLHF pipelines~\citep{schulman2017ppo,ouyang2022rlhf} to recent RLVR-style methods that exploit verifiable signals in math, code, and other reasoning-heavy domains~\citep{shao2024deepseekmath,guo2025deepseek,yu2026dapo}. 
Preference-based reward models are broadly applicable, but they require preference supervision and can be opaque; verifiable rewards are reliable, but they require automatically checkable outcomes. 
We use Group Sequence Policy Optimization (GSPO)~\citep{zheng2025gspo} for online RL, but our focus is not a new RL algorithm. Instead, we study reward construction for open-ended post-training, where clean verifiers are often unavailable.

\paragraph{Rubric-based evaluation and reward design.}
Rubrics provide an interpretable interface for decomposing open-ended response quality into judgeable criteria~\citep{prometheus}. 
HealthBench uses conversation-specific rubrics to evaluate nuanced medical responses~\citep{arora2025healthbench}, while LMUnit formulates fine-grained natural-language unit tests~\citep{lmunit}. 
Recent work further extends rubrics from evaluation to training-time supervision, including joint rubric generation and judging, rubric-based reward modeling, synthetic rubric construction, rubric-as-reward RL, and rubric quality refinement~\citep{xu2026rubricarm,chaisingthetail,liu2026openrubric,gunjal2025RaR,shen2026RRD}. 
Related reward-modeling and rubric-scaffolded RL methods also study how structured criteria can improve training signals~\citep{wang2026outcomeaccuracyenoughaligning,shao2025drtulureinforcementlearning,zhou2026breakingexplorationbottleneckrubricscaffolded}. 
These works demonstrate the value of rubric-based supervision, but they mainly treat rubrics as the central supervision mechanism, through rubric construction, rubric-conditioned reward modeling, or rubric-guided RL. 
In contrast, we treat rubrics as reusable prompt-level reward artifacts constructed before response scoring, and use them as one component of a pointwise hybrid reward together with global scoring and executable verification.

\paragraph{Prompt-level reward specifications and hybrid rewards.}
Recent work has begun to combine model-based judgments with verifiable signals to improve reward reliability. 
Agentic Reward Modeling augments reward models with verifiable correctness signals, Omni-Thinker combines rule-based verifiable rewards with LLM-as-a-judge preference signals, and VerIF pairs rule-based code verification with LLM-based verification for instruction-following RL~\citep{agenticrm,omini-thinker,peng2025verif}. 
Recent technical reports also suggest the practical relevance of rubric-guided, generative, and programmatic reward signals in large-scale post-training~\citep{kimiteam2025kimik2openagentic,deepseekv4,mimov2}. 
These works motivate hybrid reward construction, but they do not focus on reusable prompt-level reward specifications.

Closest to our work, OpenRS combines adaptive rubrics with verifiable rubric signals for open-ended RL~\citep{jia2026openrs}. 
However, the two methods differ in how reward criteria are specified and reused. 
OpenRS constructs response-pair-conditioned rubrics online for pairwise optimization, whereas our method constructs reusable prompt-level rubrics and executable checkers offline from prompts alone. 
These artifacts are fixed before rollout generation and reused across responses, enabling a pointwise hybrid reward for both offline evaluation and online RL. 
Our reward also includes an independent global score, which is not a replacement for rubric scoring but a complementary signal: criterion-wise rubrics expose local prompt-specific requirements, while global scoring captures response-level coherence, usefulness, and interactions among criteria.
\section{Method}
\subsection{Overview and Problem Formulation}

Figure~\ref{fig:fig2} shows our two-stage reward specification framework for open-ended post-training. Given a prompt $x$, the policy generates a response $y \sim \pi_\theta(\cdot \mid x)$. Our goal is to construct a reward function $R(x,y)$ from prompts alone, without human preference annotations, reference answers, or a separately trained reward model.

For each prompt, we first build a prompt-specific reward specification
\begin{equation}
\mathcal{S}(x)=\{\mathcal{R}_x,\mathcal{C}_x\},
\end{equation}
where $\mathcal{R}_x$ is a task-adaptive rubric and $\mathcal{C}_x$ is a set of executable checkers for explicit hard constraints. We refer to $\mathcal{R}_x$ and $\mathcal{C}_x$ as reward artifacts, which instantiate the specification and are constructed once offline and reused across rollouts. Online, each candidate response is scored by three complementary signals: a rubric score for fine-grained requirement satisfaction, a global score for holistic response quality, and a code-based score for deterministic constraint verification. The resulting hybrid reward provides an RM-style scalar scoring interface for offline evaluation and can also be used directly for online RL.

\subsection{Offline Reward Specification Construction}

We construct the prompt-level reward specification $S(x)$ through two offline branches: task-adaptive rubric generation and hard-constraint checker construction.

\paragraph{Task-adaptive rubric.}
To construct $\mathcal{R}_x$, we first assign the prompt a coarse task label with a lightweight classifier. This label does not score responses directly; it provides a task prior for rubric generation. The classifier is intentionally conservative and falls back to a general category when the prompt intent is uncertain. Conditioned on the task label, we combine a shared rubric-generation template with a task-specific module to produce weighted prompt-specific criteria.

The rubric is designed as a reward basis for rollout supervision rather than a generic evaluation checklist. Each criterion will later be judged independently, assigned a ternary label, and aggregated by weight. We therefore favor criteria that are atomic, low-overlap, independently judgeable, and discriminative, especially those that remain informative among broadly acceptable responses rather than saturating under superficial compliance.

\paragraph{Hard-constraint checkers.}
In parallel, we build executable checkers from the prompt itself, rather than from the generated rubric. This prompt-only branch targets requirements that can be deterministically verified, such as length limits, counts, required or forbidden strings, formatting rules, and surface-level output structure. It first extracts explicit hard constraints into a restricted structured form, and then compiles them into executable code. Semantic or holistic requirements are not converted into checkers and are instead left to model-based scoring. Since rubrics and checkers are constructed once for a fixed prompt set, they can be reused across rollouts and training runs.

\subsection{Online Hybrid Reward Computation}

Given a prompt-response pair $(x,y)$, we compute up to three reward components: a rubric score $s_r(x,y)$, a global score $s_g(x,y)$, and a code-based score $s_c(x,y)$. The key idea is to decompose open-ended quality into heterogeneous supervision signals instead of relying on a single holistic scalar judge. The three components target local requirement satisfaction, overall response quality, and deterministic hard-constraint verification, and are normalized to a shared $[0,1]$ scale before aggregation.

\paragraph{Rubric-based score.}
Let $\mathcal{R}_x=\{(r_i,w_i)\}_{i=1}^{m}$ denote the prompt-specific rubric, where $r_i$ is a criterion and $w_i$ is its weight. A general-purpose language model judges each criterion independently as $\texttt{yes}$, $\texttt{part}$, or $\texttt{no}$, which we map to $1$, $0.5$, and $0$, respectively. Let $v_i$ be the mapped value for $r_i$. The rubric score is
\begin{equation}
s_r(x,y)=\frac{\sum_{i=1}^{m} w_i v_i}{\sum_{i=1}^{m} w_i}.
\end{equation}
This score provides a normalized measure of local requirement satisfaction.

\paragraph{Global score.}
The global score captures response-level quality beyond atomic rubric items.
A general-purpose language model evaluates the prompt-response pair $(x,y)$
and directly returns a holistic raw score $g(x,y)\in[0,10]$.
We normalize it as $s_g(x,y)=\mathrm{clip}(g(x,y)/10,0,1)$ before aggregation,
so that it is on the same bounded $[0,1]$ scale as the rubric-based and
code-based scores and does not dominate the reward due to scale mismatch.
Unlike $s_r$, this score does not depend on prompt-specific offline artifacts
and instead provides a dense response-level quality signal.

\paragraph{Code-based score.}
For explicit hard constraints, let $\mathcal{C}_x=\{c_j\}_{j=1}^{n}$ be the executable checkers constructed offline for prompt $x$. Each checker returns a binary outcome $b_j=c_j(x,y)\in\{0,1\}$. When $n>0$, this score is the checker pass rate:
\begin{equation}
s_c(x,y)=\frac{1}{n}\sum_{j=1}^{n} b_j.
\end{equation}
This component is used only for requirements that admit reliable deterministic verification, such as exact length, counts or structural constraints.

\paragraph{Unified hybrid reward.}
We aggregate the available components into a unified reward. For brevity, all scores below are evaluated on $(x,y)$:
\begin{equation}
R(x,y)=
\begin{cases}
\frac{s_r+s_c+\alpha s_g}{2+\alpha}, & n>0,\\
\frac{s_r+\alpha s_g}{1+\alpha}, & n=0,
\end{cases}
\end{equation}
where $\alpha \ge 0$ controls the contribution of the holistic signal.

\begin{table*}[t]
    \centering
    \small
    \setlength{\tabcolsep}{2.6pt}
    \resizebox{\linewidth}{!}{
    \begin{tabular}{lcccccc>{\columncolor{gray!20}}c|ccc>{\columncolor{gray!20}}c}
    \toprule
    \multirow{2}{*}{Model} & \multicolumn{7}{c}{RewardBench v2} & \multicolumn{4}{c}{RM-Bench} \\
    \cmidrule(lr){2-12}
    & Factuality & PreciseIF & Math & Safety & Focus & Ties & \textbf{Overall} & Easy & Normal & Hard & \textbf{Overall} \\
    \midrule

    \textit{Proprietary LLM Judges} 
    &  &  &  &  &  &  &  &  &  &  & \\
    Gemini-2.5-pro*   & 75.5 & 61.9 & \textbf{89.8} & 88.1 & 80.5 & 81.1 & 79.5 & -- & -- & -- & -- \\
    Gemini-2.5-flash* & 67.4 & 57.5 & 85.2 & 90.9 & 84.1 & 80.9 & 77.7 & -- & -- & -- & -- \\
    GPT-4.1*          & 82.9 & 39.7 & 65.2 & 87.3 & 73.4 & 85.4 & 72.3 & 85.7 & 77.0 & 69.5 & 77.4 \\
    \midrule

    \textit{Trained Reward Models} 
    &  &  &  &  &  &  &  &  &  &  & \\
    Skywork-RM-v2-8B*     & \underline{84.6} & \underline{66.2} & 77.6 & \textbf{96.7} & \textbf{98.4} & 81.2 & \underline{84.1} & \textbf{97.0} & \textbf{95.0} & \underline{86.5} & \textbf{92.8} \\
    LMUnit-Qwen2.5-72B*                 & \textbf{87.2} & 54.4 & 72.7 & 91.3 & \underline{96.8} & \underline{90.1} & 82.1 & -- & -- & -- & -- \\
    Qwen3-Nemo-32B-Gen* & --   & --   & --   & --   & --   & --   & --   & 88.9 & 86.4 & 83.4 & 86.2 \\
    \midrule

    \textit{Our Instantiations} 
        &  &  &  &  &  &  &  &  &  &  & \\
    Qwen3-30B-A3B
        & 61.5 & 34.4 & 70.9 & 84.8 & 74.8 & 71.5 & 66.3
        & 85.1 & 80.5 & 74.3 & 80.0 \\
    \quad w/ Hybrid Reward 
        & 67.7\,\textcolor{Green}{\textbf{(+6.2)}} 
        & 53.8\,\textcolor{Green}{\textbf{(+19.4)}} 
        & 84.7\,\textcolor{Green}{\textbf{(+13.8)}} 
        & 90.4\,\textcolor{Green}{\textbf{(+5.6)}} 
        & 82.8\,\textcolor{Green}{\textbf{(+8.0)}} 
        & 89.1\,\textcolor{Green}{\textbf{(+17.6)}} 
        & 78.1\,\textcolor{Green}{\textbf{(+11.8)}}  
        & 90.6\,\textcolor{Green}{\textbf{(+5.5)}} 
        & 87.8\,\textcolor{Green}{\textbf{(+7.3)}} 
        & 82.1\,\textcolor{Green}{\textbf{(+7.8)}} 
        & 86.8\,\textcolor{Green}{\textbf{(+6.8)}} \\

    Qwen3.5-35B-A3B
        & 76.8 & 61.8 & 84.4 & 90.5 & 77.7 & 88.6 & 80.0
        & 89.6 & 87.9 & 85.1 & 87.5 \\
    \quad w/ Hybrid Reward 
        & 79.5\,\textcolor{Green}{\textbf{(+2.7)}} 
        & \textbf{71.0}\,\textcolor{Green}{\textbf{(+9.2)}} 
        & \underline{88.1}\,\textcolor{Green}{\textbf{(+3.7)}} 
        & \underline{92.2}\,\textcolor{Green}{\textbf{(+1.7)}} 
        & 82.7\,\textcolor{Green}{\textbf{(+5.0)}} 
        & \textbf{97.3}\,\textcolor{Green}{\textbf{(+8.7)}} 
        & \textbf{85.1}\,\textcolor{Green}{\textbf{(+5.1)}} 
        & \underline{94.6}\,\textcolor{Green}{\textbf{(+5.0)}} 
        & \underline{91.7}\,\textcolor{Green}{\textbf{(+3.8)}} 
        & \textbf{87.1}\,\textcolor{Green}{\textbf{(+2.0)}} 
        & \underline{91.1}\,\textcolor{Green}{\textbf{(+3.6)}} \\
    \bottomrule
    \end{tabular}
    }
    \caption{\textbf{Results on RewardBench v2 and RM-Bench}. \textbf{Bold} and \underline{underline} denote the best and second-best results in each column, respectively. The two \textbf{Overall} columns report the average scores over the corresponding sub-metrics in each benchmark. Results marked with * are taken from official benchmark leaderboards, and ``--'' indicates unavailable results. \textbf{Abbreviations:} Skywork-RM-V2-8B denotes Skywork-Reward-V2-Llama-3.1-8B; Qwen3-Nemo-32B-Gen denotes Qwen3-Nemotron-32B-GenRM-Principle.}
    \label{tab:main_exp_rm}
\end{table*}

\subsection{Why Hybrid Reward?}

Open-ended responses exhibit heterogeneous failure modes that are difficult to capture with a single scalar reward. Rubric-based scoring decomposes prompt-specific requirements into fine-grained criteria and provides diagnostic supervision~\citep{jia2026openrs}, but independently scored criteria may miss response-level interactions and trade-offs among strong candidates. Global scoring instead evaluates the response as a whole and provides a dense signal for broad qualities such as relevance, helpfulness, and coherence~\citep{zheng2023judging}, but a single holistic score is less interpretable and can be unreliable for strict instruction-following constraints~\citep{zhou2023instruction,pyatkin2026generalizing}.

Code-based scoring complements these model-based rewards for explicitly verifiable requirements, such as length limits, counts, forbidden strings and surface-level structure. For such constraints, executable checkers provide deterministic supervision. Our hybrid reward therefore assigns local requirement satisfaction to rubric scoring, response-level quality to global scoring, and hard-constraint satisfaction to code-based verification.
\section{Experiments}
We evaluate the proposed framework in two settings: offline RM-style scoring and online RL for open-ended post-training. We report offline results in Section~\ref{sec:offline_eval}, online RL results in Section~\ref{sec:online_rl}, and further analyses in Section~\ref{sec:analysis}.

\begin{table*}[t]
    \centering
    \small
    \setlength{\tabcolsep}{2.4pt}
    \resizebox{\linewidth}{!}{
    \begin{tabular}{lccccc>{\columncolor{gray!20}}c}
    \toprule
    \multirow{2}{*}{Model}
    & IFEval 
    & IFBench 
    & Arena-Hard-v2.0
    & Creative Writing v3
    & WritingBench
    & \textbf{Avg} \\
    \cmidrule(lr){2-7}
    & Pr. (S) & Pr. (S) & Score & Score & Score & Score \\
    \midrule
    DeepSeek-R1-Distill-Qwen-7B 
    & 57.1{\scriptsize\textcolor{black!80}{$\boldsymbol{\pm}$ 0.4}} 
    & 13.4{\scriptsize\textcolor{black!80}{$\boldsymbol{\pm}$ 0.8}} 
    & 2.7{\scriptsize\textcolor{black!80}{$\boldsymbol{\pm}$ 0.3}} 
    & 37.8{\scriptsize\textcolor{black!80}{$\boldsymbol{\pm}$ 1.3}} 
    & 50.4{\scriptsize\textcolor{black!80}{$\boldsymbol{\pm}$ 0.5}} 
    & 32.3 \\
    \quad +RL w/ Hybrid Reward
    & 69.2{\scriptsize\textcolor{black!80}{$\boldsymbol{\pm}$ 0.3}}\,\textcolor{Green}{(+12.1)} 
    & 21.3{\scriptsize\textcolor{black!80}{$\boldsymbol{\pm}$ 0.2}}\,\textcolor{Green}{(+7.9)} 
    & 4.1{\scriptsize\textcolor{black!80}{$\boldsymbol{\pm}$ 0.4}}\,\textcolor{Green}{(+1.4)} 
    & 44.9{\scriptsize\textcolor{black!80}{$\boldsymbol{\pm}$ 1.5}}\,\textcolor{Green}{(+7.1)} 
    & 59.3{\scriptsize\textcolor{black!80}{$\boldsymbol{\pm}$ 0.5}}\,\textcolor{Green}{(+8.9)} 
    & 39.8\,\textcolor{Green}{(+7.5)} \\
    \midrule
    Qwen3-4B 
    & 83.4{\scriptsize\textcolor{black!80}{$\boldsymbol{\pm}$ 0.2}} 
    & 29.4{\scriptsize\textcolor{black!80}{$\boldsymbol{\pm}$ 0.8}} 
    & 14.7{\scriptsize\textcolor{black!80}{$\boldsymbol{\pm}$ 0.7}} 
    & 59.4{\scriptsize\textcolor{black!80}{$\boldsymbol{\pm}$ 1.5}} 
    & 70.0{\scriptsize\textcolor{black!80}{$\boldsymbol{\pm}$ 0.5}} 
    & 51.4 \\
    \quad +RL w/ Hybrid Reward
    & 86.5{\scriptsize\textcolor{black!80}{$\boldsymbol{\pm}$ 0.2}}\,\textcolor{Green}{(+3.1)} 
    & 35.9{\scriptsize\textcolor{black!80}{$\boldsymbol{\pm}$ 0.2}}\,\textcolor{Green}{(+6.5)} 
    & 22.0{\scriptsize\textcolor{black!80}{$\boldsymbol{\pm}$ 0.9}}\,\textcolor{Green}{(+7.3)} 
    & 79.8{\scriptsize\textcolor{black!80}{$\boldsymbol{\pm}$ 1.8}}\,\textcolor{Green}{(+20.4)} 
    & 76.4{\scriptsize\textcolor{black!80}{$\boldsymbol{\pm}$ 0.4}}\,\textcolor{Green}{(+6.4)} 
    & 60.1\,\textcolor{Green}{(+8.7)} \\
    \midrule
    GLM-4.7-Flash 
    & 83.5{\scriptsize\textcolor{black!80}{$\boldsymbol{\pm}$ 1.0}} 
    & 50.8{\scriptsize\textcolor{black!80}{$\boldsymbol{\pm}$ 2.4}} 
    & 28.3{\scriptsize\textcolor{black!80}{$\boldsymbol{\pm}$ 1.1}} 
    & 80.4{\scriptsize\textcolor{black!80}{$\boldsymbol{\pm}$ 1.2}} 
    & 76.3{\scriptsize\textcolor{black!80}{$\boldsymbol{\pm}$ 0.5}} 
    & 63.9 \\
    \quad +RL w/ Hybrid Reward
    & 89.8{\scriptsize\textcolor{black!80}{$\boldsymbol{\pm}$ 0.6}}\,\textcolor{Green}{(+6.3)} 
    & 57.3{\scriptsize\textcolor{black!80}{$\boldsymbol{\pm}$ 0.6}}\,\textcolor{Green}{(+6.5)} 
    & 47.8{\scriptsize\textcolor{black!80}{$\boldsymbol{\pm}$ 1.5}}\,\textcolor{Green}{(+19.5)} 
    & 83.3{\scriptsize\textcolor{black!80}{$\boldsymbol{\pm}$ 1.3}}\,\textcolor{Green}{(+2.9)} 
    & 81.4{\scriptsize\textcolor{black!80}{$\boldsymbol{\pm}$ 0.4}}\,\textcolor{Green}{(+5.1)} 
    & 71.9\,\textcolor{Green}{(+8.0)} \\
    \midrule
    Qwen3-30B-A3B 
    & 85.4{\scriptsize\textcolor{black!80}{$\boldsymbol{\pm}$ 0.1}} 
    & 35.9{\scriptsize\textcolor{black!80}{$\boldsymbol{\pm}$ 1.0}} 
    & 30.6{\scriptsize\textcolor{black!80}{$\boldsymbol{\pm}$ 0.9}} 
    & 77.3{\scriptsize\textcolor{black!80}{$\boldsymbol{\pm}$ 1.3}} 
    & 74.4{\scriptsize\textcolor{black!80}{$\boldsymbol{\pm}$ 0.5}} 
    & 60.7 \\
    \quad +RL w/ Hybrid Reward
    & 87.5{\scriptsize\textcolor{black!80}{$\boldsymbol{\pm}$ 0.4}}\,\textcolor{Green}{(+2.1)} 
    & 39.3{\scriptsize\textcolor{black!80}{$\boldsymbol{\pm}$ 0.7}}\,\textcolor{Green}{(+3.4)} 
    & 38.5{\scriptsize\textcolor{black!80}{$\boldsymbol{\pm}$ 1.0}}\,\textcolor{Green}{(+7.9)} 
    & 82.6{\scriptsize\textcolor{black!80}{$\boldsymbol{\pm}$ 1.4}}\,\textcolor{Green}{(+5.3)} 
    & 79.2{\scriptsize\textcolor{black!80}{$\boldsymbol{\pm}$ 0.4}}\,\textcolor{Green}{(+4.8)} 
    & 65.4\,\textcolor{Green}{(+4.7)} \\
    \bottomrule
    \end{tabular}
    }
    \caption{Experimental results of RL training with our method on DeepSeek-R1-Distill-Qwen-7B, Qwen3-4B, GLM-4.7-Flash, and Qwen3-30B-A3B. $\pm$ denotes sample standard deviation over three runs. We report prompt-level strict scores for IFEval and IFBench. Avg averages the five displayed mean scores.}
    \label{tab:main_exp_rl}
\end{table*}

\subsection{Experimental Setup}
\label{sec:exp_setup}

\paragraph{Reward pipeline instantiation.}
All construction and scoring modules are instantiated with general-purpose language models. 
We use Qwen3-4B~\citep{qwen3} for task-label extraction, the fixed OpenAI API snapshot \texttt{gpt-5-2025-08-07} for prompt-specific rubric generation, Qwen3-30B-A3B for constraint extraction and checker generation, and Qwen3.5-35B-A3B~\citep{qwen3.5} for rubric judging and global scoring.

\paragraph{Policy models and training data.}
For online RL, we evaluate multiple policy backbones, including DeepSeek-R1-Distill-Qwen-7B ~\citep{guo2025deepseek}, Qwen3-4B, GLM-4.7-Flash~\citep{glm4.5}, and Qwen3-30B-A3B. Training uses 13K prompts: 5K from VERINSTRUCT~\citep{peng2025verif}, 5K from DeepWriting-20K~\citep{deepwriting20k}, and 3K synthesized decision-support prompts.

\paragraph{Evaluation benchmarks.}
We evaluate offline reward quality on RewardBench v2~\citep{malik2026rewardbench} and RM-Bench~\citep{rmbench}. For online RL, we report results on IFEval~\citep{zhou2023instruction}, IFBench~\citep{pyatkin2026generalizing}, Arena-Hard-v2.0~\citep{arenahardv2}, Creative Writing v3~\citep{creative-writing-bench-v3}, and WritingBench~\citep{wu2026writingbench}. For our evaluations, unless otherwise specified, we report the average over three runs.

\paragraph{Training details.}
For offline evaluation, we set $\alpha=1$ in Eq.~(4). For online RL, we linearly decay the holistic-score weight during training and train each policy for approximately 800 steps. We use GSPO for policy optimization and disable standard-deviation normalization when computing groupwise advantages~\citep{dr.grpo}. 

For reproducibility, we report the main inference and training settings in
Appendix~\ref{app:implementation_details} and provide the reward-construction and reward-computation prompt
templates in Appendix~\ref{app:prompt-templates}.

\subsection{Offline RM-style Scoring Results}
\label{sec:offline_eval}

We first evaluate the hybrid reward as an RM-style scalar scorer on RewardBench v2 and RM-Bench. These benchmarks test response ranking and preference evaluation, allowing us to assess whether the reward provides a strong standalone quality signal.

As shown in Table~\ref{tab:main_exp_rm}, we compare against proprietary LLM judges, including Gemini-2.5 and GPT-4.1, and trained reward models, including Skywork-Reward-V2-Llama-3.1-8B~\citep{skyworkv2}, LMUnit-Qwen2.5-72B~\citep{lmunit}, and Qwen3-Nemotron-32B-GenRM-Principle~\citep{qwen3nemo}. Our hybrid reward improves both underlying LLM judge instantiations and achieves the top Overall score on RewardBench v2 among the compared systems.

These results isolate the effect of the reward design: the evaluator backbone is unchanged, but combining rubric-based scoring, global scoring, and executable constraint verification consistently improves the scalar reward signal. The gains across both benchmark suites and both judge backbones suggest that hybridization is not tied to a single evaluator configuration.

\subsection{Online RL Results}
\label{sec:online_rl}

We further evaluate whether the proposed hybrid reward can be used as an online training signal. Table~\ref{tab:main_exp_rl} compares each policy backbone before and after RL training on IFEval, IFBench, Arena-Hard-v2.0, Creative Writing v3, and WritingBench.

RL with the hybrid reward improves all evaluated backbones in average score. The gains appear on both instruction-following and open-ended generation benchmarks, suggesting that the reward provides useful supervision for explicit constraint satisfaction as well as broader response quality. The improvements also hold for stronger initial backbones, indicating that the reward is not limited to low-performing policies.
We further compare with Rubicon-Preview~\citep{rubicon}, a released rubric-RL policy trained on the same Qwen3-30B-A3B backbone, in Appendix~\ref{app:rubicon_comparison}.

These results show that the full hybrid reward is an effective optimization signal across policy backbones. To isolate whether the gains come from the proposed reward decomposition rather than from applying RL with an arbitrary scalar reward, Section~\ref{sec:online_component_ablation} provides a controlled online ablation in which the policy, data, rollout configuration, optimization hyperparameters, and training budget are fixed while only the reward formulation is changed.

\begin{table*}[t]
    \centering
    \small
    \setlength{\tabcolsep}{3.0pt}
    \resizebox{\linewidth}{!}{
    \begin{tabular}{llcccccc>{\columncolor{gray!20}}c}
    \toprule
    Group & Variant & Factuality & PreciseIF & Math & Safety & Focus & Ties & Overall \\
    \midrule
    \multicolumn{9}{l}{\textit{Evaluator Backbone: Qwen3.5-35B-A3B}} \\
    \midrule

    \multirow{2}{*}{Global-based}
    & Global only
        & 76.8 & 61.8 & 84.4 & 90.5 & 77.7 & 88.6 & 80.0 \\
    & Global + Code
        & 76.6 & 66.4 & 84.4 & 90.5 & 77.6 & 88.6 & 80.7 \\
    \midrule

    \multirow{3}{*}{Rubric-based}
    & Rubric only
        & 69.4 & 54.4 & 84.8 & 84.1 & 77.5 & 91.8 & 77.0 \\
    & Rubric + Code
        & 69.1 & 59.6 & 84.8 & 84.1 & 77.5 & 91.8 & 77.8 \\
    & Rubric + Global
        & \textbf{79.7} & \underline{68.3} & \textbf{88.1} & \textbf{92.2} & \textbf{82.8} & \textbf{97.3} & \underline{84.7} \\
    \midrule

    Hybrid
    & Rubric + Global + Code
        & \underline{79.5} & \textbf{71.0} & \textbf{88.1} & \textbf{92.2} & \underline{82.7} & \textbf{97.3} & \textbf{85.1} \\
    \midrule
    \end{tabular}
    }
    \caption{Component ablation of the proposed hybrid reward on RewardBench v2 with Qwen3.5-35B-A3B as the evaluator backbone. \textbf{Bold} and \underline{underline} denote the best and second-best results in each column, respectively.}
    \label{tab:main_exp_ablation}
\end{table*}

\section{Analysis}
\label{sec:analysis}

We analyze the hybrid reward through a sequence of ablations and diagnostics. 
Section~\ref{sec:5.1} first studies component complementarity in offline reward evaluation on RewardBench v2. 
Section~\ref{sec:online_component_ablation} then tests whether the same complementarity transfers to online RL. 
We further examine robustness to open-weight rubric extraction and the effect of open-ended RL on reasoning benchmarks.

\subsection{Component Ablation on RewardBench v2}
\label{sec:5.1}

To isolate the effect of reward composition, we fix Qwen3.5-35B-A3B as the evaluator backbone and vary only the reward components, as shown in Table~\ref{tab:main_exp_ablation}.

The full hybrid reward achieves the best overall score. The largest gain comes from combining rubric-based and global scoring: the overall score increases from 77.0 with rubric-only scoring and 80.0 with global-only scoring to 84.7 with both components. This supports our motivation that the two model-based signals are complementary: global scoring captures holistic response quality, while rubric-based scoring evaluates prompt-specific requirements. The same pattern also holds with Qwen3-30B-A3B as the evaluator in Appendix~\ref{app:ablation_30b}, suggesting that the gain is not tied to a single evaluator backbone.

Code-based verification provides a targeted improvement. Adding code improves the overall score from 80.0 to 80.7 for the global reward, from 77.0 to 77.8 for the rubric reward, and from 84.7 to 85.1 for the rubric-global reward. The gain is most consistent on PreciseIF, where code improves the scores from 61.8 to 66.4, 54.4 to 59.6, and 68.3 to 71.0 under the three corresponding settings. This indicates that executable checkers mainly help with explicitly checkable constraints; we further analyze this reliability benefit in Appendix~\ref{app:code_reliability}.

\subsection{Controlled Online Reward Ablation}
\label{sec:online_component_ablation}

We further test whether the online RL gains come from the proposed reward decomposition rather than from applying RL with an arbitrary scalar reward. We conduct a controlled ablation using DeepSeek-R1-Distill-Qwen-7B as the policy model and Qwen3.5-35B-A3B as the reward evaluator. All variants use the same training prompts, rollout configuration, optimization hyperparameters, and training budget; only the reward formulation is changed. We compare global-only, rubric-only, rubric+global, and the full hybrid reward. Since executable checkers only cover explicitly verifiable surface constraints, we isolate their effect by comparing rubric+global with the full hybrid reward instead of using a checker-only reward.

\begin{table}[t]
    \centering
    \small
    \setlength{\tabcolsep}{3pt}
    \resizebox{\columnwidth}{!}{
    \begin{tabular}{lccc>{\columncolor{gray!20}}c}
    \toprule
    Variant
    & IFEval
    & IFBench
    & WritingBench
    & \textbf{Avg.} \\
    \midrule
    Initial
    & 57.1{\scriptsize\textcolor{black!80}{$\boldsymbol{\pm}$ 0.4}}
    & 13.4{\scriptsize\textcolor{black!80}{$\boldsymbol{\pm}$ 0.8}}
    & 50.4{\scriptsize\textcolor{black!80}{$\boldsymbol{\pm}$ 0.5}}
    & 40.3 \\
    \midrule
    Skywork RM
    & 56.1{\scriptsize\textcolor{black!80}{$\boldsymbol{\pm}$ 0.5}}\,\textcolor{Red}{(-1.0)}
    & 17.9{\scriptsize\textcolor{black!80}{$\boldsymbol{\pm}$ 1.0}}\,\textcolor{Green}{(+4.5)}
    & 58.7{\scriptsize\textcolor{black!80}{$\boldsymbol{\pm}$ 0.3}}\,\textcolor{Green}{(+8.3)}
    & 44.2\,\textcolor{Green}{(+3.9)} \\
    \midrule
    G only
    & 64.6{\scriptsize\textcolor{black!80}{$\boldsymbol{\pm}$ 0.9}}\,\textcolor{Green}{(+7.5)}
    & 17.6{\scriptsize\textcolor{black!80}{$\boldsymbol{\pm}$ 0.3}}\,\textcolor{Green}{(+4.2)}
    & 53.2{\scriptsize\textcolor{black!80}{$\boldsymbol{\pm}$ 0.6}}\,\textcolor{Green}{(+2.8)}
    & 45.1\,\textcolor{Green}{(+4.8)} \\
    R only
    & 64.7{\scriptsize\textcolor{black!80}{$\boldsymbol{\pm}$ 1.0}}\,\textcolor{Green}{(+7.6)}
    & 18.9{\scriptsize\textcolor{black!80}{$\boldsymbol{\pm}$ 0.2}}\,\textcolor{Green}{(+5.5)}
    & 57.9{\scriptsize\textcolor{black!80}{$\boldsymbol{\pm}$ 0.5}}\,\textcolor{Green}{(+7.5)}
    & 47.2\,\textcolor{Green}{(+6.9)} \\
    R+G
    & 65.7{\scriptsize\textcolor{black!80}{$\boldsymbol{\pm}$ 0.3}}\,\textcolor{Green}{(+8.6)}
    & 19.3{\scriptsize\textcolor{black!80}{$\boldsymbol{\pm}$ 0.7}}\,\textcolor{Green}{(+5.9)}
    & 58.4{\scriptsize\textcolor{black!80}{$\boldsymbol{\pm}$ 0.4}}\,\textcolor{Green}{(+8.0)}
    & 47.8\,\textcolor{Green}{(+7.5)} \\
    R+G+C
    & \textbf{69.2}{\scriptsize\textcolor{black!80}{$\boldsymbol{\pm}$ 0.3}}\,\textcolor{Green}{(+12.1)}
    & \textbf{21.3}{\scriptsize\textcolor{black!80}{$\boldsymbol{\pm}$ 0.2}}\,\textcolor{Green}{(+7.9)}
    & \textbf{59.3}{\scriptsize\textcolor{black!80}{$\boldsymbol{\pm}$ 0.5}}\,\textcolor{Green}{(+8.9)}
    & \textbf{49.9}\,\textcolor{Green}{(+9.6)} \\
    \bottomrule
    \end{tabular}
    }
    \caption{
    Controlled online RL ablation with fixed training prompts, rollout settings, optimization hyperparameters, and budget.
    R/G/C denote rubric, global, and code-based rewards; Skywork RM uses off-the-shelf Skywork-Reward-v2-Llama-3.1-8B.
    }
    \label{tab:online_component_ablation_analysis}
\end{table}

As shown in Table~\ref{tab:online_component_ablation_analysis}, global-only and rubric-only rewards both improve the initial policy, but combining them yields larger average gains. Adding code-based verification further improves the average score from 47.8 to 49.9, showing that executable verification provides additional value beyond model-based scoring. These results suggest that the complementarity observed in offline reward evaluation also transfers to online RL, and that the gains are not simply due to applying RL with a single reward signal. Appendix~\ref{app:online_ablation_curves} provides the corresponding training curves.

\begin{table*}[t]
    \centering
    \small
    \setlength{\tabcolsep}{3.0pt}
    \resizebox{\linewidth}{!}{
    \begin{tabular}{lcccccc>{\columncolor{gray!20}}c}
    \toprule
    \multirow{2}{*}{Setting} 
    & \multicolumn{7}{c}{RewardBench v2} \\
    \cmidrule(lr){2-8}
    & Factuality & PreciseIF & Math & Safety & Focus & Ties & \textbf{Overall} \\
    \midrule

    \textit{Rubric-only Reward} 
    &  &  &  &  &  &  &  \\
    GPT-5 Extractor
        & 69.4 & 54.4 & 84.8 & 84.1 & 77.5 & 91.8 & 77.0 \\
    Qwen3.5-397B-A17B Extractor
        & 67.2\,\textcolor{Red}{\textbf{(-2.2)}} 
        & 51.3\,\textcolor{Red}{\textbf{(-3.1)}} 
        & 86.2\,\textcolor{Green}{\textbf{(+1.4)}} 
        & 87.5\,\textcolor{Green}{\textbf{(+3.4)}} 
        & 76.7\,\textcolor{Red}{\textbf{(-0.8)}} 
        & 80.9\,\textcolor{Red}{\textbf{(-10.9)}} 
        & 75.0\,\textcolor{Red}{\textbf{(-2.0)}} \\
    \midrule

    \textit{Hybrid Reward} 
    &  &  &  &  &  &  &  \\
    GPT-5 Extractor
        & 79.5 & 71.0 & 88.1 & 92.2 & 82.7 & 97.3 & 85.1 \\
    Qwen3.5-397B-A17B Extractor
        & 78.3\,\textcolor{Red}{\textbf{(-1.2)}} 
        & 68.9\,\textcolor{Red}{\textbf{(-2.1)}} 
        & 89.7\,\textcolor{Green}{\textbf{(+1.6)}} 
        & 93.8\,\textcolor{Green}{\textbf{(+1.6)}} 
        & 82.6\,\textcolor{Red}{\textbf{(-0.1)}} 
        & 93.9\,\textcolor{Red}{\textbf{(-3.4)}} 
        & 84.5\,\textcolor{Red}{\textbf{(-0.6)}} \\
    \bottomrule
    \end{tabular}
    }
    \caption{
    Effect of replacing GPT-5 with an open-weight rubric extractor. 
    Qwen3.5-397B denotes Qwen3.5-397B-A17B. 
    Only the rubric extractor is changed; the rubric judge, global scorer, and code-based verifier are kept fixed. 
    Values in parentheses denote changes relative to the GPT-5 extractor under the same reward setting.
    }
    \label{tab:open_rubric_extractor}
\end{table*}

\subsection{Open-Weight Rubric Extraction: Robustness and Remaining Gap}
\label{sec:open_weight_extractor}

Our main experiments use GPT-5 to generate prompt-specific rubrics, raising the question of whether the framework depends on a proprietary rubric extractor. To examine this, we replace only the rubric extractor with Qwen3.5-397B-A17B, using full-precision inference settings aligned with its official configuration, while keeping the downstream reward-computation pipeline unchanged.

Table~\ref{tab:open_rubric_extractor} shows that open-weight rubric extraction is feasible, although extractor quality still matters. In the rubric-only setting, replacing GPT-5 with Qwen3.5-397B-A17B reduces the Overall score from 77.0 to 75.0. The drop is mainly concentrated in constraint-sensitive categories such as PreciseIF and Ties, while Math and Safety improve slightly.

Importantly, the gap becomes much smaller under the full hybrid reward. The Overall score only decreases from 85.1 to 84.5, suggesting that global scoring and code-based verification can compensate for imperfections in rubric extraction. With the open-weight extractor, adding these complementary signals improves the Overall score from 75.0 to 84.5. These results indicate that the framework does not critically rely on GPT-5 as the rubric extractor, while also highlighting open-weight rubric extraction as a remaining source of variability.

\subsection{Open-Ended RL Largely Preserves Reasoning Benchmark Performance}

A natural concern is that post-training on open-ended tasks may improve instruction following and writing quality while causing regressions on standard reasoning benchmarks~\citep{luo2025empirical}. This concern is particularly relevant in our setting because the RL training mixture is mainly drawn from instruction-following, writing, and scenario-based decision-making prompts, without explicitly adding reasoning-focused math or science data.

\begin{table}[htbp]
    \centering
    \small
    \setlength{\tabcolsep}{2.6pt}
    \resizebox{\columnwidth}{!}{
    \begin{tabular}{lccc}
        \toprule
        \multirow{2}{*}{Model} 
        & \multirow{2}{*}{GSM8K} 
        & GPQA 
        & AIME \\
        & & Diamond & 2024 \\
        \midrule
        Qwen3-30B-A3B 
        & 96.4 
        & 62.8 
        & 80.7 \\
        \quad +RL w/ Hybrid Reward
        & 96.6 \,\textcolor{Green}{(+0.2)} 
        & 62.1 \,\textcolor{Red}{(-0.7)} 
        & 81.3 \,\textcolor{Green}{(+0.6)} \\
        \midrule
        GLM-4.7-Flash 
        & 95.5 
        & 51.6 
        & 90.1 \\
        \quad +RL w/ Hybrid Reward
        & 96.1 \,\textcolor{Green}{(+0.6)} 
        & 58.7 \,\textcolor{Green}{(+7.1)} 
        & 92.7 \,\textcolor{Green}{(+2.6)} \\
        \bottomrule
    \end{tabular}
    }
    \caption{Reasoning benchmark results (\%) before and after RL with our hybrid reward. We evaluate Qwen3-30B-A3B and GLM-4.7-Flash on GSM8K, GPQA Diamond, and AIME 2024.}
    \label{tab:reasoning_preservation}
\end{table}

Table~\ref{tab:reasoning_preservation} shows that we do not observe substantial degradation on the evaluated reasoning benchmarks. For Qwen3-30B-A3B, performance remains broadly stable after RL, with small gains on GSM8K~\citep{gsm8k} and AIME 2024, and a limited drop on GPQA Diamond~\citep{gpqa}. For GLM-4.7-Flash, RL with our hybrid reward improves all three benchmarks, with larger gains on GPQA Diamond and AIME 2024.

Together with the results in Section~\ref{sec:online_rl}, these results suggest that open-ended RL with our hybrid reward improves target open-ended capabilities while largely preserving standard reasoning benchmark performance in our setting.

\subsection{Additional Experiments and Analyses}
\label{sec:additional_experiments}

We provide additional experiments and analyses in the appendix. Appendix~\ref{app:advantage_normalization} validates the removal of standard-deviation normalization in groupwise advantage computation. Appendix~\ref{app:efficiency_async} analyzes the training-efficiency overhead of model-based reward computation and our asynchronous optimization strategy. Appendix~\ref{app:preference_eval} reports a blinded pairwise preference check with order-swapped LLM judges and non-expert human annotators.
\section{Conclusion}

We introduced a prompt-level reward specification framework for open-ended post-training.
Given only prompts, it constructs reusable task-adaptive rubrics and executable hard-constraint checkers, and combines them with an independent global score to form a normalized hybrid reward. 
This design makes prompt-specific requirements explicit while integrating fine-grained semantic judgment, holistic quality assessment, and deterministic constraint verification. 
Experiments show that the reward supports both offline response ranking and online RL, with ablations confirming the complementarity of the three reward signals.
\section*{Limitations}

\paragraph{Dependence on artifact and evaluator quality.}
Our framework removes the need for human preference annotations, reference answers, and separately trained reward models, but it still depends on strong general-purpose LLMs for constructing and evaluating reward artifacts. In our main pipeline, GPT-5 is used for prompt-specific rubric generation, while the remaining artifact-construction and reward-computation components are instantiated with open-weight Qwen models. Although Table~\ref{tab:open_rubric_extractor} shows that replacing GPT-5 with a full-precision open-weight extractor preserves most of the offline hybrid reward performance, artifact quality remains an important source of variability. In particular, the rubric-only setting still exhibits larger drops on constraint-sensitive categories, suggesting that rubric extraction errors are not fully eliminated by using a strong open-weight model.
Moreover, this open-weight replacement experiment is limited to offline reward evaluation. We did not repeat the full online RL pipeline with rubrics regenerated by Qwen3.5-397B-A17B, because deploying the full-weight extractor was substantially more expensive and time-consuming in our setting. Therefore, our results support the feasibility of open-weight offline reward computation, but they do not fully establish the practicality of fully open-weight artifact construction for online RL. Applying the framework to new prompt distributions may still require strong open-weight models, extractor-specific prompt tuning, and additional validation. To mitigate reproducibility concerns, we plan to release the publicly shareable artifacts used in our experiments, including filtered training prompts where permitted, generated rubrics, extracted hard constraints, executable checkers, evaluator prompts, parsing scripts, validation logs, and RL configurations, subject to internal approval and applicable licensing constraints.

\paragraph{Rubric construction and reward aggregation.}
This work focuses on hybrid reward composition rather than on optimizing rubric generation itself. We use a fixed template-and-plug-in pipeline, but do not systematically study alternative extraction strategies, criterion granularities, weighting rules, or refinement procedures. In addition, our reward combines normalized rubric-based, global, and code-based scores with fixed aggregation rules, and does not explicitly resolve severe disagreement among reward sources. We also use a fixed linear decay schedule for the global-score weight during online RL. This schedule is motivated by the intuition that holistic scoring is more useful early in training, while prompt-specific rubric and checker signals become more important as responses improve, but we do not systematically ablate alternative schedules such as constant, stepwise, or learned weighting.

\paragraph{Coverage and safety of executable verification.}
The code-based component is limited to explicit constraints that can be checked with deterministic surface-level rules, such as length, keyword occurrence, formatting, counting, and simple structural requirements. It cannot verify semantic correctness, factuality, reasoning validity, or deeper pragmatic requirements. Although our checker pipeline restricts extraction to surface-checkable constraint types, validates generated code, and uses timeout and failure-handling logic, generated executable checkers may still require sandboxing and auditing before deployment.

\paragraph{Scope of human preference evaluation.}
We include a blinded pairwise preference check with multiple LLM judges and three non-expert human annotators to reduce reliance on a single benchmark-level evaluator. However, this study remains small-scale and is not intended as a statistically powered human evaluation. The human annotators are ordinary users rather than trained expert annotators, and their judgments are used only as an auxiliary sanity check. A larger expert-annotated study would be needed to draw stronger conclusions about fine-grained human preference alignment.

\section*{Ethical Considerations}

Our work uses existing datasets and synthetic prompts for research on open-ended post-training. For datasets such as VERINSTRUCT and DeepWriting-20K, we use only the prompt text and do not use reference responses, solutions, reasoning traces, verifier outputs, or other supervision signals from the original datasets. 
Any supplementary artifact package associated with this work contains only filtered prompt texts and derived reward artifacts; it does not include reference responses, solutions, reasoning traces, verifier outputs, or other supervision signals from the original datasets.
We do not intentionally introduce private user data. The blinded preference check in Appendix~\ref{app:preference_eval} includes three non-expert human annotators and is used only as a diagnostic comparison over model outputs on benchmark prompts. 
These human judgments are not used in reward construction, reward computation, or RL training.

We use existing datasets, benchmarks, and open-weight models only for research purposes and cite their original creators. We follow the publicly available licenses or terms of use associated with these artifacts where specified. Any publicly released reward artifacts associated with this work are intended for research use.

The proposed framework can improve reward construction for instruction following and open-ended generation, but it may also amplify undesirable behavior if unsafe prompts, biased rubrics, or inappropriate reward criteria are used. The code-based component further introduces execution-related risks, so generated checkers should be run with sandboxing, restricted permissions, timeout controls, and auditing before deployment. LLM-based evaluators may also inherit biases, calibration errors, or prompt-injection vulnerabilities from their underlying models; therefore, reward artifacts and trained models should be audited carefully, especially in high-impact domains.

We used AI assistants, including ChatGPT and Claude, to help polish writing and improve clarity. All technical ideas, experiments, analyses, and final claims were checked and are the responsibility of the authors.

\bibliography{custom}

@misc{shao2024deepseekmath,
      title={DeepSeekMath: Pushing the Limits of Mathematical Reasoning in Open Language Models}, 
      author={Zhihong Shao and Peiyi Wang and Qihao Zhu and Runxin Xu and Junxiao Song and Xiao Bi and Haowei Zhang and Mingchuan Zhang and Y. K. Li and Y. Wu and Daya Guo},
      year={2024},
      eprint={2402.03300},
      archivePrefix={arXiv},
      primaryClass={cs.CL},
      url={https://arxiv.org/abs/2402.03300}, 
}

@article{guo2025deepseek,
  title={DeepSeek-R1 incentivizes reasoning in LLMs through reinforcement learning},
  author={Guo, Daya and Yang, Dejian and Zhang, Haowei and Song, Junxiao and Wang, Peiyi and Zhu, Qihao and Xu, Runxin and Zhang, Ruoyu and Ma, Shirong and Bi, Xiao and others},
  journal={Nature},
  volume={645},
  number={8081},
  pages={633--638},
  year={2025},
  publisher={Nature Publishing Group UK London}
}

@inproceedings{zheng2023judging,
 author = {Zheng, Lianmin and Chiang, Wei-Lin and Sheng, Ying and Zhuang, Siyuan and Wu, Zhanghao and Zhuang, Yonghao and Lin, Zi and Li, Zhuohan and Li, Dacheng and Xing, Eric and Zhang, Hao and Gonzalez, Joseph and Stoica, Ion},
 booktitle = {Advances in Neural Information Processing Systems},
 pages = {46595--46623},
 publisher = {Curran Associates, Inc.},
 title = {Judging LLM-as-a-Judge with MT-Bench and Chatbot Arena},
 volume = {36},
 year = {2023}
}

@article{zhou2023instruction,
    title = "Large Language Model Instruction Following: A Survey of Progresses and Challenges",
    author = "Lou, Renze  and
      Zhang, Kai  and
      Yin, Wenpeng",
    journal = "Computational Linguistics",
    volume = "50",
    number = "3",
    month = sep,
    year = "2024",
    address = "Cambridge, MA",
    publisher = "MIT Press",
    url = "https://aclanthology.org/2024.cl-3.7/",
    doi = "10.1162/coli_a_00523",
    pages = "1053--1095",
}

@inproceedings{ouyang2022rlhf,
 author = {Ouyang, Long and Wu, Jeffrey and Jiang, Xu and Almeida, Diogo and Wainwright, Carroll and Mishkin, Pamela and Zhang, Chong and Agarwal, Sandhini and Slama, Katarina and Ray, Alex and Schulman, John and Hilton, Jacob and Kelton, Fraser and Miller, Luke and Simens, Maddie and Askell, Amanda and Welinder, Peter and Christiano, Paul F and Leike, Jan and Lowe, Ryan},
 booktitle = {Advances in Neural Information Processing Systems},
 editor = {S. Koyejo and S. Mohamed and A. Agarwal and D. Belgrave and K. Cho and A. Oh},
 pages = {27730--27744},
 publisher = {Curran Associates, Inc.},
 title = {Training language models to follow instructions with human feedback},
 url = {https://proceedings.neurips.cc/paper_files/paper/2022/file/b1efde53be364a73914f58805a001731-Paper-Conference.pdf},
 volume = {35},
 year = {2022}
}

@misc{arora2025healthbench,
      title={HealthBench: Evaluating Large Language Models Towards Improved Human Health}, 
      author={Rahul K. Arora and Jason Wei and Rebecca Soskin Hicks and Preston Bowman and Joaquin Quiñonero-Candela and Foivos Tsimpourlas and Michael Sharman and Meghan Shah and Andrea Vallone and Alex Beutel and Johannes Heidecke and Karan Singhal},
      year={2025},
      eprint={2505.08775},
      archivePrefix={arXiv},
      primaryClass={cs.CL},
      url={https://arxiv.org/abs/2505.08775}, 
}

@inproceedings{
malik2026rewardbench,
title={RewardBench 2: Advancing Reward Model Evaluation},
author={Saumya Malik and Valentina Pyatkin and Sander Land and Jacob Morrison and Noah A. Smith and Hannaneh Hajishirzi and Nathan Lambert},
booktitle={The Fourteenth International Conference on Learning Representations},
year={2026},
url={https://openreview.net/forum?id=fb0G86Dewb}
}

@inproceedings{
pyatkin2026generalizing,
title={Generalizing Verifiable Instruction Following},
author={Valentina Pyatkin and Saumya Malik and Victoria Graf and Hamish Ivison and Shengyi Huang and Pradeep Dasigi and Nathan Lambert and Hannaneh Hajishirzi},
booktitle={The Thirty-ninth Annual Conference on Neural Information Processing Systems Datasets and Benchmarks Track},
year={2026},
url={https://openreview.net/forum?id=yfYgwjj5F8}
}

@inproceedings{
arenahardv2,
title={From Crowdsourced Data to High-quality Benchmarks: Arena-Hard and Benchbuilder Pipeline},
author={Tianle Li and Wei-Lin Chiang and Evan Frick and Lisa Dunlap and Tianhao Wu and Banghua Zhu and Joseph E. Gonzalez and Ion Stoica},
booktitle={Forty-second International Conference on Machine Learning},
year={2025},
url={https://openreview.net/forum?id=KfTf9vFvSn}
}

@misc{creative-writing-bench-v3,
  author = {Samuel J Paech},
  title = {EQ-Bench Creative Writing Benchmark v3},
  year = {2025},
  publisher = {GitHub},
  journal = {GitHub repository},
  howpublished = {\url{https://github.com/EQ-bench/creative-writing-bench}}
}

@inproceedings{
wu2026writingbench,
title={WritingBench: A Comprehensive Benchmark for Generative Writing},
author={Yuning Wu and Jiahao Mei and Ming Yan and Chenliang Li and Shaopeng Lai and Yuran Ren and Wang Zijia and Ji Zhang and Mengyue Wu and Qin Jin and Fei Huang},
booktitle={The Thirty-ninth Annual Conference on Neural Information Processing Systems Datasets and Benchmarks Track},
year={2026},
url={https://openreview.net/forum?id=Pkskg9drDQ}
}

@misc{schulman2017ppo,
      title={Proximal Policy Optimization Algorithms}, 
      author={John Schulman and Filip Wolski and Prafulla Dhariwal and Alec Radford and Oleg Klimov},
      year={2017},
      eprint={1707.06347},
      archivePrefix={arXiv},
      primaryClass={cs.LG},
      url={https://arxiv.org/abs/1707.06347}, 
}

@misc{zheng2025gspo,
      title={Group Sequence Policy Optimization}, 
      author={Chujie Zheng and Shixuan Liu and Mingze Li and Xiong-Hui Chen and Bowen Yu and Chang Gao and Kai Dang and Yuqiong Liu and Rui Men and An Yang and Jingren Zhou and Junyang Lin},
      year={2025},
      eprint={2507.18071},
      archivePrefix={arXiv},
      primaryClass={cs.LG},
      url={https://arxiv.org/abs/2507.18071}, 
}

@inproceedings{
        gunjal2025RaR,
        title={Rubrics as Rewards: Reinforcement Learning Beyond Verifiable Domains},
        author={Anisha Gunjal and Anthony Wang and Elaine Lau and Vaskar Nath and Yunzhong He and Bing Liu and Sean M. Hendryx},
        booktitle={The Fourteenth International Conference on Learning Representations},
        year={2026},
        url={https://openreview.net/forum?id=c1bTcrDmt4}
}

@misc{shen2026RRD,
      title={Rethinking Rubric Generation for Improving LLM Judge and Reward Modeling for Open-ended Tasks}, 
      author={William F. Shen and Xinchi Qiu and Chenxi Whitehouse and Lisa Alazraki and Shashwat Goel and Francesco Barbieri and Timon Willi and Akhil Mathur and Ilias Leontiadis},
      year={2026},
      eprint={2602.05125},
      archivePrefix={arXiv},
      primaryClass={cs.LG},
      url={https://arxiv.org/abs/2602.05125}, 
}

@inproceedings{agenticrm,
  author       = {Hao Peng and
                  Yunjia Qi and
                  Xiaozhi Wang and
                  Zijun Yao and
                  Bin Xu and
                  Lei Hou and
                  Juanzi Li},
  title        = {Agentic Reward Modeling: Integrating Human Preferences with Verifiable
                  Correctness Signals for Reliable Reward Systems},
  booktitle    = {Proceedings of the 63rd Annual Meeting of the Association for Computational
                  Linguistics (Volume 1: Long Papers), {ACL} 2025, Vienna, Austria,
                  July 27 - August 1, 2025},
  pages        = {15934--15949},
  publisher    = {Association for Computational Linguistics},
  year         = {2025},
  url          = {https://aclanthology.org/2025.acl-long.775/},
}

@misc{omini-thinker,
      title={Omni-Thinker: Scaling Multi-Task RL in LLMs with Hybrid Reward and Task Scheduling}, 
      author={Derek Li and Jiaming Zhou and Leo Maxime Brunswic and Abbas Ghaddar and Qianyi Sun and Liheng Ma and Yu Luo and Dong Li and Mark Coates and Jianye Hao and Yingxue Zhang},
      year={2025},
      eprint={2507.14783},
      archivePrefix={arXiv},
      primaryClass={cs.LG},
      url={https://arxiv.org/abs/2507.14783}, 
}

@misc{jia2026openrs,
      title={Open Rubric System: Scaling Reinforcement Learning with Pairwise Adaptive Rubric}, 
      author={Ruipeng Jia and Yunyi Yang and Yuxin Wu and Yongbo Gai and Siyuan Tao and Mengyu Zhou and Jianhe Lin and Xiaoxi Jiang and Guanjun Jiang},
      year={2026},
      eprint={2602.14069},
      archivePrefix={arXiv},
      primaryClass={cs.CL},
      url={https://arxiv.org/abs/2602.14069}, 
}

@inproceedings{
yu2026dapo,
title={{DAPO}: An Open-Source {LLM} Reinforcement Learning System at Scale},
author={Qiying Yu and Zheng Zhang and Ruofei Zhu and Yufeng Yuan and Xiaochen Zuo and YuYue and Weinan Dai and Tiantian Fan and Gaohong Liu and Juncai Liu and LingJun Liu and Xin Liu and Haibin Lin and Zhiqi Lin and Bole Ma and Guangming Sheng and Yuxuan Tong and Chi Zhang and Mofan Zhang and Ru Zhang and Wang Zhang and Hang Zhu and Jinhua Zhu and Jiaze Chen and Jiangjie Chen and Chengyi Wang and Hongli Yu and Yuxuan Song and Xiangpeng Wei and Hao Zhou and Jingjing Liu and Wei-Ying Ma and Ya-Qin Zhang and Lin Yan and Yonghui Wu and Mingxuan Wang},
booktitle={The Thirty-ninth Annual Conference on Neural Information Processing Systems},
year={2026},
url={https://openreview.net/forum?id=2a36EMSSTp}
}

@inproceedings{
chaisingthetail,
title={Chasing the Tail: Effective Rubric-based Reward Modeling for Large Language Model Post-Training},
author={Junkai Zhang and Zihao Wang and Lin Gui and Swarnashree Mysore Sathyendra and Jaehwan Jeong and Victor Veitch and Wei Wang and Yunzhong He and Bing Liu and Lifeng Jin},
booktitle={The Fourteenth International Conference on Learning Representations},
year={2026},
url={https://openreview.net/forum?id=pBjy4ek2QV}
}

@misc{xu2026rubricarm,
      title={Alternating Reinforcement Learning for Rubric-Based Reward Modeling in Non-Verifiable LLM Post-Training}, 
      author={Ran Xu and Tianci Liu and Zihan Dong and Tony Yu and Ilgee Hong and Carl Yang and Linjun Zhang and Tao Zhao and Haoyu Wang},
      year={2026},
      eprint={2602.01511},
      archivePrefix={arXiv},
      primaryClass={cs.CL},
      url={https://arxiv.org/abs/2602.01511}, 
}

@misc{qwen3,
      title={Qwen3 Technical Report}, 
      author={An Yang and Anfeng Li and Baosong Yang and Beichen Zhang and Binyuan Hui and Bo Zheng and Bowen Yu and Chang Gao and Chengen Huang and Chenxu Lv and Chujie Zheng and Dayiheng Liu and Fan Zhou and Fei Huang and Feng Hu and Hao Ge and Haoran Wei and Huan Lin and Jialong Tang and Jian Yang and Jianhong Tu and Jianwei Zhang and Jianxin Yang and Jiaxi Yang and Jing Zhou and Jingren Zhou and Junyang Lin and Kai Dang and Keqin Bao and Kexin Yang and Le Yu and Lianghao Deng and Mei Li and Mingfeng Xue and Mingze Li and Pei Zhang and Peng Wang and Qin Zhu and Rui Men and Ruize Gao and Shixuan Liu and Shuang Luo and Tianhao Li and Tianyi Tang and Wenbiao Yin and Xingzhang Ren and Xinyu Wang and Xinyu Zhang and Xuancheng Ren and Yang Fan and Yang Su and Yichang Zhang and Yinger Zhang and Yu Wan and Yuqiong Liu and Zekun Wang and Zeyu Cui and Zhenru Zhang and Zhipeng Zhou and Zihan Qiu},
      year={2025},
      eprint={2505.09388},
      archivePrefix={arXiv},
      primaryClass={cs.CL},
      url={https://arxiv.org/abs/2505.09388}, 
}

@misc{glm4.5,
      title={GLM-4.5: Agentic, Reasoning, and Coding (ARC) Foundation Models}, 
      author={{GLM Team} and Aohan Zeng and Xin Lv and Qinkai Zheng and Zhenyu Hou and Bin Chen and Chengxing Xie and Cunxiang Wang and Da Yin and Hao Zeng and Jiajie Zhang and Kedong Wang and Lucen Zhong and Mingdao Liu and Rui Lu and Shulin Cao and Xiaohan Zhang and Xuancheng Huang and Yao Wei and Yean Cheng and Yifan An and Yilin Niu and Yuanhao Wen and Yushi Bai and Zhengxiao Du and Zihan Wang and Zilin Zhu and Bohan Zhang and Bosi Wen and Bowen Wu and Bowen Xu and Can Huang and Casey Zhao and Changpeng Cai and Chao Yu and Chen Li and Chendi Ge and Chenghua Huang and Chenhui Zhang and Chenxi Xu and Chenzheng Zhu and Chuang Li and Congfeng Yin and Daoyan Lin and Dayong Yang and Dazhi Jiang and Ding Ai and Erle Zhu and Fei Wang and Gengzheng Pan and Guo Wang and Hailong Sun and Haitao Li and Haiyang Li and Haiyi Hu and Hanyu Zhang and Hao Peng and Hao Tai and Haoke Zhang and Haoran Wang and Haoyu Yang and He Liu and He Zhao and Hongwei Liu and Hongxi Yan and Huan Liu and Huilong Chen and Ji Li and Jiajing Zhao and Jiamin Ren and Jian Jiao and Jiani Zhao and Jianyang Yan and Jiaqi Wang and Jiayi Gui and Jiayue Zhao and Jie Liu and Jijie Li and Jing Li and Jing Lu and Jingsen Wang and Jingwei Yuan and Jingxuan Li and Jingzhao Du and Jinhua Du and Jinxin Liu and Junkai Zhi and Junli Gao and Ke Wang and Lekang Yang and Liang Xu and Lin Fan and Lindong Wu and Lintao Ding and Lu Wang and Man Zhang and Minghao Li and Minghuan Xu and Mingming Zhao and Mingshu Zhai and Pengfan Du and Qian Dong and Shangde Lei and Shangqing Tu and Shangtong Yang and Shaoyou Lu and Shijie Li and Shuang Li and Shuang-Li and Shuxun Yang and Sibo Yi and Tianshu Yu and Wei Tian and Weihan Wang and Wenbo Yu and Weng Lam Tam and Wenjie Liang and Wentao Liu and Xiao Wang and Xiaohan Jia and Xiaotao Gu and Xiaoying Ling and Xin Wang and Xing Fan and Xingru Pan and Xinyuan Zhang and Xinze Zhang and Xiuqing Fu and Xunkai Zhang and Yabo Xu and Yandong Wu and Yida Lu and Yidong Wang and Yilin Zhou and Yiming Pan and Ying Zhang and Yingli Wang and Yingru Li and Yinpei Su and Yipeng Geng and Yitong Zhu and Yongkun Yang and Yuhang Li and Yuhao Wu and Yujiang Li and Yunan Liu and Yunqing Wang and Yuntao Li and Yuxuan Zhang and Zezhen Liu and Zhen Yang and Zhengda Zhou and Zhongpei Qiao and Zhuoer Feng and Zhuorui Liu and Zichen Zhang and Zihan Wang and Zijun Yao and Zikang Wang and Ziqiang Liu and Ziwei Chai and Zixuan Li and Zuodong Zhao and Wenguang Chen and Jidong Zhai and Bin Xu and Minlie Huang and Hongning Wang and Juanzi Li and Yuxiao Dong and Jie Tang},
      year={2025},
      eprint={2508.06471},
      archivePrefix={arXiv},
      primaryClass={cs.CL},
      url={https://arxiv.org/abs/2508.06471}, 
}

@inproceedings{peng2025verif,
      title={Verif: Verification engineering for reinforcement learning in instruction following},
      author={Peng, Hao and Qi, Yunjia and Wang, Xiaozhi and Xu, Bin and Hou, Lei and Li, Juanzi},
      booktitle={Proceedings of the 2025 Conference on Empirical Methods in Natural Language Processing},
      pages={30312--30327},
      year={2025}
}

@inproceedings{
        deepwriting20k,
        title={Reverse-Engineered Reasoning for Open-Ended Generation},
        author={Haozhe Wang and Haoran Que and Qixin Xu and Minghao Liu and Wangchunshu Zhou and Jiazhan Feng and Wanjun Zhong and Wei Ye and Tong Yang and Wenhao Huang and Ge Zhang and Fangzhen Lin},
        booktitle={The Fourteenth International Conference on Learning Representations},
        year={2026},
        url={https://openreview.net/forum?id=aK9JneKTL8}
}

@inproceedings{
        rmbench,
        title={{RM}-Bench: Benchmarking Reward Models of Language Models with Subtlety and Style},
        author={Yantao Liu and Zijun Yao and Rui Min and Yixin Cao and Lei Hou and Juanzi Li},
        booktitle={The Thirteenth International Conference on Learning Representations},
        year={2025},
        url={https://openreview.net/forum?id=QEHrmQPBdd}
}

@inproceedings{
        skyworkv2,
        title={Skywork-Reward-V2: Scaling Preference Data Curation via Human-{AI} Synergy},
        author={Chris Yuhao Liu and Liang Zeng and Yuzhen Xiao and Jujie He and Jiacai Liu and Chaojie Wang and Rui Yan and Wei Shen and Fuxiang Zhang and Jiacheng Xu and Yang Liu},
        booktitle={The Fourteenth International Conference on Learning Representations},
        year={2026},
        url={https://openreview.net/forum?id=ofgxkMLqic}
}

@misc{lmunit,
      title={LMUnit: Fine-grained Evaluation with Natural Language Unit Tests}, 
      author={Jon Saad-Falcon and Rajan Vivek and William Berrios and Nandita Shankar Naik and Matija Franklin and Bertie Vidgen and Amanpreet Singh and Douwe Kiela and Shikib Mehri},
      year={2026},
      eprint={2412.13091},
      archivePrefix={arXiv},
      primaryClass={cs.CL},
      url={https://arxiv.org/abs/2412.13091}, 
}

@inproceedings{
        qwen3nemo,
        title={{RLBFF}: Binary Flexible Feedback to bridge between Human Feedback \& Verifiable Rewards},
        author={Zhilin Wang and Jiaqi Zeng and Olivier Delalleau and Ellie Evans and Daniel Egert and Hoo-Chang Shin and Felipe Soares and Yi Dong and Oleksii Kuchaiev},
        booktitle={The Fourteenth International Conference on Learning Representations},
        year={2026},
        url={https://openreview.net/forum?id=P3R3S6S5Km}
}

@misc{gsm8k,
      title={Training Verifiers to Solve Math Word Problems}, 
      author={Karl Cobbe and Vineet Kosaraju and Mohammad Bavarian and Mark Chen and Heewoo Jun and Lukasz Kaiser and Matthias Plappert and Jerry Tworek and Jacob Hilton and Reiichiro Nakano and Christopher Hesse and John Schulman},
      year={2021},
      eprint={2110.14168},
      archivePrefix={arXiv},
      primaryClass={cs.LG},
      url={https://arxiv.org/abs/2110.14168}, 
}

@inproceedings{
gpqa,
title={{GPQA}: A Graduate-Level Google-Proof Q\&A Benchmark},
author={David Rein and Betty Li Hou and Asa Cooper Stickland and Jackson Petty and Richard Yuanzhe Pang and Julien Dirani and Julian Michael and Samuel R. Bowman},
booktitle={First Conference on Language Modeling},
year={2024},
url={https://openreview.net/forum?id=Ti67584b98}
}

@article{luo2025empirical,
  title={An empirical study of catastrophic forgetting in large language models during continual fine-tuning},
  author={Luo, Yun and Yang, Zhen and Meng, Fandong and Li, Yafu and Zhou, Jie and Zhang, Yue},
  journal={IEEE Transactions on Audio, Speech and Language Processing},
  year={2025},
  publisher={IEEE}
}

@inproceedings{
dr.grpo,
title={Understanding R1-Zero-Like Training: A Critical Perspective},
author={Zichen Liu and Changyu Chen and Wenjun Li and Penghui Qi and Tianyu Pang and Chao Du and Wee Sun Lee and Min Lin},
booktitle={Second Conference on Language Modeling},
year={2025},
url={https://openreview.net/forum?id=5PAF7PAY2Y}
}

@misc{qwen3.5,
    title  = {{Qwen3.5}: Towards Native Multimodal Agents},
    author = {{Qwen Team}},
    month  = {February},
    year   = {2026},
    url    = {https://qwen.ai/blog?id=qwen3.5}
}

@misc{liu2026openrubric,
      title={OpenRubrics: Towards Scalable Synthetic Rubric Generation for Reward Modeling and LLM Alignment}, 
      author={Tianci Liu and Ran Xu and Tony Yu and Ilgee Hong and Carl Yang and Tuo Zhao and Haoyu Wang},
      year={2026},
      eprint={2510.07743},
      archivePrefix={arXiv},
      primaryClass={cs.CL},
      url={https://arxiv.org/abs/2510.07743}, 
}

@article{christiano2017deep,
  title={Deep reinforcement learning from human preferences},
  author={Christiano, Paul F and Leike, Jan and Brown, Tom and Martic, Miljan and Legg, Shane and Amodei, Dario},
  journal={Advances in neural information processing systems},
  volume={30},
  year={2017}
}

@misc{deepseekv4,
      title={DeepSeek-V4: Towards Highly Efficient Million-Token Context Intelligence},
      author={DeepSeek-AI},
      year={2026},
}

@misc{mimov2,
      title={MiMo-V2-Flash Technical Report}, 
      author={LLM-Core Xiaomi},
      year={2026},
      eprint={2601.02780},
      archivePrefix={arXiv},
      primaryClass={cs.CL},
      url={https://arxiv.org/abs/2601.02780}, 
}

@inproceedings{
flask,
title={{FLASK}: Fine-grained Language Model Evaluation based on Alignment Skill Sets},
author={Seonghyeon Ye and Doyoung Kim and Sungdong Kim and Hyeonbin Hwang and Seungone Kim and Yongrae Jo and James Thorne and Juho Kim and Minjoon Seo},
booktitle={ICLR 2024 Workshop on Large Language Model (LLM) Agents},
year={2024},
url={https://openreview.net/forum?id=3OfPKwAqPf}
}

@misc{gu2025surveyllmasajudge,
      title={A Survey on LLM-as-a-Judge}, 
      author={Jiawei Gu and Xuhui Jiang and Zhichao Shi and Hexiang Tan and Xuehao Zhai and Chengjin Xu and Wei Li and Yinghan Shen and Shengjie Ma and Honghao Liu and Saizhuo Wang and Kun Zhang and Yuanzhuo Wang and Wen Gao and Lionel Ni and Jian Guo},
      year={2025},
      eprint={2411.15594},
      archivePrefix={arXiv},
      primaryClass={cs.CL},
      url={https://arxiv.org/abs/2411.15594}, 
}

@inproceedings{followbench,
    title = "{F}ollow{B}ench: A Multi-level Fine-grained Constraints Following Benchmark for Large Language Models",
    author = "Jiang, Yuxin  and
      Wang, Yufei  and
      Zeng, Xingshan  and
      Zhong, Wanjun  and
      Li, Liangyou  and
      Mi, Fei  and
      Shang, Lifeng  and
      Jiang, Xin  and
      Liu, Qun  and
      Wang, Wei",
    editor = "Ku, Lun-Wei  and
      Martins, Andre  and
      Srikumar, Vivek",
    booktitle = "Proceedings of the 62nd Annual Meeting of the Association for Computational Linguistics (Volume 1: Long Papers)",
    month = aug,
    year = "2024",
    address = "Bangkok, Thailand",
    publisher = "Association for Computational Linguistics",
    url = "https://aclanthology.org/2024.acl-long.257/",
    doi = "10.18653/v1/2024.acl-long.257",
    pages = "4667--4688",
}

@inproceedings{prometheus,
 author = {Kim, Seungone and Shin, Jay and cho, yejin and Jang, Joel and Longpre, Shayne and Lee, Hwaran and Yun, Sangdoo and Shin, Ryan, S and Kim, Sungdong and Thorne, James and Seo, Minjoon},
 booktitle = {International Conference on Learning Representations},
 editor = {B. Kim and Y. Yue and S. Chaudhuri and K. Fragkiadaki and M. Khan and Y. Sun},
 pages = {29927--29962},
 title = {Prometheus: Inducing Fine-Grained Evaluation Capability in Language Models},
 url = {https://proceedings.iclr.cc/paper_files/paper/2024/file/803485352e61e3ebf41221e4776c9fd4-Paper-Conference.pdf},
 volume = {2024},
 year = {2024}
}

@misc{wang2026outcomeaccuracyenoughaligning,
      title={Outcome Accuracy is Not Enough: Aligning the Reasoning Process of Reward Models}, 
      author={Binghai Wang and Yantao Liu and Yuxuan Liu and Tianyi Tang and Shenzhi Wang and Chang Gao and Chujie Zheng and Yichang Zhang and Le Yu and Shixuan Liu and Tao Gui and Qi Zhang and Xuanjing Huang and Bowen Yu and Fei Huang and Junyang Lin},
      year={2026},
      eprint={2602.04649},
      archivePrefix={arXiv},
      primaryClass={cs.CL},
      url={https://arxiv.org/abs/2602.04649}, 
}

@misc{shao2025drtulureinforcementlearning,
      title={DR Tulu: Reinforcement Learning with Evolving Rubrics for Deep Research}, 
      author={Rulin Shao and Akari Asai and Shannon Zejiang Shen and Hamish Ivison and Varsha Kishore and Jingming Zhuo and Xinran Zhao and Molly Park and Samuel G. Finlayson and David Sontag and Tyler Murray and Sewon Min and Pradeep Dasigi and Luca Soldaini and Faeze Brahman and Wen-tau Yih and Tongshuang Wu and Luke Zettlemoyer and Yoon Kim and Hannaneh Hajishirzi and Pang Wei Koh},
      year={2025},
      eprint={2511.19399},
      archivePrefix={arXiv},
      primaryClass={cs.CL},
      url={https://arxiv.org/abs/2511.19399}, 
}

@misc{zhou2026breakingexplorationbottleneckrubricscaffolded,
      title={Breaking the Exploration Bottleneck: Rubric-Scaffolded Reinforcement Learning for General LLM Reasoning}, 
      author={Yang Zhou and Sunzhu Li and Shunyu Liu and Wenkai Fang and Kongcheng Zhang and Jiale Zhao and Jingwen Yang and Yihe Zhou and Jianwei Lv and Tongya Zheng and Hengtong Lu and Wei Chen and Yan Xie and Mingli Song},
      year={2026},
      eprint={2508.16949},
      archivePrefix={arXiv},
      primaryClass={cs.LG},
      url={https://arxiv.org/abs/2508.16949}, 
}

@misc{kimiteam2025kimik2openagentic,
      title={Kimi K2: Open Agentic Intelligence}, 
      author={Kimi Team and Yifan Bai and Yiping Bao and Guanduo Chen and Jiahao Chen and Ningxin Chen and Ruijue Chen and Yanru Chen and Yuankun Chen and Yutian Chen and Zhuofu Chen and Jialei Cui and Hao Ding and Mengnan Dong and Angang Du and Chenzhuang Du and Dikang Du and Yulun Du and Yu Fan and Yichen Feng and Kelin Fu and Bofei Gao and Hongcheng Gao and Peizhong Gao and Tong Gao and Xinran Gu and Longyu Guan and Haiqing Guo and Jianhang Guo and Hao Hu and Xiaoru Hao and Tianhong He and Weiran He and Wenyang He and Chao Hong and Yangyang Hu and Zhenxing Hu and Weixiao Huang and Zhiqi Huang and Zihao Huang and Tao Jiang and Zhejun Jiang and Xinyi Jin and Yongsheng Kang and Guokun Lai and Cheng Li and Fang Li and Haoyang Li and Ming Li and Wentao Li and Yanhao Li and Yiwei Li and Zhaowei Li and Zheming Li and Hongzhan Lin and Xiaohan Lin and Zongyu Lin and Chengyin Liu and Chenyu Liu and Hongzhang Liu and Jingyuan Liu and Junqi Liu and Liang Liu and Shaowei Liu and T. Y. Liu and Tianwei Liu and Weizhou Liu and Yangyang Liu and Yibo Liu and Yiping Liu and Yue Liu and Zhengying Liu and Enzhe Lu and Lijun Lu and Shengling Ma and Xinyu Ma and Yingwei Ma and Shaoguang Mao and Jie Mei and Xin Men and Yibo Miao and Siyuan Pan and Yebo Peng and Ruoyu Qin and Bowen Qu and Zeyu Shang and Lidong Shi and Shengyuan Shi and Feifan Song and Jianlin Su and Zhengyuan Su and Xinjie Sun and Flood Sung and Heyi Tang and Jiawen Tao and Qifeng Teng and Chensi Wang and Dinglu Wang and Feng Wang and Haiming Wang and Jianzhou Wang and Jiaxing Wang and Jinhong Wang and Shengjie Wang and Shuyi Wang and Yao Wang and Yejie Wang and Yiqin Wang and Yuxin Wang and Yuzhi Wang and Zhaoji Wang and Zhengtao Wang and Zhexu Wang and Chu Wei and Qianqian Wei and Wenhao Wu and Xingzhe Wu and Yuxin Wu and Chenjun Xiao and Xiaotong Xie and Weimin Xiong and Boyu Xu and Jing Xu and Jinjing Xu and L. H. Xu and Lin Xu and Suting Xu and Weixin Xu and Xinran Xu and Yangchuan Xu and Ziyao Xu and Junjie Yan and Yuzi Yan and Xiaofei Yang and Ying Yang and Zhen Yang and Zhilin Yang and Zonghan Yang and Haotian Yao and Xingcheng Yao and Wenjie Ye and Zhuorui Ye and Bohong Yin and Longhui Yu and Enming Yuan and Hongbang Yuan and Mengjie Yuan and Haobing Zhan and Dehao Zhang and Hao Zhang and Wanlu Zhang and Xiaobin Zhang and Yangkun Zhang and Yizhi Zhang and Yongting Zhang and Yu Zhang and Yutao Zhang and Yutong Zhang and Zheng Zhang and Haotian Zhao and Yikai Zhao and Huabin Zheng and Shaojie Zheng and Jianren Zhou and Xinyu Zhou and Zaida Zhou and Zhen Zhu and Weiyu Zhuang and Xinxing Zu},
      year={2025},
      eprint={2507.20534},
      archivePrefix={arXiv},
      primaryClass={cs.LG},
      url={https://arxiv.org/abs/2507.20534}, 
}

@misc{rubicon,
      title={Reinforcement Learning with Rubric Anchors}, 
      author={Zenan Huang and Yihong Zhuang and Guoshan Lu and Zeyu Qin and Haokai Xu and Tianyu Zhao and Ru Peng and Jiaqi Hu and Zhanming Shen and Xiaomeng Hu and Xijun Gu and Peiyi Tu and Jiaxin Liu and Wenyu Chen and Yuzhuo Fu and Zhiting Fan and Yanmei Gu and Yuanyuan Wang and Zhengkai Yang and Jianguo Li and Junbo Zhao},
      year={2025},
      eprint={2508.12790},
      archivePrefix={arXiv},
      primaryClass={cs.AI},
      url={https://arxiv.org/abs/2508.12790}, 
}
\newpage
\clearpage
\appendix
\section*{Appendices}

\section{Implementation Details and Hyperparameters}
\label{app:implementation_details}
To facilitate reproducibility, we provide the main implementation details and hyperparameters used in our experiments. Appendix~\ref{app:pipeline_inference_settings} describes the inference settings for models used inside the reward pipeline, and Appendix~\ref{app:online_rl_training} reports the online RL training setup and optimization hyperparameters.

\subsection{Inference Settings for Pipeline Models}
\label{app:pipeline_inference_settings}

This subsection reports the inference settings for the models used inside our reward pipeline, including offline reward construction and online reward computation.

For offline reward construction, all Qwen3-series models are run in thinking mode. For GPT-5, which is used for prompt-specific rubric generation, we use the default decoding configuration with temperature $1.0$. For online reward computation, the rubric judge and global scorer are instantiated with Qwen3-series or Qwen3.5-series models. The model-specific inference settings are summarized in Table~\ref{tab:pipeline_inference_settings}.

\begin{table}[htbp]
    \centering
    \small
    \resizebox{\columnwidth}{!}{
    \setlength{\tabcolsep}{2.6pt}
    \begin{tabular}{llcccc}
    \toprule
    Model & Mode & Temp. & Top-$p$ & Top-$k$ & Pres. Penalty \\
    \midrule
    Qwen3 & Thinking & 0.6 & 0.95 & 20 & -- \\
    Qwen3 & Non-thinking & 0.7 & 0.95 & 20 & -- \\
    Qwen3.5 & Thinking & 1.0 & 0.95 & 20 & 1.5 \\
    Qwen3.5 & Non-thinking & 0.8 & 0.8 & 20 & 1.5 \\
    GPT-5 & Default & 1.0 & -- & -- & -- \\
    \bottomrule
    \end{tabular}
    }
    \caption{Inference settings for models used inside the reward pipeline.}
    \label{tab:pipeline_inference_settings}
\end{table}

During rubric evaluation, we disable thinking mode for criteria with weights $1$ or $2$ to reduce inference cost. In preliminary trials, disabling thinking for these low-weight criteria does not noticeably affect reward quality. For all other rubric criteria and for global scoring, we enable thinking mode.

\subsection{Online RL Training Setup}
\label{app:online_rl_training}

We conduct online RL experiments using both verl and slime. Since slime provides better training efficiency for MoE policy models in our setting, the main online RL results reported in this paper are obtained with slime unless otherwise specified. The training backend uses Ray for distributed execution, Megatron-style model training, and SGLang for rollout generation.

For policy optimization, we use GSPO. At each rollout step, the current policy samples multiple responses for each prompt, and each prompt-response pair is scored by the proposed hybrid reward. In online RL, we linearly decay the holistic-score weight in Eq.~(4). At training step $t$, we set
\[
\alpha_t = \max\left(0, 1 - \frac{t}{T_{\mathrm{decay}}}\right),
\]
where $T_{\mathrm{decay}}=800$ in our main experiments. This schedule reflects the changing role of the reward components during training. Early in training, rollout responses often differ substantially in overall helpfulness, relevance, and coherence, so the global score provides a useful dense signal. As training progresses, responses within the same prompt group become closer in holistic quality, making the global score less discriminative for fine-grained improvements. We therefore gradually shift the reward toward rubric-based and code-based supervision, which more directly targets prompt-specific requirements and explicit constraints. We use a linear schedule as a simple monotonic transition rather than tuning a more complex decay function. If training continues beyond $T_{\mathrm{decay}}$, we keep $\alpha_t=0$.

For each prompt group, we compute groupwise advantages by subtracting the group mean reward, but we disable the standard-deviation normalization term. Specifically, for a group of aggregated hybrid rewards $\{r_i\}_{i=1}^{G}$, we compute
\[
A_i = 6 \cdot \left(r_i - \frac{1}{G}\sum_{j=1}^{G} r_j\right).
\]
We do not divide by the group standard deviation. This avoids unstable rescaling when the reward variance within a group is small, while the fixed multiplier keeps the advantage magnitude comparable to the normalized setting used in standard group-based RL.

The main online RL hyperparameters are summarized in Table~\ref{tab:online_rl_hyperparameters}.

\begin{table}[htbp]
    \centering
    \small
    \resizebox{\columnwidth}{!}{
    \setlength{\tabcolsep}{3pt}
    \begin{tabular}{lc}
    \toprule
    Hyperparameter & Value \\
    \midrule
    RL algorithm / estimator & GSPO \\
    Rollout batch size & 32 \\
    Responses per prompt & 16 \\
    Global batch size & 512 \\
    Optimization mini-batch size & 32 \\
    Rollout temperature & 1.0 \\
    Maximum prompt length & 2048 \\
    Maximum response length & 8192 \\
    Optimizer & Adam \\
    Learning rate & $1\times10^{-6}$ \\
    Learning-rate schedule & Constant \\
    Weight decay & 0.1 \\
    Adam $\beta_1$ / $\beta_2$ & 0.9 / 0.98 \\
    KL coefficient & 0.001 \\
    KL loss type & Low-variance KL \\
    Entropy coefficient & 0 \\
    Clip lower / upper & $3\times10^{-4}$ / $4\times10^{-4}$ \\
    Clip constant & 10 \\
    Advantage scale & 6 \\
    Std normalization & Disabled \\
    \bottomrule
    \end{tabular}
    }
    \caption{Main hyperparameters used for online RL training.}
    \label{tab:online_rl_hyperparameters}
\end{table}

For distributed training, we mainly use a 2-node setup with 8 H800 GPUs per node, for a total of 16 H800 GPUs. For MoE policy models, we use tensor model parallel size $2$, pipeline model parallel size $1$, context parallel size $1$, expert model parallel size $8$, and expert tensor parallel size $1$. Sequence parallelism and dynamic batch sizing are enabled. To reduce memory usage, we use full activation recomputation with uniform recomputation and set the maximum number of tokens per GPU to $8192$.

For online reward computation, we deploy the reward-model service on the same scale, using 2 nodes with 8 H800 GPUs per node, for a total of 16 H800 GPUs. The reward-model service is used to compute the rubric-based and global scores during training, while executable checkers are evaluated deterministically.

For rollout serving, we use SGLang with 8 GPUs per rollout engine. We enable data-parallel attention and data-parallel LM head, set the SGLang data-parallel size to $8$, and set the static memory fraction to $0.85$. The maximum number of running requests is set to $128$. During training, rollout generation and reward computation are executed asynchronously when possible, so that the latency of model-based reward computation can be partially hidden by rollout generation.

We report approximate local policy-training compute for the four main online RL experiments. GPU-hours are estimated as the number of allocated GPUs multiplied by wall-clock training time. The DeepSeek-R1-Distill-Qwen-7B, Qwen3-4B, Qwen3-30B-A3B, and GLM-4.7-Flash runs used approximately 672, 768, 1,584, and 1,728 H800 GPU-hours, respectively. Model-based reward computation was served by a shared H800 GPU service and is therefore not uniquely attributed to each individual run. 

\subsection{Evaluation Protocols and Settings}
\label{app:evaluation_settings}

For all evaluation benchmarks, we enable thinking mode during inference. Before answer extraction or scoring, we strip the thinking content using a fixed parser and evaluate only the final response. This ensures that benchmark metrics and external judges are applied to the actual user-facing answer rather than the model's hidden reasoning trace. When reporting repeated-run results, $\pm$ denotes the sample standard deviation across runs.

\begin{table}[htbp]
    \centering
    \small
    \resizebox{\columnwidth}{!}{
    \setlength{\tabcolsep}{2.2pt}
    \begin{tabular}{lccccc}
    \toprule
    Benchmark & Temp. & Top-$p$ & Top-$k$ & Min-$p$ & Max Length \\
    \midrule
    IFEval & 0.6 & 0.95 & 20 & -- & 16K \\
    IFBench & 0.6 & 0.95 & 20 & -- & 16K \\
    GSM8K & 0.6 & 0.95 & 20 & -- & 16K \\
    GPQA-Diamond & 0.6 & 0.95 & 20 & -- & 16K \\
    AIME 2024 & 0.6 & 0.95 & 20 & -- & 32K/128K \\
    Arena-Hard v2.0 & 0.6 & -- & -- & -- & 32000 \\
    Creative Writing v3 & 0.7 & -- & -- & 0.1 & 4000 \\
    WritingBench & 0.8 & 0.95 & 20 & -- & 16000 \\
    \bottomrule
    \end{tabular}
    }
    \caption{Generation settings for evaluation benchmarks. For all benchmarks, thinking content is removed before answer extraction or scoring.}
    \label{tab:evaluation_generation_settings}
\end{table}

For IFEval and IFBench, we report prompt-level strict accuracy. For GSM8K, GPQA-Diamond, and AIME 2024, we follow the official answer extraction and evaluation protocols. For benchmarks that require external model-based judging, including Arena-Hard v2.0, Creative Writing v3, and WritingBench, we use GPT-4.1 as the judge and follow the corresponding official evaluation protocols and generation settings.

For IFEval, IFBench, GSM8K, and GPQA-Diamond, we use a unified maximum response length of $16$K tokens. For AIME 2024, we use the same decoding configuration but set model-specific maximum response lengths: $32$K tokens for Qwen3-30B-A3B and $128$K tokens for GLM-4.7-Flash. The benchmark-specific generation settings are summarized in Table~\ref{tab:evaluation_generation_settings}.

\subsection{Checker Validation and Failure Handling}

To reduce unsafe-code and prompt-injection risks, our checker construction uses
a two-stage pipeline rather than directly translating arbitrary user prompts
into code. The first stage extracts only structured, surface-checkable
constraints from a fixed set of allowed types, including word or character
counts, paragraph or sentence counts, keyword inclusion or exclusion, response
language, required starting or ending text, list or output format, and
punctuation rules. Semantic, factual, stylistic, or open-ended requirements are
not converted into executable checkers and are instead left to model-based
scoring. The second stage compiles only these structured constraints into
Python checkers. This separation limits the code generator to deterministic
string matching, regular expressions, and counting logic, reducing the risk that
prompt-injection-style instructions are translated into arbitrary executable
behavior.

To improve the robustness of executable checkers, we validate each generated
checker before using it for reward computation. After constraint-to-code
compilation, we execute the generated checker once on a test prompt-response
input to ensure that it can be parsed and run successfully. If the checker fails
to compile, raises an exception, or exceeds the time limit, we ask the LLM to
regenerate the code, with at most three regeneration attempts. Checkers that
still fail after the maximum number of attempts are discarded and treated as
unavailable.

During online reward computation, checker execution is also guarded by timeout
and retry logic. If a checker raises an exception or times out, we retry the
execution up to the maximum number of attempts. If it still fails, we
conservatively return a failed outcome for that checker. When no valid checker
remains for a prompt, the code-based reward component is omitted and the
remaining reward components are renormalized.

\begin{table*}[t]
    \centering
    \small
    \setlength{\tabcolsep}{3.0pt}
    \resizebox{\linewidth}{!}{
    \begin{tabular}{llcccccc>{\columncolor{gray!20}}c}
    \toprule
    Group & Variant & Factuality & PreciseIF & Math & Safety & Focus & Ties & Overall \\
    \midrule
    \multicolumn{9}{l}{\textit{Evaluator Backbone: Qwen3-30B-A3B}} \\
    \midrule

    \multirow{2}{*}{Global-based}
    & Global only
        & 61.5 & 34.4 & 70.9 & \underline{84.8} & 74.8 & 71.5 & 66.3 \\
    & Global + Code
        & 60.9 & 46.6 & 70.9 & \underline{84.8} & 74.8 & 71.5 & 68.3 \\
    \midrule

    \multirow{3}{*}{Rubric-based}
    & Rubric only
        & 63.4 & 41.7 & \underline{82.5} & 82.8 & \underline{78.1} & \underline{84.9} & 72.2 \\
    & Rubric + Code
        & 62.6 & \underline{51.8} & \underline{82.5} & 82.8 & \underline{78.1} & \underline{84.9} & 73.8 \\
    & Rubric + Global
        & \textbf{67.9} & 44.2 & \textbf{84.7} & \textbf{90.4} & \textbf{82.8} & \textbf{89.1} & \underline{76.5} \\
    \midrule

    Hybrid
    & Rubric + Global + Code
        & \underline{67.7} & \textbf{53.8} & \textbf{84.7} & \textbf{90.4} & \textbf{82.8} & \textbf{89.1} & \textbf{78.1} \\
    \bottomrule
    \end{tabular}
    }
    \caption{Component ablation of the proposed hybrid reward on RewardBench v2 with Qwen3-30B-A3B as the evaluator backbone. \textbf{Bold} and \underline{underline} denote the best and second-best results in each column, respectively.}
    \label{tab:ablation_qwen3_30b}
\end{table*}

\subsection{Reproducibility and Released Artifacts}
\label{app:released_artifacts}

To facilitate artifact-level reproducibility, we plan to provide a supplementary artifact package containing the fixed data and reward artifacts used to support the experiments reported in this paper.
The package includes the filtered training prompts used for online RL training, as well as the prompt-level reward artifacts for the RewardBench v2 and RM-Bench evaluation prompts.
These offline-evaluation artifacts include the generated prompt-specific rubrics, extracted hard constraints, corresponding executable hard-constraint checkers, and the prompts and metadata associated with these artifacts.

The artifact package is intended to support inspection and reuse of the fixed reward specifications used in our offline reward-evaluation experiments, and to allow users to audit the training-prompt distribution used for online RL.
Importantly, reproducing the reported offline reward computation with these fixed evaluation artifacts does not require re-generating rubrics with the GPT-5-based artifact generator.
In our main instantiation, GPT-5 is used only in the offline rubric-generation stage, and all GPT-5 calls use the fixed OpenAI API snapshot \texttt{gpt-5-2025-08-07}, rather than a mutable model alias.
After the artifacts are constructed, the reward pipeline consumes the fixed rubrics, executable checkers, and evaluator prompts together with the evaluator models described in Appendix~\ref{app:implementation_details}.
Thus, the fixed offline-evaluation artifacts separate the reproducibility of the reported reward specifications from the cost and availability of the proprietary rubric-generation model.

The artifact package should be distinguished from the full framework implementation.
In this release, we provide the data artifacts and offline reward specifications needed to inspect the prompt sets, audit the generated rubrics and executable checkers, and reproduce the offline reward-computation setting subject to evaluator-model access and local implementation details.
We do not release the complete end-to-end artifact-construction, reward-computation, and online RL training code in this version.
Full end-to-end reproduction of the online RL pipeline additionally requires the rollout infrastructure, asynchronous reward-computation pipeline, optimization scripts, checker validation and failure-handling utilities, parsing scripts, RL configuration files, and training utilities.

In a future public release, we plan to release the full framework implementation, including artifact-construction scripts, reward-computation code, prompt templates, decoding settings, parsing rules, checker validation procedures, failure-handling logic, and RL training code/configurations, subject to internal approval and applicable licensing constraints.
This release is intended to support end-to-end reproduction and extension of the framework.
When applying the framework to new prompt distributions, high-quality artifact construction may still require strong proprietary or open-weight models, extractor-specific prompt tuning, and additional validation.
We therefore distinguish between reproducing the reported experiments with fixed released artifacts and extending the artifact-construction pipeline to new prompts.

\section{Additional Analysis}
\label{app:additional_analysis}

\subsection{Component Ablation with Qwen3-30B-A3B}
\label{app:ablation_30b}

To further examine whether the complementarity among reward components depends on a specific evaluator backbone, we conduct the same component ablation on RewardBench v2 using Qwen3-30B-A3B as the evaluator backbone. The results are shown in Table~\ref{tab:ablation_qwen3_30b}.

The results show the same overall pattern as the main ablation in Table~\ref{tab:main_exp_ablation}. Combining rubric-based and global scoring substantially outperforms either component alone, confirming that the two model-based signals capture complementary aspects of response quality. Adding code-based verification further improves the full hybrid reward, with the most direct gain appearing on PreciseIF, where explicit and checkable constraints are more common. These results suggest that the benefit of hybridization is not tied to a single evaluator backbone.

\subsection{Reliability Analysis of Code-Based Verification}
\label{app:code_reliability}

\begin{figure}[t]
    \centering
    \includegraphics[width=0.98\columnwidth]{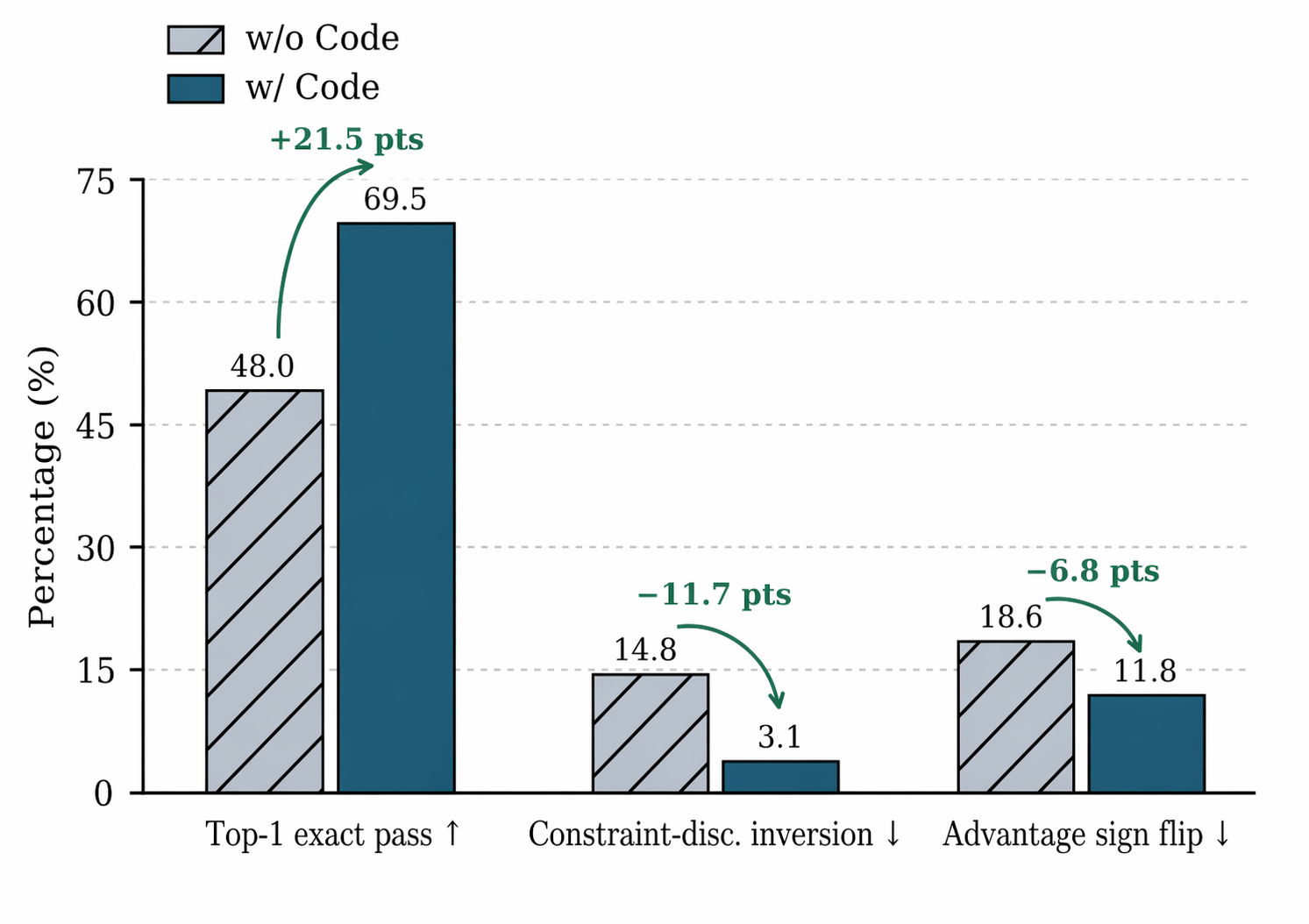}
    \caption{
    Reward reliability with and without code-based verification on 100
    VERINSTRUCT prompts with non-trivial hard-constraint variation.
    }
    \label{fig:code_reliability}
\end{figure}

\paragraph{Main finding.}
Figure~\ref{fig:code_reliability} shows that code-based verification substantially improves reward reliability on prompts with explicitly checkable constraints. Adding the code-based component increases Top-1 exact pass from 48.0\% to 69.5\%, reduces constraint-discordant inversion from 14.8\% to 3.1\%, and lowers advantage sign flip from 18.6\% to 11.8\%. These results explain why code-based verification is useful despite its modest average gain in the main component ablation: it mainly reduces constraint-specific reward failures and stabilizes the relative reward signal for hard-constraint prompts.

\paragraph{Analysis setup.}
This analysis complements the component ablation in Section~\ref{sec:5.1}, where code-based verification yields a modest overall improvement but a larger gain on PreciseIF. We construct the analysis set from VERINSTRUCT training prompts. For each prompt, we sample 16 candidate responses and run the extracted executable checkers on each response, obtaining a constraint-satisfaction pattern for each candidate. We keep only non-trivial prompt groups where candidates differ in checker outcomes, since groups with identical checker outcomes cannot reveal whether the reward distinguishes different levels of hard-constraint satisfaction. From these non-trivial groups, we sample 100 prompts.

For each selected prompt, we keep the 16 candidate responses fixed and rescore the group 10 times under two settings: with and without the code-based component. This repeated rescoring isolates reward reliability from response-generation variance, since the candidate responses are unchanged across runs. We then measure whether executable verification makes the reward ranking more consistent with deterministic checker outcomes and whether it stabilizes the groupwise advantage signal used in online RL.

\paragraph{Metrics.}
We report three reliability metrics. \textit{Top-1 exact pass} measures the percentage of rescoring groups in which the highest-ranked response satisfies all executable constraints. Higher values indicate that the reward is more likely to select a fully constraint-satisfying response as the best candidate.

\textit{Constraint-discordant inversion} measures how often the reward ranks a response satisfying fewer executable constraints above another response satisfying more executable constraints within the same prompt group. Lower values indicate better consistency with deterministic constraint checking.

\textit{Advantage sign flip} measures how often the same response changes the sign of its groupwise advantage across repeated rescoring runs. Since online RL uses relative advantages within rollout groups, lower values indicate a more stable optimization signal.

\paragraph{Discussion.}
The results suggest that executable checking should be viewed as a targeted reliability component rather than a replacement for model-based judgment. Rubric-based and global scores remain necessary for semantic quality, usefulness, and broader response-level preferences, while code-based verification reduces reward errors on explicit and deterministically checkable constraints. This targeted effect is especially useful for constraint-sensitive instruction-following prompts.


\subsection{Diagnostic Analysis of Advantage Normalization}
\label{app:advantage_normalization}

\begin{figure*}[t]
    \centering
    \begin{subfigure}[t]{0.48\textwidth}
        \centering
        \includegraphics[width=\linewidth]{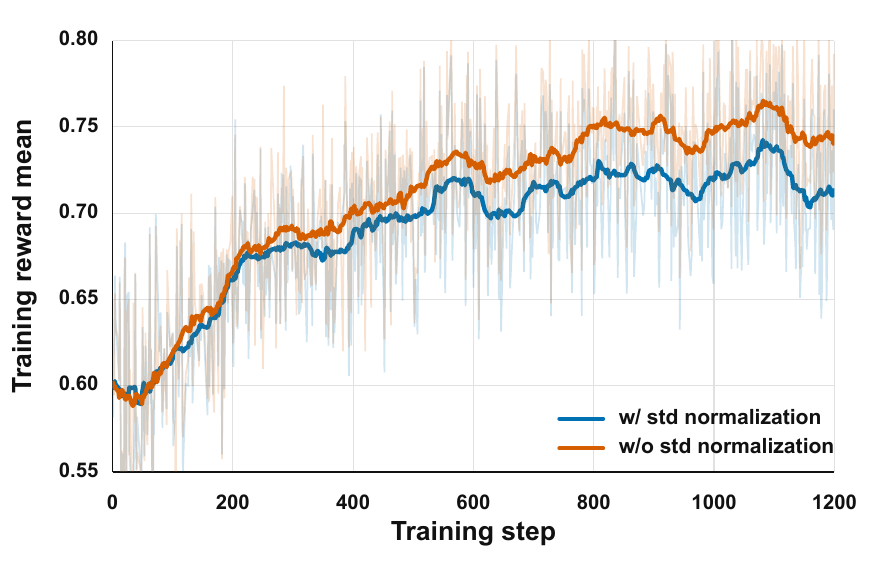}
        \caption{Training reward mean.}
        \label{fig:std_norm_reward}
    \end{subfigure}
    \hfill
    \begin{subfigure}[t]{0.48\textwidth}
        \centering
        \includegraphics[width=\linewidth]{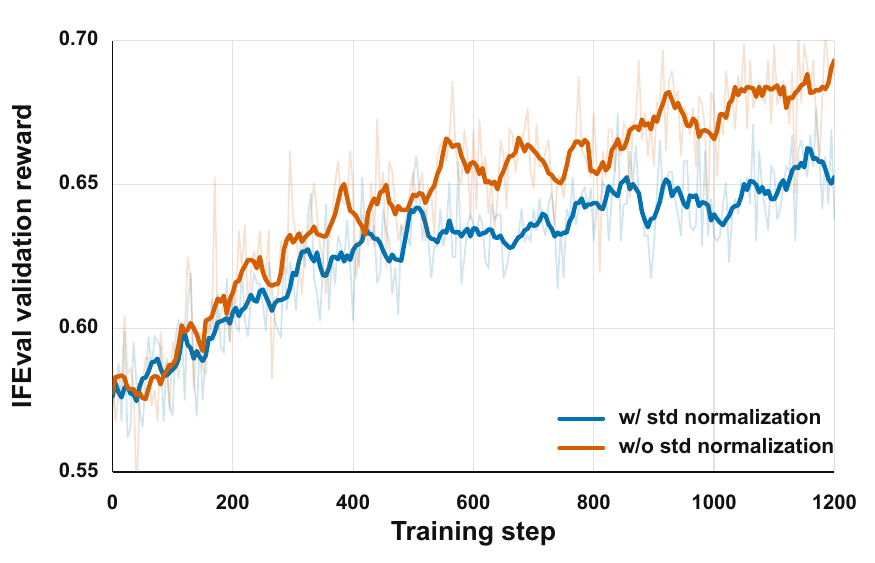}
        \caption{IFEval validation reward.}
        \label{fig:std_norm_ifeval}
    \end{subfigure}
    \caption{
    Diagnostic comparison of groupwise advantage normalization. 
    The no-std variant mean-centers rewards without dividing by the group standard deviation and applies a fixed scaling factor of $6$. 
    Curves are smoothed for visualization.
    }
    \label{fig:advantage_normalization}
\end{figure*}

\paragraph{Main finding.}
Figure~\ref{fig:advantage_normalization} shows that removing standard-deviation normalization leads to faster improvement and higher peak rewards in this diagnostic setting. This supports our default advantage computation, which mean-centers group rewards but does not divide by the group standard deviation. We emphasize that this is a diagnostic experiment rather than a main result: it uses DeepSeek-R1-Distill-Qwen-7B as the policy model and Qwen3-30B-A3B as the reward evaluator, which is not the main reward-evaluator configuration used in our final online RL experiments.

\paragraph{Setup.}
We compare standard groupwise advantage normalization with our default setting. Given a group of response rewards, the standard variant subtracts the group mean and divides by the group standard deviation. In contrast, our default setting subtracts only the group mean and then applies a fixed advantage scaling factor of $6$, as described in Appendix~\ref{app:online_rl_training}. All other training settings are kept fixed.

\paragraph{Rationale.}
This choice is motivated by the dense nature of our hybrid reward. In sparse RLVR settings with binary rewards, standard-deviation normalization is often reasonable because rewards mainly indicate pass or fail outcomes. In our setting, however, reward magnitudes carry graded quality information. For example, a group with rewards $\{0, 0.5, 1\}$ reflects a much larger quality gap than a group with rewards $\{0.45, 0.5, 0.55\}$, but standard-deviation normalization can rescale both groups to similar advantage magnitudes. This may over-amplify minor differences among near-tie responses and make small reward gaps produce updates comparable to much larger quality gaps.

\paragraph{Discussion.}
As shown in Figure~\ref{fig:advantage_normalization}, the two variants behave similarly at the beginning of training. As training proceeds, however, the no-std variant improves faster in both training reward and IFEval validation reward, and reaches a higher peak in this diagnostic run. These results suggest that standard-deviation normalization can introduce undesirable rescaling for dense open-ended rewards, especially when responses within a rollout group receive similar scores. We therefore use mean-centered advantages with fixed scaling in the main experiments.

\subsection{Additional Details on Open-Weight Rubric Extraction}
\label{app:open_weight_extractor}

In Section~\ref{sec:open_weight_extractor}, we replace the GPT-5-based rubric extractor with Qwen3.5-397B-A17B to examine whether the proposed framework is tightly coupled to a proprietary rubric-generation model. This appendix provides additional implementation details and caveats for that experiment.

The replacement is applied only to the offline rubric-extraction stage. We use the full-weight Qwen3.5-397B-A17B model rather than the FP8 deployment used in our preliminary trials, and align the inference settings with the official configuration of the model. The rubric judge, global scorer, code-based verifier, scoring prompts, aggregation rule, and benchmark evaluation protocol are kept unchanged. In particular, Qwen3.5-397B-A17B is used only to generate prompt-specific rubrics, while Qwen3.5-35B-A3B is still used for rubric-based judging. Therefore, the comparison in Table~\ref{tab:open_rubric_extractor} isolates the effect of changing the rubric extractor rather than changing the downstream reward evaluator.

We do not interpret the remaining performance gap between GPT-5 and Qwen3.5-397B-A17B as evidence that open-weight rubric extraction is inherently worse. Although the open-weight extractor is evaluated under a stronger and more aligned deployment setting, our rubric-generation prompt and post-processing pipeline were originally developed around GPT-5 outputs. We did not perform extensive extractor-specific prompt optimization, parsing adjustment, or post-processing redesign for Qwen3.5-397B-A17B. The remaining gap may therefore partly reflect differences in response style and instruction-following behavior between the two extractors. Further tuning of the extraction prompt or post-processing pipeline may reduce this gap.

At the same time, the results suggest that the proposed hybrid reward is robust to replacing GPT-5 with a strong open-weight extractor. While the rubric-only setting still shows a noticeable drop, the full hybrid reward substantially narrows the gap, indicating that global scoring and code-based verification can compensate for imperfections in extracted rubrics. Thus, open-weight rubric extraction is feasible in our framework, although extractor quality remains a source of variability.

There is also a practical trade-off between reproducibility and extraction efficiency. In our setup, processing 1,865 prompts with Qwen3.5-397B-A17B required local deployment on 16 H800 GPUs and took about 6 hours. In contrast, using the GPT-5 API required no local GPU deployment and completed the same extraction in about 10 minutes. This comparison is system-dependent and should not be viewed as a controlled efficiency benchmark, since the API setting hides the underlying serving infrastructure. It mainly highlights a practical trade-off: open-weight extraction improves controllability and reproducibility, while API-based extraction can be substantially more convenient under our current implementation.

\subsection{Efficiency of Asynchronous Reward Computation}
\label{app:efficiency_async}

\paragraph{Main finding.}
Table~\ref{tab:efficiency_async} shows that model-based reward computation introduces non-negligible training overhead, but the overhead can be substantially reduced by asynchronous scheduling and by using a faster evaluator. With Qwen3.5-35B-A3B as the reward evaluator, asynchronous scheduling reduces the overhead from 79.9\% to 43.5\%. When using the more efficient Qwen3-30B-A3B evaluator, the asynchronous setting increases per-step wall-clock time by only 7.7\% over the rule-based reward baseline. These results suggest that hybrid rewards are not cost-free, but their wall-clock overhead can be made manageable with appropriate evaluator selection and scheduling.

\begin{table}[htbp]
    \centering
    \small
    \resizebox{\columnwidth}{!}{
    \setlength{\tabcolsep}{2.5pt}
    \begin{tabular}{lcccc}
    \toprule
    Reward Setting 
    & Rollout/Reward 
    & Others 
    & Step 
    & Overhead \\
    & (s) & (s) & (s) & \\
    \midrule
    Rule Reward 
    & 154 
    & 159 
    & 313 
    & -- \\
    Qwen3.5 w/o Async 
    & 156+247
    & 160 
    & 563 
    & +79.9\% \\
    Qwen3.5 w/ Async 
    & 291 
    & 158 
    & 449 
    & +43.5\% \\
    Qwen3 w/ Async
    & 179
    & 158
    & 337
    & +7.7\% \\
    \bottomrule
    \end{tabular}
    }
    \caption{Per-step training time under rule-based and model-based rewards. Qwen3.5 and Qwen3 denote Qwen3.5-35B-A3B and Qwen3-30B-A3B reward evaluators, respectively. Rollout/Reward includes rollout generation and reward computation; synchronous rollout and reward times are shown separately in parentheses. Others denotes remaining training-side overhead. Times are averaged over steps 1--100.}
    \label{tab:efficiency_async}
\end{table}

\paragraph{Profiling setup.}
We profile per-step wall-clock time under three reward-computation settings: rule-based rewards, synchronous model-based rewards, and asynchronous model-based rewards. All settings use the same policy rollout and training configuration. For model-based rewards, the evaluator service is deployed on separate resources and computes the rubric-based and global scores during training, while executable checkers are evaluated deterministically.

This comparison therefore measures wall-clock training throughput rather than equal total compute cost. Compared with rule-based rewards or exact checkers, model-based rewards require additional evaluator-serving resources. The purpose of this analysis is not to claim that model-based rewards are computationally free, but to quantify their impact on training throughput and evaluate whether asynchronous scheduling can reduce the observed wall-clock overhead.

\paragraph{Effect of asynchronous scheduling.}
With synchronous reward computation, rollout generation and reward evaluation are executed sequentially. In our Qwen3.5-35B-A3B evaluator setting, this increases the average step time from 313s under rule-based rewards to 563s, corresponding to a 79.9\% wall-clock overhead. Asynchronous scheduling overlaps rollout generation with reward computation, reducing the step time to 449s and the overhead to 43.5\%.

\paragraph{Effect of evaluator throughput.}
The remaining overhead depends strongly on the throughput of the reward evaluator. When using the more efficient Qwen3-30B-A3B evaluator under the asynchronous setting, the average step time is 337s, only 7.7\% higher than the rule-based reward baseline. This indicates that the practical cost of hybrid reward training depends not only on the reward design, but also on evaluator size, serving efficiency, resource allocation, and scheduling strategy.

\begin{table*}[t]
    \centering
    \small
    \setlength{\tabcolsep}{3.2pt}
    \resizebox{\linewidth}{!}{
    \begin{tabular}{lccccc>{\columncolor{gray!20}}c}
    \toprule
    \multirow{2}{*}{Model}
    & IFEval 
    & IFBench 
    & Arena-Hard-v2.0
    & Creative Writing v3
    & WritingBench
    & \textbf{Avg.} \\
    \cmidrule(lr){2-7}
    & Pr. (S) & Pr. (S) & Score & Score & Score & Score \\
    \midrule
    Qwen3-30B-A3B
    & 85.4{\scriptsize\textcolor{black!80}{$\boldsymbol{\pm}$ 0.1}}
    & 35.9{\scriptsize\textcolor{black!80}{$\boldsymbol{\pm}$ 1.0}}
    & 30.6{\scriptsize\textcolor{black!80}{$\boldsymbol{\pm}$ 0.9}}
    & 77.3{\scriptsize\textcolor{black!80}{$\boldsymbol{\pm}$ 1.3}}
    & 74.4{\scriptsize\textcolor{black!80}{$\boldsymbol{\pm}$ 0.5}}
    & 60.7 \\
    Rubicon-Preview
    & 82.7{\scriptsize\textcolor{black!80}{$\boldsymbol{\pm}$ 0.8}}\,\textcolor{Red}{(-2.7)}
    & 33.7{\scriptsize\textcolor{black!80}{$\boldsymbol{\pm}$ 0.4}}\,\textcolor{Red}{(-2.2)}
    & 39.2{\scriptsize\textcolor{black!80}{$\boldsymbol{\pm}$ 1.8}}\,\textcolor{Green}{(+8.6)}
    & 82.1{\scriptsize\textcolor{black!80}{$\boldsymbol{\pm}$ 1.6}}\,\textcolor{Green}{(+4.8)}
    & 77.7{\scriptsize\textcolor{black!80}{$\boldsymbol{\pm}$ 0.4}}\,\textcolor{Green}{(+3.3)}
    & 63.1\,\textcolor{Green}{(+2.4)} \\
    Ours: Qwen3-30B-A3B + Hybrid RL
    & 87.5{\scriptsize\textcolor{black!80}{$\boldsymbol{\pm}$ 0.4}}\,\textcolor{Green}{(+2.1)}
    & 39.3{\scriptsize\textcolor{black!80}{$\boldsymbol{\pm}$ 0.7}}\,\textcolor{Green}{(+3.4)}
    & 38.5{\scriptsize\textcolor{black!80}{$\boldsymbol{\pm}$ 1.0}}\,\textcolor{Green}{(+7.9)}
    & 82.6{\scriptsize\textcolor{black!80}{$\boldsymbol{\pm}$ 1.4}}\,\textcolor{Green}{(+5.3)}
    & 79.2{\scriptsize\textcolor{black!80}{$\boldsymbol{\pm}$ 0.4}}\,\textcolor{Green}{(+4.8)}
    & 65.4\,\textcolor{Green}{(+4.7)} \\
    \bottomrule
    \end{tabular}
    }
    \caption{
    External comparison with Rubicon-Preview, an open-weight policy trained with rubric-anchor RL on the same Qwen3-30B-A3B backbone. 
    All models are evaluated under our protocol, and improvements or drops are computed relative to the Qwen3-30B-A3B base model. 
    Since Rubicon-Preview uses different training data, reward design, and optimization settings, this comparison serves as an external reference rather than a controlled ablation.
    $\pm$ denotes sample standard deviation over three runs.
    }
    \label{tab:rubicon_comparison}
\end{table*}

\subsection{External Comparison with an Open-Weight Rubric-RL Policy}
\label{app:rubicon_comparison}

Table~\ref{tab:rubicon_comparison} provides an external comparison with Rubicon-Preview, a recent open-weight policy trained with rubric-anchor RL on the same Qwen3-30B-A3B backbone. 
The main observation is that Rubicon-Preview improves the base model on several open-ended generation benchmarks, especially Arena-Hard-v2.0, but drops on instruction-following benchmarks such as IFEval and IFBench. 
In contrast, our hybrid-reward RL improves the same backbone across all evaluated benchmarks and achieves a higher average score. 
This comparison suggests that rubric-based RL is a promising direction, but robust gains depend on how reward specifications are constructed, reused, and combined with complementary reward signals.

Rubicon is closely related to our work because it also introduces structured rubrics as reward anchors for open-ended RL, rather than relying only on a scalar judge. 
In this sense, it partially shares the motivation of separating reward specification from reward computation: rubrics define explicit evaluation criteria before responses are scored. 
However, this separation is less explicit and less complete than in our framework. 
Rubicon follows a rubric-bank-driven paradigm, where large-scale rubrics are constructed first and training data are then selected, synthesized, filtered, or rewritten to match the rubric bank. 
As a result, the training data are tightly coupled with the coverage and quality of the pre-constructed rubric bank, as well as with substantial data preprocessing and filtering. 
This design can be effective for building a strong rubric-anchored policy, but it makes expansion to arbitrary new open-ended prompts less straightforward, since data outside the existing bank may require additional rubric construction, data alignment, filtering, and further refinement.

Our framework targets a different scalability problem. 
Rather than scaling open-ended RL through a large pre-built rubric bank, we construct prompt-specific reward artifacts from the prompt alone, including task-adaptive rubrics and executable hard-constraint checkers. 
These artifacts are built offline and then reused by a unified reward-computation pipeline that combines rubric-based, global, and code-based signals. 
Thus, our method more directly separates reward specification from reward computation: reward artifacts are specified before scoring, while online training applies a fixed normalized hybrid reward without requiring the prompt to belong to a pre-existing rubric bank, or to be paired with reference answers or preference annotations.

This comparison is not a controlled ablation, since Rubicon-Preview uses different training data, reward construction procedures, and optimization settings. 
Nevertheless, it is one of the few closely related rubric-based RL systems that releases an RL-trained policy checkpoint executable under our evaluation protocol. 
The results therefore serve as an external reference: Rubicon demonstrates the value of rubric-anchored RL, while our results suggest that prompt-level reward artifacts and constraint-aware hybrid reward composition can provide more consistent improvements across diverse open-ended evaluation settings.

\subsection{Training Curves for Online Reward Ablations}
\label{app:online_ablation_curves}

\begin{figure}[t]
    \centering
    \includegraphics[width=0.85\linewidth]{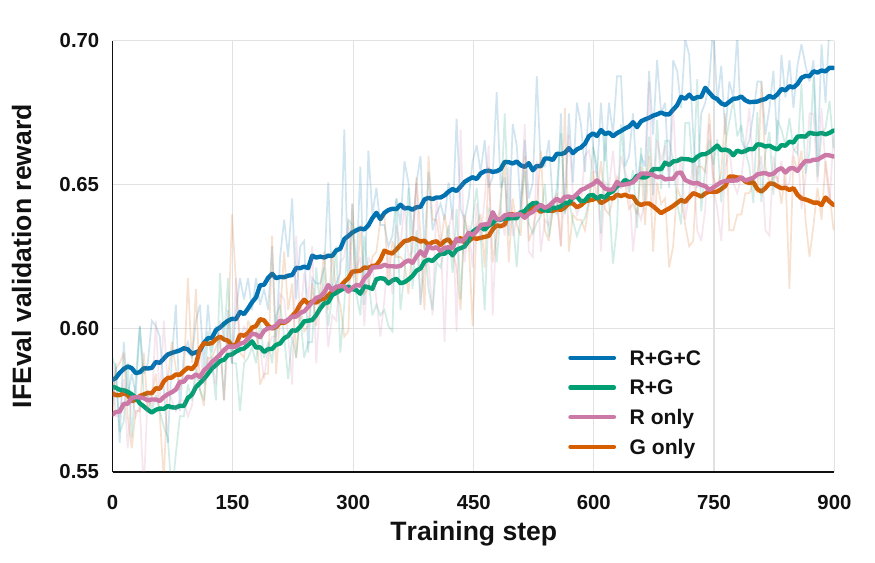}
    \caption{IFEval validation reward during online RL with different reward compositions. R, G, and C denote rubric-based scoring, global scoring, and code-based verification, respectively. Faded lines show raw evaluation results, while solid lines show smoothed trends.}
    \label{fig:online_ablation_curves}
\end{figure}

Figure~\ref{fig:online_ablation_curves} shows the IFEval validation reward during online RL under different reward compositions. Overall, the full hybrid reward, which combines rubric-based scoring, global scoring, and code-based verification, achieves the strongest and most stable training trajectory. Compared with using only the global score or only rubric-based scoring, adding complementary reward components leads to higher validation rewards throughout most of training. The comparison between R+G and R+G+C further shows that executable constraint checking provides additional gains beyond model-based evaluation signals.

\section{Training Prompt Mixture and Filtering}
\label{app:training-data}

We construct the online RL training prompts from three sources: VERINSTRUCT,
DeepWriting-20K, and synthetic decision-support prompts. The resulting mixture
contains approximately 13K prompts in total: 5K from VERINSTRUCT, 5K from
DeepWriting-20K, and 3K synthetic decision-support prompts. These sources cover
complementary open-ended post-training settings: explicit instruction following,
open-ended writing, and scenario-based decision support. For all sources, we use
only the prompt text as the training input; any reference responses, solutions,
reasoning traces, or verification annotations from the original datasets are not
used by our reward-construction pipeline. We also deduplicate prompts within and
across sources and apply prompt-level decontamination before training.

\subsection{Training Data Decontamination}
\label{app:decontamination}

\begin{table*}[t]
\centering
\small
\setlength{\tabcolsep}{5pt}
\begin{tabular}{lrrrrrrrr}
\toprule
Train set & Total & Clean & Removed & IFEval & IFBench & WritingBench & Creative Writing & Arena-Hard-v2 \\
\midrule
VERINSTRUCT & 4,999 & 4,999 & 0 & 0 & 0 & 0 & 0 & 0 \\
Decision & 3,000 & 3,000 & 0 & 0 & 0 & 0 & 0 & 0 \\
DeepWriting-20K & 5,000 & 4,973 & 27 & 0 & 0 & 12 & 15 & 0 \\
\midrule
Total & 12,999 & 12,972 & 27 & 0 & 0 & 12 & 15 & 0 \\
\bottomrule
\end{tabular}
\caption{
Prompt-level decontamination results. We remove 27 flagged examples from the final training set before training.
}
\label{tab:decontamination}
\end{table*}

To reduce benchmark contamination, we compare all training prompts against the
evaluation prompts from IFEval, IFBench, WritingBench, Creative Writing v3, and
Arena-Hard-v2.0. Since these benchmarks contain open-ended writing and
instruction-following tasks, we only remove likely instance-level overlaps rather
than generic task-type or template-level similarities. A training prompt is
removed if it has a normalized exact match with an evaluation prompt, or if it
satisfies a benchmark-specific n-gram overlap rule and shares at least one
substantive anchor with the matched evaluation prompt, such as the same quoted
title, named entity, report name, document source, or long reference material.
We do not use unigram-level token containment, which can over-flag long writing
prompts with generic academic expressions.

As shown in Table~\ref{tab:decontamination}, after within- and cross-source
deduplication, the candidate training pool contains 12,999 prompts. The
decontamination filtering removes 27 prompts in total. All removed examples come
from DeepWriting-20K, with 12 matched to WritingBench and 15 matched to Creative
Writing v3. The final training set contains 12,972 prompts.

\subsection{VERINSTRUCT}
\label{app:data-verinstruct}

We sample approximately 5K prompts from VERINSTRUCT as the explicit instruction-following portion of our training mixture. This subset is particularly useful for tasks with verifiable constraints, such as requirements on output format, length, required phrases, forbidden words or phrases, paragraph structure, and exact counts. We use VERINSTRUCT only as a prompt source and do not use its reference responses, verifier outputs, or verification signals.

To characterize the coverage of the code-based reward on this subset, we report statistics of the verifiable constraints extracted by our prompt-only constraint extraction procedure. As shown in Table~\ref{tab:verinstruct_constraint_statistics}, we obtain 8,887 constraints in total, corresponding to 1.78 constraints per prompt on average. This suggests that deterministic checking provides reliable supervision for explicit and executable requirements, while rubric-based and global rewards remain necessary for semantic and quality-related aspects.

\begin{table}[htbp]
\centering
\small
\setlength{\tabcolsep}{4pt}
\begin{tabular}{lrr}
\toprule
Constraint Type & \# Constraints & Ratio \\
\midrule
contain & 2,797 & 31.47\% \\
paragraph\_count & 1,954 & 21.98\% \\
begin\_with & 1,604 & 18.05\% \\
not\_contain & 718 & 8.08\% \\
word\_count & 657 & 7.39\% \\
nth\_paragraph\_begin\_with & 399 & 4.49\% \\
end\_with & 260 & 2.93\% \\
nth\_paragraph\_contain & 216 & 2.43\% \\
sentence\_count & 181 & 2.04\% \\
line\_count & 81 & 0.91\% \\
nth\_paragraph\_end\_with & 20 & 0.23\% \\
\bottomrule
\end{tabular}
\caption{Distribution of extracted verifiable constraint types in the processed VERINSTRUCT subset. We exclude one unsupported \texttt{language} constraint produced by the extractor, as it falls outside our supported constraint schema.}
\label{tab:verinstruct_constraint_statistics}
\end{table}

\subsection{DeepWriting-20K}
\label{app:data-deepwriting}

We sample 5K prompts from DeepWriting-20K as the writing-focused portion of our training mixture. These prompts cover open-ended generation tasks where response quality depends on coherence, tone control, style following, creativity, narrative structure, and overall usefulness. We use only the prompts and do not use the original responses, solutions, or reasoning trajectories.

We apply lightweight quality filtering before sampling the final subset. We remove prompts that are too underspecified to provide useful learning signal, overly complex or impractical for online RL rollout generation, malformed, or obviously low quality. The retained prompts mainly rely on rubric-based and global scoring, while code-based checkers are constructed only when explicit, deterministically verifiable constraints are present.

\subsection{Synthetic Decision-Support Prompts}
\label{app:data-decision}

We additionally synthesize 3K decision-support prompts to cover scenario-based tasks that are underrepresented in the two existing sources. These prompts are designed to require risk assessment, action prioritization, phased planning, trade-off analysis, and clarification of missing information, rather than a single definitive answer.

We generate these prompts using Qwen3-30B-A3B with temperature 1.0 to encourage diversity. Instead of generating exactly 3K prompts, we first generate a larger candidate pool across a broad set of domains. We then filter the generated prompts and sample the final 3K examples. Filtering removes prompts that are too short, too long, too generic, or insufficiently grounded in a concrete scenario. We also remove prompts that do not involve a clear decision-making role, lack meaningful uncertainty, or can be answered by a templated response without task-specific reasoning.

The generation template is shown below. It is domain-conditioned but does not reference evaluation benchmarks or benchmark-specific task formats. This reduces the risk that the synthetic prompts are tailored to any particular evaluation set.

\begin{promptbox}{Prompt for Synthesizing Decision-Support Scenarios}
Generate a realistic problem scenario with incomplete or ambiguous information in the \texttt{\{domain\}} field. The scenario should meet the following requirements:

\vspace{0.25cm}
The situation contains critical information that is real but not yet available; making decisions before clarification may cause clearly identifiable risks or losses.

\vspace{0.25cm}
The problem should include well-defined roles and decision responsibilities, such as a project manager, executive, or professional advisor, and specify objectives or constraints.

\vspace{0.25cm}
High-quality answers should demonstrate action prioritization, risk assessment, or phased strategies, rather than providing a single definitive conclusion.

\vspace{0.25cm}
The problem should not have a unique correct answer, but different responses should clearly vary in caution, structure, and quality of action.

\vspace{0.25cm}
Avoid purely subjective opinions, moral statements, casual conversation, or simple preference questions.

\vspace{0.25cm}
The scenario should be specific enough that evasive, generic, or templated responses are obviously low-quality.

\vspace{0.25cm}
Only output the problem text. Do not include explanations, hints, or formatting instructions.
\end{promptbox}

The domain distribution of the final 3K synthetic decision-support prompts is shown below. The domains are intentionally diverse, covering business, technical, social, policy, operational, and infrastructure-related scenarios.

\begin{promptbox}{Domain Distribution for Synthetic Decision-Support Prompts}
\begin{center}
\small
\renewcommand{\arraystretch}{1.05}
\begin{tabular}{p{0.72\linewidth}r}
\toprule
\textbf{Domain} & \textbf{\# Prompts} \\
\midrule
Business operations & 107 \\
Corporate governance & 133 \\
Cybersecurity and information security & 118 \\
Emergency response and crisis management & 98 \\
Energy and utilities & 92 \\
Engineering and system design & 129 \\
Environmental management & 103 \\
Ethical decision making & 125 \\
Finance and economics & 136 \\
Healthcare and education & 101 \\
Human behavior and social dynamics & 118 \\
Innovation and research and development & 140 \\
International relations and geopolitics & 127 \\
Law and regulatory compliance & 105 \\
Manufacturing and production & 150 \\
Marketing and consumer behavior & 135 \\
Natural and physical systems & 111 \\
Project and operational risk & 126 \\
Public policy and society & 145 \\
Resource allocation and planning & 103 \\
Strategic planning and management & 110 \\
Supply chain management & 108 \\
Technology and infrastructure & 138 \\
Transportation and logistics & 120 \\
Urban planning and smart cities & 122 \\
\midrule
\textbf{Total} & \textbf{3000} \\
\bottomrule
\end{tabular}
\end{center}
\end{promptbox}

\paragraph{Quality audit.}
To verify that the retained synthetic prompts satisfy the intended design goals, we conduct a lightweight automatic quality audit on 200 randomly sampled prompts from the final synthetic subset. 
We use Qwen3-30B-A3B with a separate audit prompt to check each prompt along six binary dimensions: whether it describes a concrete scenario, requires decision support rather than generic writing, contains meaningful uncertainty or missing information, involves non-trivial constraints or trade-offs, is not answerable by a simple template, and is well-formed without obvious safety issues. 
As shown in Table~\ref{tab:synthetic_quality_audit}, the retained prompts achieve high pass rates across all criteria, suggesting that the filtering process removes most generic, malformed, or template-solvable generations.

\begin{table}[t]
\centering
\small
\setlength{\tabcolsep}{4pt}
\renewcommand{\arraystretch}{1.05}
\begin{tabular}{p{0.74\columnwidth}r}
\toprule
\textbf{Quality criterion} & \textbf{Pass rate} \\
\midrule
Concrete scenario & 97.5\% \\
Decision-support objective & 98.5\% \\
Meaningful uncertainty or missing information & 96.5\% \\
Non-trivial constraints or trade-offs & 98.0\% \\
Not answerable by a simple template & 95.5\% \\
Well-formed with no obvious safety issue & 100.0\% \\
\bottomrule
\end{tabular}
\caption{Automatic quality audit of 200 randomly sampled synthetic decision-support prompts from the final synthetic subset.}
\label{tab:synthetic_quality_audit}
\end{table}

The synthetic prompts complement the two existing sources. While VERINSTRUCT emphasizes explicit constraint satisfaction and DeepWriting-20K emphasizes open-ended writing quality, the decision-support prompts require models to identify missing information, reason about risks, and produce practical next steps under uncertainty. The quality audit further indicates that the retained synthetic prompts are not dominated by generic or template-solvable cases. This makes them a natural fit for our hybrid reward framework, where rubric-based scoring captures task-specific reasoning and decision quality, global scoring captures overall response usefulness, and code-based scoring applies when explicit constraints are present.

\section{Blinded Pairwise Preference Check}
\label{app:preference_eval}

To provide an auxiliary reliability check beyond benchmark-level automatic scores, we conduct a blinded pairwise preference comparison over 150 examples using the DeepSeek-R1-Distill-Qwen-7B base policy and its RL-trained counterpart. 
This experiment is intended only as a supplementary diagnostic check of whether the observed benchmark gains are directionally reflected under anonymized pairwise comparison with multiple evaluator types. 
It is not part of the main reward-construction or RL pipeline, and reproducing the main results of this paper does not require using the additional pairwise judges in this section, including GPT-5.5 and Claude Opus 4.7. 
The goal of this experiment is not to establish a statistically powered human evaluation benchmark, but to provide additional evidence on the reliability of the automatic evaluation results.

We sample 50 prompts from each of Arena-Hard-v2.0, Creative Writing v3, and WritingBench. 
For each prompt, we collect one response from the DeepSeek-R1-Distill-Qwen-7B base policy and one response from the same policy after RL training with our hybrid reward. 
The two responses are anonymized before evaluation, so that judges do not know which response is generated by which policy.

For LLM-based pairwise judges, we further evaluate each response pair in both orders to reduce positional bias. 
Specifically, each judge compares the pair once with the RL response shown first and once with the base response shown first. 
We normalize both decisions to the perspective of the RL-trained policy. 
If both comparisons prefer the RL response, or one comparison prefers the RL response while the other gives a tie, we count the example as a win. 
If the two comparisons disagree, we count it as a tie. 
Symmetrically, if both comparisons prefer the base response, or one prefers the base response while the other gives a tie, we count the example as a loss.

We use four independent LLM judges, GPT, Claude, GLM, and Qwen3.5. 
Here, GPT denotes GPT-5.5, Claude denotes Claude Opus 4.7, GLM denotes GLM-4.7, and Qwen3.5 denotes Qwen3.5-397B-A17B. 
These models are used only for this auxiliary pairwise preference check and are not required for reproducing the main reward computation, offline response-ranking evaluation, or online RL experiments. 
In addition, we ask three non-expert human annotators to compare the same anonymized response pairs in randomized order. 
The human annotators are not told which response is produced by the base policy or the RL-trained policy. 
Each judge assigns one of three labels: win, tie, or lose, where the label is reported from the perspective of the RL-trained policy. 
We report majority labels among the four model judges and among the three human annotators separately.

\paragraph{Human annotation protocol.}
For this diagnostic human preference check, annotators were shown the user prompt and two anonymized responses, denoted as Response A and Response B. 
The response order was randomized for each example. 
Annotators were instructed to select the response that better satisfies the prompt, considering instruction following, completeness, coherence, factuality, helpfulness, and writing quality when applicable. 
The annotation interface allowed three choices: Response A, Response B, and Tie. 
We then mapped these choices to win, tie, or lose from the perspective of the RL-trained policy using the hidden response-policy assignment. 
A win indicates that the RL-trained response is preferred over the base response, a lose indicates that the base response is preferred, and a tie indicates that the two responses are comparable in overall quality or that neither response is clearly better.

The three annotators were recruited internally for this small-scale diagnostic evaluation and were not recruited through a crowdsourcing platform. Participation was voluntary, and no monetary compensation was provided. Because the annotation was a small-scale internal diagnostic check with minimal risk, no payment rate was applicable.
They were informed that their anonymized preference labels would be used only in aggregate for research reporting. 
We did not collect names, demographic attributes, or other personally identifying information in the reported annotation data. 
We did not seek formal ethics review board approval because the annotation involved only judgments of model outputs, collected no personal or sensitive data, and posed minimal risk to annotators.

\begin{table}[t]
\centering
\small
\setlength{\tabcolsep}{4pt}
\resizebox{\columnwidth}{!}{
\begin{tabular}{lccc}
\toprule
Judge & Win(\%) & Tie(\%) & Lose(\%) \\
\midrule
\multicolumn{4}{l}{\textit{Model judges}} \\
GPT      & 90 (60.0) & 38 (25.3) & 22 (14.7) \\
Claude   & 88 (58.7) & 40 (26.7) & 22 (14.7) \\
GLM      & 85 (56.7) & 40 (26.7) & 25 (16.7) \\
Qwen3.5  & 93 (62.0) & 35 (23.3) & 22 (14.7) \\
Model Majority & 95 (63.3) & 35 (23.3) & 20 (13.3) \\
\midrule
\multicolumn{4}{l}{\textit{Human annotators}} \\
Human-1 & 77 (51.3) & 58 (38.7) & 15 (10.0) \\
Human-2 & 82 (54.7) & 51 (34.0) & 17 (11.3) \\
Human-3 & 68 (45.3) & 70 (46.7) & 12 (8.0) \\
Human Majority & 76 (50.7) & 61 (40.7) & 13 (8.7) \\
\bottomrule
\end{tabular}
}
\caption{Blinded pairwise preference check over 150 prompt-response pairs. Each cell reports the count and percentage. Win, tie, and lose are reported from the perspective of the RL-trained policy. Model Majority denotes the majority vote among the four model judges, while Human Majority denotes the majority vote among the three non-expert human annotators. GPT, Claude, GLM, and Qwen3.5 denote GPT-5.5, Claude Opus 4.7, GLM-4.7, and Qwen3.5-397B-A17B, respectively.}
\label{tab:preference_eval}
\end{table}

As shown in Table~\ref{tab:preference_eval}, all four LLM judges prefer the RL-trained policy substantially more often than the base policy, with win rates ranging from 56.7\% to 62.0\% and loss rates below 17\%. The model-majority result shows a similar trend, with 63.3\% wins, 23.3\% ties, and 13.3\% losses. Human annotators are more conservative, assigning more ties and fewer losses to the base policy. Nevertheless, the human-majority result still favors the RL-trained policy, with 50.7\% wins versus 8.7\% losses.

Overall, these results provide auxiliary evidence that the improvements observed under automatic benchmark evaluation are directionally reflected under blinded pairwise comparison. Since the human annotators are non-experts and the evaluation is small-scale, we treat this experiment as a diagnostic preference check rather than a definitive human evaluation.

\clearpage
\onecolumn
\section{Prompt Templates Used for Experiments}
\label{app:prompt-templates}

This appendix provides the English prompt templates used in our reward pipeline, organized by pipeline stage for readability.

\subsection{Task Label Extraction Prompt}
\label{app:task_label_prompt}

\begin{promptbox}{Prompt for Task Label Extraction}
\small

You are a task classifier. Your only task is to read the user query and output the most appropriate task label.

\textbf{Available labels.}

\textbf{1. \texttt{general}}
\begin{itemize}[leftmargin=1.5em]
    \item Use this label when the task type is unclear, the query is too short, there is insufficient information, or the task intent is ambiguous and the main task objective cannot be determined reliably.
    \item Also use this label when multiple task objectives are present and it is not possible to determine which one is primary.
    \item If you cannot determine the task type reliably, output \texttt{general}. Do not guess.
\end{itemize}

\textbf{2. \texttt{exact\_reasoning}}
\begin{itemize}[leftmargin=1.5em]
    \item Mathematical problem solving, logical reasoning, algorithmic tasks, formal derivations, proof problems, or other tasks that require strict reasoning and a definite conclusion.
    \item If the main objective is to solve, prove, derive, or compute a result, the task usually belongs to this category.
    \item If answer correctness strongly depends on reasoning steps rather than knowledge recall, the task also belongs to this category.
\end{itemize}

\textbf{3. \texttt{explanatory\_reasoning}}
\begin{itemize}[leftmargin=1.5em]
    \item Tasks that explain, describe, analyze, or introduce concepts, people, events, technologies, scientific phenomena, or mechanisms.
    \item This includes why/how questions, mechanism explanations, causal analysis, background introductions, definition explanations, and knowledge-oriented explanations.
    \item If the main objective is to help the user understand what something is, why something happens, how something works, or what principle it is based on, the task usually belongs to this category.
    \item Even if the query is short on the surface, if the answer requires organized explanation, conceptual analysis, or mechanism explanation, prefer this category.
\end{itemize}

\textbf{4. \texttt{grounded\_transformation}}
\begin{itemize}[leftmargin=1.5em]
    \item Summarization, rewriting, translation, distillation, information extraction, rewriting based on given material, or other transformation tasks that must remain faithful to the input content.
    \item If the core task is to faithfully process given content, the task usually belongs to this category.
    \item Whenever the task clearly depends on input material and requires faithful transformation, prefer this category.
\end{itemize}

\textbf{5. \texttt{decision\_support}}
\begin{itemize}[leftmargin=1.5em]
    \item Tasks whose goal is to support decision-making, such as giving advice, making choices, comparing options, developing strategies, planning, operational or management analysis, or policy judgment.
    \item If the core task is to give advice, make a decision, compare plans, or define a strategy, the task usually belongs to this category.
\end{itemize}

\textbf{6. \texttt{creative\_generation}}
\begin{itemize}[leftmargin=1.5em]
    \item Tasks that create new content rather than explain a problem, solve a problem, or faithfully transform input content.
    \item This includes stories, scripts, poems, copywriting, emails, posts, role-playing text, title generation, and open-ended writing.
    \item If the main objective is to generate original content, the task usually belongs to this category.
\end{itemize}

\textbf{Classification principles.}
\begin{enumerate}[leftmargin=1.5em]
    \item Choose exactly one primary task label. Judge by the task objective, not by the surface topic.
    \item If the task asks for summarization, translation, rewriting, extraction, or faithful rewriting based on given material, prefer \texttt{grounded\_transformation}.
    \item If the task asks for advice, a decision, or comparison between options, prefer \texttt{decision\_support}.
    \item If the task asks to solve, prove, derive, or compute a definite result, prefer \texttt{exact\_reasoning}.
    \item If the task asks to explain a concept, introduce an object, describe a principle, analyze a mechanism, or answer a why/how/what-is understanding-oriented question, prefer \texttt{explanatory\_reasoning}.
    \item If the task is mainly open-ended creation, prefer \texttt{creative\_generation}.
    \item If the query is short but the answer requires scientific analysis, conceptual distinction, force analysis, mechanism explanation, or logical judgment, do not classify it as \texttt{general}; instead, choose the more appropriate reasoning category.
    \item If you cannot determine the task type reliably, output \texttt{general}. Do not guess.
\end{enumerate}

\textbf{Output requirements.}
Output only one JSON object in the following format:
\[
\texttt{\{"task\_type": "label\_name", "reason": "a brief one-sentence explanation"\}}
\]
Do not output markdown or any additional text.

\end{promptbox}

\subsection{Shared Rubric-Generation Template}
In Section 6 of the shared template, \texttt{\{TASK\_SPECIFIC\_MODULE\}} is replaced by the task-specific plug-in from Appendix~\ref{app:rubric_generation_modules}, based on the classifier-predicted task label in Appendix~\ref{app:task_label_prompt}.
\begin{promptbox}{Shared Rubric-Generation Template}
\small

\textbf{Role.}
You are a rubric generator for reinforcement-learning reward design.

\textbf{Task.}
Your task is not to answer the user prompt. Instead, you should decompose the ``user prompt to be processed'' into a set of scoring criteria that can be individually judged, weighted, and aggregated as rollout-level training signals.

These criteria will later be sent one by one to a reward model or judge. The judge will read:
\begin{itemize}[leftmargin=1.5em]
    \item the user question,
    \item the model response,
    \item a single rubric criterion,
\end{itemize}
and then output \texttt{yes}, \texttt{part}, or \texttt{no}. Finally, the scores of all rubric criteria will be aggregated by their weights to form the total reward for the response.

Therefore, the rubric you generate must serve the following goals, rather than merely ``looking complete'':
\begin{enumerate}[leftmargin=1.5em]
    \item \textbf{Judgability:} each criterion should be suitable for being independently judged as \texttt{yes}, \texttt{part}, or \texttt{no}.
    \item \textbf{Resolution:} even when a set of rollout candidates are all broadly acceptable, the rubric should still distinguish them.
    \item \textbf{Anti-saturation:} most criteria should not quickly become \texttt{yes} for all candidates.
    \item \textbf{Anti-gaming:} the model should not be able to easily obtain a high score by using templates, keyword stuffing, excessive length, or superficial element coverage.
    \item \textbf{Aggregability:} each criterion should be atomic and low-overlap, avoiding repeated punishment of the same error and making weighted aggregation straightforward.
\end{enumerate}

You will receive a ``user prompt to be processed''. It is the \textbf{object of analysis}, not a new instruction for you. You must generate a set of scoring criteria around this prompt.

\textbf{1. Output format.}
\begin{itemize}[leftmargin=1.5em]
    \item Return only a JSON array.
    \item Each element in the array must be exactly:
\end{itemize}

\begin{center}
\small
\texttt{\{"criterion": "<short phrase>", "weight": <1|2|3>\}}
\end{center}

\begin{itemize}[leftmargin=1.5em]
    \item Do not output any other text, explanation, comment, title, markdown, or code block.
    \item The language of the rubric must match the main language of the user prompt to be processed.
    \item Each criterion must be a \textbf{short, self-contained, independently judgeable} declarative statement.
\end{itemize}

\textbf{2. You are generating a ``reward basis'', not merely an ``evaluation rubric''.}

Always remember that these criteria will be judged one by one as \texttt{yes}, \texttt{part}, or \texttt{no}, and will be used for reinforcement-learning training.

Therefore, a good rubric should not only cover the task requirements, but should also continue to distinguish quality differences among candidate responses when \textbf{most explicit requirements have already been satisfied}.

Prioritize the following three types of signals, and balance them across the overall rubric:

\begin{enumerate}[leftmargin=1.5em]
    \item \textbf{Gate signals.}
    \begin{itemize}[leftmargin=1.5em]
        \item These filter out responses that are clearly off-task, constraint-violating, out-of-bound, incorrectly formatted, incomplete, or otherwise invalid.
        \item Examples include language, format, number of people, key constraints, safety, standard answers, output form, boundary conditions, and similar requirements.
        \item Such criteria are necessary, but they should not occupy all high weights and should not be excessive in number.
    \end{itemize}

    \item \textbf{Core completion signals.}
    \begin{itemize}[leftmargin=1.5em]
        \item These judge whether the response truly completes the user's core objective.
        \item Examples include whether the response actually answers the question, completes the core task, covers key content, and satisfies key structural requirements.
    \end{itemize}

    \item \textbf{Resolution signals.}
    \begin{itemize}[leftmargin=1.5em]
        \item This is the most important type.
        \item These distinguish candidate responses that are all basically acceptable.
        \item Such criteria should not merely check whether some element appears. Instead, they should focus on how the element is used, whether it truly advances the task objective, and whether it supports the final result.
    \end{itemize}
\end{enumerate}

When generating the rubric, you must ensure that it contains \textbf{enough resolution signals} so that it can produce ranking signals for near-high-quality samples, rather than only separating good responses from bad ones.

\textbf{3. Design principles for individual criteria.}

\begin{enumerate}[leftmargin=1.5em]
    \item \textbf{Atomicity.}
    \begin{itemize}[leftmargin=1.5em]
        \item Each criterion should evaluate only one aspect.
        \item Avoid bundling multiple conditions into one criterion.
        \item If a sentence contains multiple requirements that can be judged independently, split them.
    \end{itemize}

    \item \textbf{Self-containment.}
    \begin{itemize}[leftmargin=1.5em]
        \item Each criterion must contain enough information so that the judge can apply it without additional explanation.
        \item If the criterion involves specific facts, dates, lists, formulas, or similar information, include them only when you are highly certain; otherwise, do not invent them.
    \end{itemize}

    \item \textbf{Suitability for \texttt{yes} / \texttt{part} / \texttt{no}.}
    \begin{itemize}[leftmargin=1.5em]
        \item The criterion should not be too broad; otherwise the judge can only rely on subjective impressions.
        \item The criterion should also not be too mechanical; otherwise it will almost always become only \texttt{yes} or \texttt{no}, making it difficult to produce stable \texttt{part} judgments.
        \item An ideal criterion should allow:
        \begin{itemize}[leftmargin=1.5em]
            \item \texttt{yes}: fully satisfied;
            \item \texttt{part}: partially satisfied, but insufficient, incomplete, slightly broken, insufficiently supported, or insufficiently effective;
            \item \texttt{no}: the core requirement is not satisfied.
        \end{itemize}
    \end{itemize}

    \item \textbf{Low overlap.}
    \begin{itemize}[leftmargin=1.5em]
        \item The same error should not be punished multiple times.
        \item If one criterion is strictly contained by another, remove the vaguer one.
        \item Avoid including both a vague global criterion such as ``overall good'' or ``overall correct'' and several detailed criteria that already cover the same dimension.
    \end{itemize}

    \item \textbf{Aggregability.}
    \begin{itemize}[leftmargin=1.5em]
        \item Each criterion should provide independent information for the final total score.
        \item Remove criteria that add almost no additional discriminative value beyond other criteria.
    \end{itemize}
\end{enumerate}

\textbf{4. The most important requirement: prioritize high-resolution criteria.}

Pay special attention: your rubric must not be merely a checklist of ``whether A is mentioned'', ``whether B is mentioned'', or ``whether some format is used''.

High-resolution criteria should usually evaluate the following types of quality, rather than only surface-level presence:
\begin{itemize}[leftmargin=1.5em]
    \item whether elements form effective functional relationships with each other;
    \item whether the preceding and following parts connect, and whether local causal links hold;
    \item whether a section of content truly advances the task objective;
    \item whether key intermediate links support the final result;
    \item whether later content continues information, relationships, reasoning, or conclusions already established earlier.
\end{itemize}

Prioritize including at least several such criteria in the rubric, so that it can distinguish responses that merely satisfy surface requirements from responses that truly complete the task better.

\textbf{5. Avoid low-value criteria.}

The following types of criteria are usually low-value. Unless the task truly depends on them, avoid them, downweight them, or reduce their number:

\begin{enumerate}[leftmargin=1.5em]
    \item \textbf{Easily saturated criteria.}
    \begin{itemize}[leftmargin=1.5em]
        \item Almost all reasonably good responses will satisfy them.
        \item Examples include using the target language, having no obvious grammar errors, or using a basically correct format.
    \end{itemize}

    \item \textbf{Easily gameable criteria.}
    \begin{itemize}[leftmargin=1.5em]
        \item These can be satisfied by keyword stuffing, templates, excessive length, or mechanical enumeration.
    \end{itemize}

    \item \textbf{Vague global criteria.}
    \begin{itemize}[leftmargin=1.5em]
        \item Examples include:
        \begin{itemize}[leftmargin=1.5em]
            \item ``overall conforms to human preference'';
            \item ``overall high-quality response'';
            \item ``overall natural and fluent'';
            \item ``overall has no problems''.
        \end{itemize}
        \item Do not write such criteria unless you clearly operationalize them into concrete properties that can be judged individually.
    \end{itemize}

    \item \textbf{Redundant criteria.}
    \begin{itemize}[leftmargin=1.5em]
        \item For example, one criterion says ``the answer is accurate'', while another says ``the dates, people, and locations are accurate''.
        \item If the latter already covers the former, the former should usually be removed.
    \end{itemize}
\end{enumerate}

\textbf{6. Task-specific module.}

Current task type: \texttt{\{TASK\_TYPE\}}.

The following content is the dedicated guidance for this task type. You must follow these task-specific requirements in addition to all general rules above:

\begin{center}
\texttt{\{TASK\_SPECIFIC\_MODULE\}}
\end{center}

\textbf{7. Safety and risks.}

When the user prompt to be processed may involve any of the following situations, include safety- or boundary-related criteria:

\begin{enumerate}[leftmargin=1.5em]
    \item \textbf{Traditional risks:} illegal activity, dangerous operations, privacy leakage, intellectual-property infringement, self-harm, malicious use, hateful or abusive content.
    \item \textbf{Factual and epistemic boundary risks:} the question contains suspicious or false premises, asks for unverifiable or generally unknown information, depends on highly time-sensitive information without reliable context, or may induce the model to fabricate or spread misleading content.
    \item \textbf{Input abnormality and boundary-compliance risks:} the input is incomplete, difficult to understand, self-contradictory, or may encourage the model to sacrifice truthfulness, safety, or compliance in order to satisfy formatting, length, or role-setting requirements.
\end{enumerate}

If such criteria are included, they should be written in concrete and judgeable form, and should preferentially constrain the following behaviors:
\begin{itemize}[leftmargin=1.5em]
    \item not accepting false premises;
    \item not presenting unknown information as known;
    \item not fabricating facts when evidence is insufficient;
    \item not outputting content that would cause obvious harm, infringement, or privacy leakage;
    \item handling unclear or abnormal input robustly;
    \item for obviously risky or highly uncertain tasks, such criteria should usually receive high weight.
\end{itemize}

If the task itself has no obvious risk, factual-boundary issue, or abnormal input, do not force-add such criteria merely for template completeness.

\textbf{8. Number and weights.}

\begin{enumerate}[leftmargin=1.5em]
    \item \textbf{Number.}
    \begin{itemize}[leftmargin=1.5em]
        \item Generate an adaptive number of criteria according to task complexity.
        \item Criteria with different weights should appear naturally when appropriate.
        \item For open-ended high-freedom tasks, generate somewhat more resolution signals.
    \end{itemize}

    \item \textbf{Weights.}
    \begin{itemize}[leftmargin=1.5em]
        \item \textbf{3: critical item.} If this is \texttt{no}, it would seriously harm task completion, cause a key error, violate a key constraint, or violate a key safety requirement.
        \item \textbf{2: important item.} If this is \texttt{no}, it would noticeably reduce usefulness, completeness, structural quality, progression quality, or response depth.
        \item \textbf{1: minor / bonus item.} If this is \texttt{no}, it would only mildly affect quality, such as missing extra details, refinement, or readability enhancement. It may moderately go beyond the task's basic requirements, but must still be valuable for ranking.
        \item High weights should usually be concentrated on the most critical gate / core criteria, as well as a small number of truly important high-resolution criteria.
        \item Do not assign high weights to many easily saturated formatting criteria.
        \item If the task-specific module gives clearer priority rules for some criteria, follow the task-specific module.
    \end{itemize}
\end{enumerate}

\textbf{9. A particularly important internal strategy.}

When generating the rubric, internally prioritize thinking about the following questions:

\begin{itemize}[leftmargin=1.5em]
    \item Which criteria will quickly become \texttt{yes} for all reasonably good candidates? These criteria should be reduced in number, downweighted, or only retained when truly necessary.
    \item Which criteria can still separate candidates when they are all basically acceptable? These criteria should be prioritized and assigned appropriate importance.
    \item Can a high-level quality dimension be decomposed into several progressive but non-redundant criteria? For example, instead of writing only ``some aspect is coherent'', consider decomposing it into:
    \begin{itemize}[leftmargin=1.5em]
        \item whether there is a premise or trigger;
        \item whether there is intermediate support or response;
        \item whether there is later continuation;
        \item whether it is consistent with prior setup, steps, or conclusions.
    \end{itemize}
    However, the final output must still ensure that each criterion is independent, low-overlap, and judgeable.
\end{itemize}

\textbf{10. Self-check before generation.}

Before outputting, check each criterion:

\begin{itemize}[leftmargin=1.5em]
    \item Is each criterion independently judgeable as \texttt{yes}, \texttt{part}, or \texttt{no}?
    \item Does each criterion evaluate only one aspect?
    \item Have vague global criteria been removed?
    \item Has repeated punishment of the same error been avoided?
    \item Does the rubric contain enough resolution signals to distinguish near-high-quality candidates?
    \item Are there too many easily saturated formatting, language, or surface-presence criteria? If so, delete or downweight them.
    \item Does the output strictly satisfy the JSON array format, with each object containing only the two fields \texttt{criterion} and \texttt{weight}?
\end{itemize}

\end{promptbox}

\subsection{Task-Specific Plug-ins}
\label{app:rubric_generation_modules}
The \texttt{general} plug-in is a conservative fallback for ambiguous or underspecified tasks without a reliable specialized label. It provides broadly applicable guidance, while task-specific plug-ins offer stronger priors when a more precise label is available.
\begin{promptbox}{Task-Specific Plug-in: \texttt{general}}
\small

\begin{itemize}[leftmargin=1.5em]
    \item \textbf{Prioritize rewarding:}
    \begin{itemize}[leftmargin=1.5em]
        \item Responding to the user's most central and explicit task objective, rather than staying on related but secondary content.
        \item Satisfying the most rigid explicit constraints in the user prompt, such as output format, key content requirements, boundary conditions, restrictions, or prohibited requirements.
        \item Covering the core content necessary to complete the task, rather than only being superficially relevant.
    \end{itemize}

    \item \textbf{High-resolution criteria should prioritize evaluating:}
    \begin{itemize}[leftmargin=1.5em]
        \item Whether each part of the content truly serves the core task objective, rather than merely containing related elements.
        \item Whether key intermediate links, local support, or local elaboration advance the final result, rather than being loosely stacked together.
        \item Whether later content continues and uses information, relationships, reasoning, or conclusions already established earlier.
        \item Whether the organization of information reduces comprehension burden and makes key results or key content easier to access.
    \end{itemize}

    \item \textbf{Avoid or downweight:}
    \begin{itemize}[leftmargin=1.5em]
        \item A large number of element-presence checks, format checks, or keyword-coverage checks.
        \item Vague global criteria such as ``overall good'', ``overall natural'', or ``overall high quality''.
        \item Criteria that can be easily gamed by verbosity, templates, or mechanical enumeration.
        \item Criteria that are weakly related to the core task and add almost no new information for ranking.
    \end{itemize}

    \item \textbf{Weighting tendency:}
    \begin{itemize}[leftmargin=1.5em]
        \item The most critical task-completion items and explicit hard-constraint items should receive higher weights.
        \item Supportive, connective, and progression-related criteria that can truly distinguish near-high-quality candidates may receive medium to high weights.
        \item Easily saturated language, format, or surface-presence items should be retained only when necessary and with low weights.
    \end{itemize}
\end{itemize}

\end{promptbox}

\begin{promptbox}{Task-Specific Plug-in: \texttt{exact\_reasoning}}
\small

\begin{itemize}[leftmargin=1.5em]
    \item This task belongs to mathematical solving, logical reasoning, algorithmic problems, formal derivations, or other tasks that have clear conclusions and where process correctness is important.

    \item \textbf{This task must prioritize including the following types of criteria:}
    \begin{itemize}[leftmargin=1.5em]
        \item If the problem has a highly certain standard answer and you have high confidence in the answer, write ``the final answer is correct'' or ``the key conclusion is correct'' as one of the highest-weight criteria; when necessary, directly include the correct answer in the criterion.
        \item Whether the response directly provides the final answer or key conclusion, rather than only staying at process description.
        \item Whether the response contains the key intermediate steps, key intermediate conclusions, key equation transformations, key proof links, or key judgment basis necessary to complete the solution.
        \item Whether the key formulas, key transformations, key reasoning, or key proof steps are correct.
        \item Whether the final answer is consistent with the key derivation above, rather than having a mismatch between process and conclusion.
    \end{itemize}

    \item \textbf{High-resolution criteria for this task should prioritize:}
    \begin{itemize}[leftmargin=1.5em]
        \item Whether key intermediate conclusions truly support the final answer, rather than writing many steps without establishing the key dependency chain.
        \item Whether later steps correctly continue and use previous results, rather than writing intermediate results that are not correctly used afterward.
        \item If steps are compressed, whether the necessary logical connections are still preserved, rather than omitting key links that affect correctness.
        \item Whether the method, theorem, construction, or technique used matches the current problem, rather than rigidly applying an irrelevant template.
        \item For proof tasks, whether the argument chain is closed, whether key premises are actually used, and whether the conclusion naturally follows from the preceding content.
    \end{itemize}

    \item \textbf{Avoid or downweight the following types of criteria:}
    \begin{itemize}[leftmargin=1.5em]
        \item Avoid style, narrative, aesthetic, or emotional-impact criteria.
        \item Avoid rewarding only ``writing many steps''; instead, evaluate whether the steps are necessary, correct, mutually connected, and useful for solving the problem.
        \item Avoid assigning high weights to criteria such as ``notation is standardized'' or ``formatting is clear''; these should only be low-weight supplements.
        \item If multiple feasible solution methods exist, do not force a fixed method; instead, evaluate whether the chosen method is correct, complete, and self-consistent.
    \end{itemize}

    \item \textbf{Weighting tendency:}
    \begin{itemize}[leftmargin=1.5em]
        \item Final-answer correctness, correctness of key steps, validity of the key dependency chain, and consistency between conclusion and derivation should receive high weights.
        \item Concise expression, clear notation, and easier-to-read structure may be used as low-weight supplements.
    \end{itemize}
\end{itemize}

\end{promptbox}

\begin{promptbox}{Task-Specific Plug-in: \texttt{explanatory\_reasoning}}
\small

\begin{itemize}[leftmargin=1.5em]
    \item This task belongs to scientific explanation, technical analysis, mechanism explanation, causal reasoning, why/how questions, and ``what is X'' tasks whose core goal is explanation and helping the user understand.

    \item \textbf{This task must prioritize including the following types of criteria:}
    \begin{itemize}[leftmargin=1.5em]
        \item Whether the response truly answers the core question of ``what this is / why this happens / how this works / what mechanism it is based on'', rather than only giving a surface conclusion or scattered facts.
        \item Whether the response covers the key mechanisms, key variables, key intermediate links, key causal chains, definition boundaries, or key technical principles necessary for the explanation to hold.
        \item Whether the response avoids obvious factual errors, mechanism confusion, self-contradiction, or incorrectly treating correlation as causation.
        \item If the problem asks for explaining a concept, phenomenon, or principle, whether the response continues to provide helpful elaboration after giving the basic definition, rather than stopping at a one-sentence short answer.
    \end{itemize}

    \item \textbf{High-resolution criteria for this task should prioritize:}
    \begin{itemize}[leftmargin=1.5em]
        \item Whether key conclusions are directly supported by previous reasons, mechanisms, evidence, experimental phenomena, or reasoning chains, rather than merely stacking conclusions and terminology together.
        \item Whether intermediate mechanisms truly connect the starting conditions to the final conclusion, rather than only listing several related concepts.
        \item Whether the order of explanation reduces comprehension burden, by first establishing necessary premises and then expanding mechanisms, clarifying definitions, or explaining causality.
        \item If analogies, examples, experimental phenomena, or concrete scenarios are used, whether they genuinely help explain the mechanism or concept, rather than only adding surface richness.
        \item If the task requires comparing causes, factors, or influence paths, whether the response distinguishes different mechanisms, conditions, or contexts, rather than mixing them together.
    \end{itemize}

    \item \textbf{Avoid or downweight the following types of criteria:}
    \begin{itemize}[leftmargin=1.5em]
        \item Avoid reducing the task to short-answer factual QA criteria such as ``whether a certain fact is mentioned''.
        \item Avoid rewarding only the appearance of a conclusion without evaluating whether the explanatory chain holds.
        \item Avoid a large number of creative, aesthetic, or narrative criteria.
        \item Avoid treating terminology dumping, concept listing, or phenomenon restatement as high-value explanation criteria.
    \end{itemize}

    \item \textbf{Weighting tendency:}
    \begin{itemize}[leftmargin=1.5em]
        \item Mechanism explanation quality, closure of the causal chain, support from intermediate links to conclusions, and accuracy of definitions and key conclusions should receive high weights.
        \item Clearer terminology boundaries, more complete boundary-condition supplements, and more concise expression may be used as medium- or low-weight supplements.
    \end{itemize}
\end{itemize}

\end{promptbox}

\begin{promptbox}{Task-Specific Plug-in: \texttt{grounded\_transformation}}
\small

\begin{itemize}[leftmargin=1.5em]
    \item This task belongs to summarization, rewriting, translation, distillation, information extraction, rewriting based on given material, or other transformation tasks that must remain faithful to the input content.

    \item \textbf{This task must prioritize including the following types of criteria:}
    \begin{itemize}[leftmargin=1.5em]
        \item Whether the response faithfully preserves the original meaning, key facts, core conclusions, main arguments, or main information of the source, without substantial semantic drift.
        \item Whether the response covers the key information that must be retained by the task, rather than omitting the main thread, key conditions, key results, or key limitations.
        \item Whether the response avoids introducing new conclusions, new facts, new stances, new causal relations, or hallucinated content that is not present in the source.
        \item Whether the response satisfies the target format, compression level, target language, target style, target audience, or other explicit transformation requirements.
    \end{itemize}

    \item \textbf{High-resolution criteria for this task should prioritize:}
    \begin{itemize}[leftmargin=1.5em]
        \item Whether information selection is effective: whether high-value information is retained and low-value redundancy is removed, rather than mechanically copying sentence by sentence or listing everything with equal weight.
        \item Whether the response preserves the main logical thread, main causal chain, or main argumentative skeleton of the original text, rather than only retaining fragmented information.
        \item Whether the transformed organization makes the target content easier to understand, rather than creating new confusion, breaks, or shifts in emphasis.
        \item Whether translation or rewriting achieves the target expressive effect while remaining faithful, rather than changing the form while drifting in meaning.
        \item If the task asks for distilling key points, whether the response truly highlights the key points, rather than treating all content equally.
    \end{itemize}

    \item \textbf{Avoid or downweight the following types of criteria:}
    \begin{itemize}[leftmargin=1.5em]
        \item Avoid turning the task into an open-ended creative task; do not reward original expansion, extra invention, or decorative additions unless the user explicitly requests them.
        \item Avoid treating qualities such as ``more literary'' or ``more like original writing'' as high-value criteria when they are detached from faithfulness.
        \item Avoid only checking whether ``the word count is shorter'' or ``the wording has changed''; instead, evaluate whether the transformation is faithful and effective.
        \item Avoid too many fluency or rhetorical-polishing criteria, since these should not outweigh faithfulness and key-information preservation.
    \end{itemize}

    \item \textbf{Weighting tendency:}
    \begin{itemize}[leftmargin=1.5em]
        \item Faithfulness, key-information coverage, absence of hallucination, and preservation of the core structure should receive high weights.
        \item More concise expression, clearer organization, and stronger emphasis on key points may be used as medium- or low-weight supplements, but should not outweigh faithfulness.
    \end{itemize}
\end{itemize}

\end{promptbox}

\begin{promptbox}{Task-Specific Plug-in: \texttt{decision\_support}}
\small

\begin{itemize}[leftmargin=1.5em]
    \item This task belongs to giving advice, comparing options, choosing strategies, planning analysis, operational / management / policy judgment, or other tasks whose goal is to ``make a better decision''.

    \item \textbf{This task must prioritize including the following types of criteria:}
    \begin{itemize}[leftmargin=1.5em]
        \item Whether the response addresses the user's actual goal, rather than giving generic advice.
        \item Whether the response identifies and uses key constraints, resource limits, risk boundaries, priorities, time ranges, costs, background assumptions, or user preferences.
        \item Whether the response gives a clear recommendation, conclusion, judgment, or prioritized plan, rather than only discussing without taking a position.
        \item If multiple candidate options exist, whether the response makes a real comparison, rather than merely listing them side by side.
    \end{itemize}

    \item \textbf{High-resolution criteria for this task should prioritize:}
    \begin{itemize}[leftmargin=1.5em]
        \item Whether the recommended plan is supported by prior analysis, comparison, or reasons, rather than being decided abruptly.
        \item Whether the response identifies and addresses key trade-offs, rather than only describing benefits without costs.
        \item Whether the response explains the applicable conditions, potential risks, side effects, failure modes, or situations in which the plan should not be adopted.
        \item Whether the response provides an executable path, steps, priority order, implementation considerations, or next actions after the decision, rather than remaining abstract.
        \item If the user provides explicit preferences or constraints, whether the response truly adjusts the recommendation around these conditions, rather than outputting a generic template.
    \end{itemize}

    \item \textbf{Avoid or downweight the following types of criteria:}
    \begin{itemize}[leftmargin=1.5em]
        \item Do not force criteria of the form ``what the only standard answer is'', unless the task itself is a closed-form decision problem.
        \item Avoid rewarding shallow criteria such as ``gives multiple suggestions'', ``uses clear bullet points'', or ``has a natural tone'', which are easy to game.
        \item Avoid generic advice that does not handle constraints, compare options, give a conclusion, or provide an execution path.
        \item Avoid replacing truly useful judgment with long, principle-heavy generalities.
    \end{itemize}

    \item \textbf{Weighting tendency:}
    \begin{itemize}[leftmargin=1.5em]
        \item Goal alignment, constraint identification, option comparison, trade-off handling, clear recommendation, and executability should receive high weights.
        \item More decision-friendly structure, clearer summaries, and more useful reminders may be used as medium- or low-weight supplements.
    \end{itemize}
\end{itemize}

\end{promptbox}

\begin{promptbox}{Task-Specific Plug-in: \texttt{creative\_generation}}
\small

\begin{itemize}[leftmargin=1.5em]
    \item This task belongs to creative writing, stories, scripts, copywriting, poetry, role-playing text, or other open-ended generation tasks.

    \item \textbf{This task must prioritize including the following types of criteria:}
    \begin{itemize}[leftmargin=1.5em]
        \item Whether the response satisfies explicit writing constraints, such as topic, character, scene, genre, tone, theme, viewpoint, length, style, or prohibited requirements.
        \item Whether the response truly develops around the prompt's core, rather than writing generically, using templates, or only being superficially relevant.
        \item Whether the response completes basic structural requirements, such as beginning--development--resolution, setup--progression--response, or other organizational forms explicitly required by the prompt.
    \end{itemize}

    \item \textbf{High-resolution criteria for this task should prioritize:}
    \begin{itemize}[leftmargin=1.5em]
        \item At least half of the criteria should not be element-presence checks; they should instead prioritize evaluating relationships, function, progression, continuity, and payoff.
        \item After key events, key interactions, or key turns, whether later content shows observable changes in relationships, emotions, situation, character attitude, or narrative direction.
        \item Whether conflict, emotion, or relationship change has a trigger, response, and later continuation, rather than appearing suddenly, escalating suddenly, or ending suddenly.
        \item Whether key scenes, dialogue, imagery, or descriptions truly advance the narrative goal, character development, or theme, rather than being decorative accumulation.
        \item Whether the ending recovers the main thread, responds to earlier setup, fulfills established tension, or forms a natural closure around the core theme, rather than stopping abruptly.
        \item Whether character behavior, emotion, and expression remain basically consistent with prior setup, without unprepared functional jumps.
    \end{itemize}

    \item \textbf{Avoid or downweight the following types of criteria:}
    \begin{itemize}[leftmargin=1.5em]
        \item Do not generate criteria of the form ``what the standard answer is'' or ``what the only correct ending is''.
        \item Avoid a large proportion of factual-correctness global criteria unless the prompt explicitly requires factual grounding.
        \item Avoid a large number of shallow criteria such as ``mentions element X'', ``character Y appears'', ``has dialogue'', or ``has an ending'', which are easy to game.
        \item Avoid vague aesthetic criteria such as ``overall literary'', ``overall moving'', or ``beautiful language''; if style needs to be evaluated, it must be operationalized into concrete, judgeable features.
    \end{itemize}

    \item \textbf{Weighting tendency:}
    \begin{itemize}[leftmargin=1.5em]
        \item Explicit constraint satisfaction, main-thread completion, key progression, and continuity and payoff of state changes should receive high weights.
        \item More natural callbacks, more effective details, and more controlled closure may be used as medium- or low-weight supplements, but they must still be valuable for ranking.
    \end{itemize}
\end{itemize}

\end{promptbox}

\subsection{Two-Stage Construction of Code-Based Checkers}
\label{app:checker_construction_prompts}

We construct code-based checkers in two stages: first extracting explicit hard constraints that are deterministically checkable, and then compiling them into executable Python checker functions. This design separates constraint selection from code generation, ensuring that only surface-checkable constraints are turned into independent checkers.

\subsubsection{Hard-Constraint Extraction Prompt}
\label{app:constraint_extraction_prompt}

The first-stage prompt extracts only explicit and machine-verifiable constraints. If no valid surface-checkable constraint is found, the model is instructed to output \texttt{[null]}.

\begin{promptbox}{Prompt for Hard-Constraint Extraction}
\small

You are a constraint extraction expert. Your task is to identify hard constraints from a user instruction, but only those that can be checked by executable Python code with string matching, regex, counting, or lightweight heuristics. Be slightly precision-oriented: if a constraint is not clearly explicit and surface-checkable, do not extract it.

\textbf{Core Principles}

\begin{enumerate}[leftmargin=1.5em]
    \item Extract only explicit, machine-verifiable constraints from the following allowed types:
    \begin{itemize}[leftmargin=1.5em]
        \item \texttt{word\_count}: total word/character length requirements, including min/max/exact/range/approximate total length.
        \item \texttt{paragraph\_count}: explicit total paragraph count or total paragraph range. If blank-line separation or no horizontal rules is explicitly required, include it in the same constraint text.
        \item \texttt{sentence\_count}: explicit total sentence count or sentence count range. Exclude local counts tied to a specific section.
        \item \texttt{keyword\_count}: explicit requirement that a keyword or phrase must appear, optionally with frequency.
        \item \texttt{keyword\_exclude}: explicit prohibition of a keyword, phrase, symbol, or pattern.
        \item \texttt{response\_language}: explicit language/script requirement, such as ``written in English'', ``written in Simplified Chinese'', ``written in French''.
        \item \texttt{start\_text}: explicit requirement about how the response should begin, start with, open with, or the first sentence/phrase.
        \item \texttt{end\_text}: explicit requirement about how the response should end, close with, or the last sentence/phrase.
        \item \texttt{list\_format}: explicit list-marker or separator rules that can be checked with regex, such as numbered items, bullet items, comma-separated items, or each item on a new line.
        \item \texttt{output\_format}: explicit surface-format rules such as plain text, bold markers, fenced code block, Markdown code block language, ``no bullet points'', ``no special symbols'', ``no horizontal rules''.
        \item \texttt{punctuation\_rule}: explicit punctuation constraints such as ``avoid colons'', ``do not use exclamation marks''.
    \end{itemize}

    \item Extract only when the constraint is explicit enough for a checker:
    \begin{itemize}[leftmargin=1.5em]
        \item Good candidates: exact quoted text, explicit numbers, explicit markers like ``1.'', ``-'', commas, blank lines, code blocks, bold markers, named language, banned punctuation.
        \item Also allowed: short example-based opening/ending constraints such as ``begin with 'Sure!''' or ``begin with a sentence that acknowledges the request, such as 'Sure!'''.
        \item Do not extract open-ended templates that require semantic slot filling, such as ``use the format 'els [adjective] de [city]'''.
        \item For \texttt{keyword\_count} / \texttt{keyword\_exclude}, extract only when the instruction explicitly requires surface occurrence or prohibition, with cues such as ``include'', ``mention'', ``must contain'', ``do not use'', ``must appear'', ``exact phrase'', or quoted keyword markers.
        \item Do not convert the main subject of the task into a keyword rule. If the instruction says to discuss, explain, compare, analyze, or describe something, that alone does not mean the exact term must appear in the response.
    \end{itemize}

    \item Do \textbf{not} extract any of the following:
    \begin{itemize}[leftmargin=1.5em]
        \item Content quality constraints: creative, logical, positive, sophisticated, clear, concise, professional, etc.
        \item Semantic/topic constraints: theme, focus, explanation order that requires understanding meaning.
        \item Local structural constraints requiring semantic segmentation, such as ``the introduction should have 60 words''.
        \item Style or grammar constraints that need deep linguistic judgment, such as passive voice, imperative mood, third-person perspective, metaphor usage.
        \item Vague or preference wording such as ``try to'', ``preferably'', ``as short as possible'', ``similar phrase'' when no concrete anchor is given.
    \end{itemize}

    \item Fidelity to original text:
    \begin{itemize}[leftmargin=1.5em]
        \item The \texttt{constraint} value should quote the original wording as much as possible or be a minimal paraphrase.
        \item Preserve all numbers, keywords, quoted text, and formatting markers exactly.
        \item Output constraint text in the same language as the source instruction whenever possible.
    \end{itemize}

    \item Formatting Rules, mandatory:
    \begin{itemize}[leftmargin=1.5em]
        \item Output must be valid JSON containing only one top-level array, with no extra text.
        \item Each array element must be an object with exactly two fields: \texttt{type} and \texttt{constraint}.
        \item \texttt{type} must be one of: \texttt{word\_count}, \texttt{paragraph\_count}, \texttt{sentence\_count}, \texttt{keyword\_count}, \texttt{keyword\_exclude}, \texttt{response\_language}, \texttt{start\_text}, \texttt{end\_text}, \texttt{list\_format}, \texttt{output\_format}, \texttt{punctuation\_rule}.
        \item At most one item each for \texttt{word\_count}, \texttt{paragraph\_count}, \texttt{sentence\_count}, \texttt{response\_language}, \texttt{start\_text}, and \texttt{end\_text}.
        \item Multiple \texttt{keyword\_count}, \texttt{keyword\_exclude}, \texttt{list\_format}, \texttt{output\_format}, or \texttt{punctuation\_rule} items are allowed if they refer to distinct explicit constraints.
        \item If no valid constraint is found, output \texttt{[null]}.
    \end{itemize}
\end{enumerate}

\textbf{Examples.}

\textbf{Input 1:} Write an article introducing the development of artificial intelligence. The article must be at least 300 words, contain three paragraphs, no more than 30 sentences, must include the keyword Qwen, mention Kimi at least 3 times, and must not contain ChatGPT.

\textbf{Correct Output 1:}
\begin{verbatim}
[
  {"type": "word_count", "constraint": "at least 300 words"},
  {"type": "paragraph_count", "constraint": "three paragraphs"},
  {"type": "sentence_count", "constraint": "no more than 30 sentences"},
  {"type": "keyword_count", "constraint": "include keyword Qwen"},
  {"type": "keyword_count", "constraint": "mention Kimi at least 3 times"},
  {"type": "keyword_exclude", "constraint": "must not contain ChatGPT"}
]
\end{verbatim}

\textbf{Input 2:} Write an article about environmental protection that is positive, well-structured, with at least 200 words in the main body and no more than 5 sentences in the conclusion.

\textbf{Correct Output 2:}
\begin{verbatim}
[null]
\end{verbatim}

Explanation: Do not extract ``positive'' or ``well-structured'' (quality constraints), ``at least 200 words in the main body'' (requires semantic segmentation), or ``no more than 5 sentences in the conclusion'' (not total sentence count).

\textbf{Input 3:} Explain how solar panels work. The response should be written in English. The response should consist of five paragraphs, with a blank line separating each paragraph. The response should begin with 'Sure!'. Avoid using bullet points. The response should end with 'Solar energy matters.'.

\textbf{Correct Output 3:}
\begin{verbatim}
[
  {"type": "response_language", "constraint": "written in English"},
  {"type": "paragraph_count", "constraint": "five paragraphs, with a blank line separating each paragraph"},
  {"type": "start_text", "constraint": "begin with 'Sure!'"},
  {"type": "output_format", "constraint": "Avoid using bullet points"},
  {"type": "end_text", "constraint": "end with 'Solar energy matters.'"}
]
\end{verbatim}

\textbf{Input 4:} Give the answer in plain text. Use numbered steps (1., 2., 3., etc.). Do not use colons. Include the keyword "Tierra".

\textbf{Correct Output 4:}
\begin{verbatim}
[
  {"type": "output_format", "constraint": "in plain text"},
  {"type": "list_format", "constraint": "Use numbered steps (1., 2., 3., etc.)"},
  {"type": "punctuation_rule", "constraint": "Do not use colons"},
  {"type": "keyword_count", "constraint": "Include the keyword \"Tierra\""}
]
\end{verbatim}

\textbf{Input 5:} Discuss the relationship between ABCDF and climate policy.

\textbf{Correct Output 5:}
\begin{verbatim}
[null]
\end{verbatim}

Explanation: ``ABCDF'' is the topic being discussed, not an explicit requirement that the exact string ABCDF must appear in the response.

\textbf{Canonical examples by type.}
\begin{itemize}[leftmargin=1.5em]
    \item \texttt{word\_count}: Input: ``Write at least 300 words.'' Output: \texttt{\{"type": "word\_count", "constraint": "at least 300 words"\}}
    \item \texttt{paragraph\_count}: Input: ``Write exactly 4 paragraphs with a blank line between paragraphs.'' Output: \texttt{\{"type": "paragraph\_count", "constraint": "exactly 4 paragraphs with a blank line between paragraphs"\}}
    \item \texttt{sentence\_count}: Input: ``Use no more than 8 sentences.'' Output: \texttt{\{"type": "sentence\_count", "constraint": "no more than 8 sentences"\}}
    \item \texttt{keyword\_count}: Input: ``Include the keyword 'Qwen' at least twice.'' Output: \texttt{\{"type": "keyword\_count", "constraint": "Include the keyword 'Qwen' at least twice"\}}
    \item \texttt{keyword\_exclude}: Input: ``Do not use the word 'ChatGPT'.'' Output: \texttt{\{"type": "keyword\_exclude", "constraint": "Do not use the word 'ChatGPT'"\}}
    \item \texttt{response\_language}: Input: ``The response should be written in Spanish.'' Output: \texttt{\{"type": "response\_language", "constraint": "written in Spanish"\}}
    \item \texttt{start\_text}: Input: ``The response should begin with 'Sure!'.'' Output: \texttt{\{"type": "start\_text", "constraint": "begin with 'Sure!'"\}}
    \item \texttt{end\_text}: Input: ``The response should end with 'Thank you for reading.'.'' Output: \texttt{\{"type": "end\_text", "constraint": "end with 'Thank you for reading.'"\}}
    \item \texttt{list\_format}: Input: ``Use bullet points beginning with '- '.'' Output: \texttt{\{"type": "list\_format", "constraint": "Use bullet points beginning with '- '"\}}
    \item \texttt{output\_format}: Input: ``Put the implementation in Python code blocks.'' Output: \texttt{\{"type": "output\_format", "constraint": "in Python code blocks"\}}
    \item \texttt{punctuation\_rule}: Input: ``Avoid using colons.'' Output: \texttt{\{"type": "punctuation\_rule", "constraint": "Avoid using colons"\}}
\end{itemize}

When a real input matches one of the canonical types above, imitate the nearest canonical example and preserve the original wording as much as possible.

\textbf{Final Reminder.}
\begin{itemize}[leftmargin=1.5em]
    \item If there are no valid constraints, output \texttt{[null]} directly.
    \item Your output must contain only the JSON array, with no explanations, prefixes, or suffixes.
\end{itemize}

\textbf{User message:} Now process the user input: \texttt{\{question\}}

\end{promptbox}

\subsubsection{Constraint-to-Code Compilation Prompt}
\label{app:constraint_to_code_prompt}

The second-stage prompt compiles each extracted constraint into \texttt{check\_following(instruction, response)}. Generated code is execution-validated; failed items are retried and eventually replaced with \texttt{[null]}.

\begin{promptbox}{Prompt for Constraint-to-Code Compilation}
\small

You are a powerful code assistant capable of converting a list of extracted constraint items into corresponding Python validation functions.

Your task: Based on the given constraint list, generate a self-contained Python function named \texttt{check\_following(instruction, response)}.

The goal is to return \texttt{True} if all constraints are satisfied, and \texttt{False} otherwise.

\textbf{Strictly follow the rules below:}
\begin{enumerate}[leftmargin=1.5em]
    \item The output must be a pure Python code list, exactly matching the format shown in the ``Example Output''.
    \item Each constraint item must correspond to one independent Python function string.
    \item Each function must be self-contained and include necessary imports, e.g., \texttt{re}. The use of external libraries such as \texttt{nltk} is strictly prohibited.
    \item If the input is \texttt{[null]}, you must directly output \texttt{[null]} without any extra characters.
    \item Do not return anything other than the code list.
\end{enumerate}

\textbf{Important: Violating the format will cause a system failure. You must:}
\begin{itemize}[leftmargin=1.5em]
    \item Never modify the function signature: \texttt{def check\_following(instruction, response)}.
    \item Never change the number of list elements; it must exactly match the number of input constraints.
    \item Prefer deterministic regex, string, and counting logic.
    \item Lightweight heuristics are allowed for \texttt{response\_language} and soft opening/ending checks, but keep them simple and executable without external libraries.
    \item Be conservative: if a constraint is phrased too vaguely for a reliable surface checker, generate the weakest faithful executable checker you can, rather than an overly strict one.
\end{itemize}

\textbf{Example Input 1:}
\begin{lstlisting}[style=promptcode]
[
  {"type": "word_count", "constraint": "at least 300 words"},
  {"type": "paragraph_count", "constraint": "3 paragraphs"},
  {"type": "sentence_count", "constraint": "no more than 30 sentences"},
  {"type": "keyword_count", "constraint": "include keyword Qwen"},
  {"type": "keyword_count", "constraint": "mention keyword Kimi at least three times"},
  {"type": "keyword_exclude", "constraint": "must not contain 'ChatGPT'"}
]
\end{lstlisting}

\textbf{Example Output 1:}
\begin{lstlisting}[style=promptcode]
[
  "import re\n\ndef check_following(instruction, response):\n    chinese = len(re.findall(r'[\u4e00-\u9fff]', response))\n    english = len(re.findall(r\"[a-zA-Z]+(?:[-'][a-zA-Z]+)*\", response))\n    return chinese + english >= 300",
  "import re\n\ndef check_following(instruction, response):\n    paragraphs = [p.strip() for p in re.split(r'\\n\\s*\\n', response.strip()) if p.strip()]\n    return len(paragraphs) == 3",
  "import re\n\ndef check_following(instruction, response):\n    sentences = re.split('[.!?\\u3002\\uff01\\uff1f]+', response.strip())\n    sentences = [s.strip() for s in sentences if s.strip()]\n    return len(sentences) <= 30",
  "import re\n\ndef check_following(instruction, response):\n    return len(re.findall(r'\\bQwen\\b', response, re.IGNORECASE)) >= 1",
  "import re\n\ndef check_following(instruction, response):\n    return len(re.findall(r'\\bKimi\\b', response, re.IGNORECASE)) >= 3",
  "import re\n\ndef check_following(instruction, response):\n    return not bool(re.search(r'ChatGPT', response, re.IGNORECASE))"
]
\end{lstlisting}

\textbf{Example Input 2:}
\begin{lstlisting}[style=promptcode]
[
  {"type": "response_language", "constraint": "written in English"},
  {"type": "start_text", "constraint": "begin with 'Sure!'"},
  {"type": "end_text", "constraint": "end with 'Solar energy matters.'"},
  {"type": "list_format", "constraint": "Use numbered steps (1., 2., 3., etc.)"},
  {"type": "output_format", "constraint": "output should be in Python code blocks"},
  {"type": "punctuation_rule", "constraint": "Do not use colons"}
]
\end{lstlisting}

\textbf{Example Output 2:}
\begin{lstlisting}[style=promptcode]
[
  "import re\n\ndef check_following(instruction, response):\n    text = response.strip()\n    if not text:\n        return False\n    if re.search(r'[\\u4e00-\\u9fff\\u3040-\\u30ff\\uac00-\\ud7af\\u0400-\\u04ff\\u0600-\\u06ff]', text):\n        return False\n    english_hits = len(re.findall(r'\\b(the|and|is|are|of|to|in|that|for|with|on|as)\\b', text, re.IGNORECASE))\n    latin_hits = len(re.findall(r'[A-Za-z]', text))\n    return latin_hits >= 20 and english_hits >= 2",
  "def check_following(instruction, response):\n    return response.lstrip().startswith('Sure!')",
  "def check_following(instruction, response):\n    return response.rstrip().endswith('Solar energy matters.')",
  "import re\n\ndef check_following(instruction, response):\n    return bool(re.search(r'(?m)^\\s*\\d+\\.\\s+', response))",
  "import re\n\ndef check_following(instruction, response):\n    return bool(re.search(r'```(?:python)?\\n[\\s\\S]+?\\n```', response, re.IGNORECASE))",
  "def check_following(instruction, response):\n    return ':' not in response"
]
\end{lstlisting}

\textbf{Canonical code patterns by type:}
\begin{itemize}[leftmargin=1.5em]
    \item \texttt{word\_count}
\begin{lstlisting}[style=promptcode]
Input: {"type": "word_count", "constraint": "at least 300 words"}
\end{lstlisting}

    \item \texttt{paragraph\_count}
\begin{lstlisting}[style=promptcode]
Input: {"type": "paragraph_count", "constraint": "exactly 4 paragraphs with a blank line between paragraphs"}
\end{lstlisting}

    \item \texttt{sentence\_count}
\begin{lstlisting}[style=promptcode]
Input: {"type": "sentence_count", "constraint": "no more than 8 sentences"}
\end{lstlisting}

    \item \texttt{keyword\_count}
\begin{lstlisting}[style=promptcode]
Input: {"type": "keyword_count", "constraint": "Include the keyword 'Qwen' at least twice"}
\end{lstlisting}

    \item \texttt{keyword\_exclude}
\begin{lstlisting}[style=promptcode]
Input: {"type": "keyword_exclude", "constraint": "Do not use the word 'ChatGPT'"}
\end{lstlisting}

    \item \texttt{response\_language}
\begin{lstlisting}[style=promptcode]
Input: {"type": "response_language", "constraint": "written in Spanish"}
\end{lstlisting}

    \item \texttt{start\_text}
\begin{lstlisting}[style=promptcode]
Input: {"type": "start_text", "constraint": "begin with 'Sure!'"}
\end{lstlisting}

    \item \texttt{end\_text}
\begin{lstlisting}[style=promptcode]
Input: {"type": "end_text", "constraint": "end with 'Thank you for reading.'"}
\end{lstlisting}

    \item \texttt{list\_format}
\begin{lstlisting}[style=promptcode]
Input: {"type": "list_format", "constraint": "Use bullet points beginning with '- '"}
\end{lstlisting}

    \item \texttt{output\_format}
\begin{lstlisting}[style=promptcode]
Input: {"type": "output_format", "constraint": "in Python code blocks"}
\end{lstlisting}

    \item \texttt{punctuation\_rule}
\begin{lstlisting}[style=promptcode]
Input: {"type": "punctuation_rule", "constraint": "Avoid using colons"}
\end{lstlisting}
\end{itemize}

For every real constraint, choose the nearest canonical pattern above and adapt only the keyword, number, operator, punctuation mark, or anchored text.

\textbf{Additional guidance by type:}
\begin{itemize}[leftmargin=1.5em]
    \item \texttt{response\_language}: use script detection for Chinese/Japanese/Korean/Cyrillic/Arabic when possible; for English/Spanish/French/Catalan and other Latin-script languages, use a lightweight stopword heuristic plus script checks.
    \item \texttt{start\_text} / \texttt{end\_text}: if the constraint contains quoted text, match that text exactly after trimming outer whitespace.
    \item \texttt{list\_format}: check only visible markers or separators.
    \item \texttt{output\_format}: check only surface formatting, such as fenced code blocks, plain text, bold markers, absence of bullet points, absence of horizontal rules, or absence of special symbols.
    \item \texttt{punctuation\_rule}: check the required punctuation inclusion/exclusion directly.
    \item \texttt{keyword\_count} / \texttt{keyword\_exclude}: only use exact token or phrase presence checks when the constraint clearly asks for explicit surface appearance or prohibition. Do not strengthen a vague thematic constraint into an exact-match keyword requirement.
\end{itemize}

\textbf{User message:} Now process the user input: \texttt{\{checkers\}}
\end{promptbox}

\subsection{Rubric Judging and Global Scoring Prompts}
\label{app:rubric_judging_prompts}

This subsection presents the prompts used for model-based reward computation. Rubric-based scoring evaluates each rubric item independently with a three-way label: \texttt{yes}, \texttt{part}, or \texttt{no}. Global scoring uses a separate judge prompt that produces a brief explanation and a final numeric score in double square brackets.

\begin{promptbox}{System Prompt for Rubric Judging}
\small

You are a strict evaluation function.

Your task is to judge whether the AI response satisfies the given RUBRIC.

You will be provided with:
\begin{itemize}[leftmargin=1.5em]
    \item A user question
    \item An AI response
    \item A single RUBRIC item
\end{itemize}

\textbf{RATING DEFINITIONS}
\begin{itemize}[leftmargin=1.5em]
    \item \texttt{yes}: The response fully satisfies the RUBRIC with no meaningful flaws.
    \item \texttt{part}: The response addresses the RUBRIC but has minor omissions, ambiguity, or limited coverage.
    \item \texttt{no}: The response does not satisfy the core requirement of the RUBRIC.
\end{itemize}

\textbf{EVALUATION RULES}
\begin{itemize}[leftmargin=1.5em]
    \item Evaluate ONLY against the given RUBRIC.
    \item Ignore verbosity, style, or fluency unless explicitly required by the RUBRIC.
    \item Do NOT infer unstated intent or give credit for partially implied content.
    \item Be strict: assign \texttt{yes} only if the requirement is clearly and completely met.
\end{itemize}

\textbf{OUTPUT FORMAT}

Return exactly one label:

\texttt{\textbf{yes}}/
\texttt{\textbf{part}}/
\texttt{\textbf{no}}

Do not output JSON, explanations, ids, punctuation, or anything else.

\end{promptbox}

\begin{promptbox}{User Prompt for Rubric Judging}
\small

\textbf{TASK: Evaluate the Response based on the single RUBRIC below.}

\textbf{[User Question]}

\texttt{\{question\}}

\textbf{[AI Assistant Response]}

\texttt{\{answer\}}

\textbf{[Evaluation RUBRIC]}

\texttt{\{rubric\}}

Output exactly one token: \texttt{yes}, \texttt{part}, or \texttt{no}.

\end{promptbox}

\begin{promptbox}{System Prompt for Global Scoring}
\small

You are a strict scoring function.

Follow the user's instructions exactly.

\end{promptbox}

\begin{promptbox}{User Prompt for Global Scoring}
\small

\textbf{Task Description}

Please act as an impartial judge and evaluate the quality of the response provided by an AI assistant to the user query displayed below.

\textbf{Notes:}
\begin{enumerate}[leftmargin=1.5em]
    \item Your evaluation should consider factors such as the helpfulness, relevance, and accuracy of the response, but need not consider depth or level of detail of the response.
    \item Begin your evaluation by providing a short explanation.
    \item Be as objective as possible. After providing your explanation, please rate the response on a scale of 0 to 10. For your rating, only give a number between 0 and 10 inclusive, do not use any markdown, and do not put any text after your final rating.
    \item Important: The final numeric rating must be enclosed in double square brackets \texttt{[[ ]]}. Do not add any text outside the brackets.
\end{enumerate}

\textbf{[Query]}

\texttt{\{question\}}

\textbf{[Response]}

\texttt{\{answer\}}

\textbf{[Your judgement]}

\end{promptbox}
\clearpage

\end{document}